\documentclass[10pt]{article} 
\usepackage[preprint]{tmlr}

\usepackage{amsmath,amsfonts,bm}

\def\eqref#1{equation~\ref{#1}}

\def\1{\bm{1}}

\def\vc{{\bm{c}}}
\def\vd{{\bm{d}}}
\def\ve{{\bm{e}}}

\def\vk{{\bm{k}}}

\def\vw{{\bm{w}}}

\def\mV{{\bm{V}}}
\def\mW{{\bm{W}}}
\def\mX{{\bm{X}}}

\DeclareMathAlphabet{\mathsfit}{\encodingdefault}{\sfdefault}{m}{sl}
\SetMathAlphabet{\mathsfit}{bold}{\encodingdefault}{\sfdefault}{bx}{n}

\def\sR{{\mathbb{R}}}
\def\sS{{\mathbb{S}}}

\def\sV{{\mathbb{V}}}

\usepackage{amsmath}
\usepackage{amssymb}
\usepackage{amsthm}
\usepackage{mathtools}
\usepackage[retainorgcmds]{IEEEtrantools}
\usepackage{hyperref}
\usepackage{url}
\usepackage{algorithm}
\usepackage{algpseudocode}
\usepackage{orcidlink}
\usepackage{graphicx}
\usepackage{svg}
\usepackage{subcaption}
\usepackage{orcidlink}

\newcommand*\mytitle{Analyzing Transformer Dynamics as Movement through Embedding Space} 
\title{\mytitle} 

\author{\name Sumeet S. Singh \orcidlink{0000-0002-5323-9678} \email ssingh@turnitin.com \\
      \addr Turnitin LLC \\
      Oakland, CA 94612, USA
      }

\def\month{MM}  
\def\year{YYYY} 
\def\openreview{\url{https://openreview.net/forum?id=XXXX}} 

\newcommand{\seqw}{{\left< \vw_i \right>}}
\newcommand{\seqd}{{\left< \vd_i \right>}}
\newcommand{\seqc}{{\left< \vc_i \right>}}

\newcommand{\seqdotw}{{\left< \dot{\vw}_i \right>}}

\theoremstyle{definition}
\newtheorem{prop}{Proposition}
\newtheorem{cor}{Corollary}[prop]
\newtheorem{result}{Result}
\newtheorem*{disc}{Discussion}
\newtheorem{property}{Property}
\theoremstyle{remark}
\newtheorem*{pf}{Argument}
\newtheorem*{defn}{Definition}

\begin{document}
\maketitle
\begin{abstract}
Transformer based language models exhibit intelligent behaviors such as understanding natural language, recognizing patterns, acquiring knowledge, reasoning, planning, reflecting and using tools. This paper explores how their underlying mechanics give rise to intelligent behaviors. Towards that end, we propose framing Transformer dynamics as movement through embedding space. Examining Transformers through this lens reveals key insights, establishing a Theory of Transformers: 1) Intelligent behaviours map to paths in Embedding Space which, the Transformer random-walks through during inferencing. 2) LM training learns a probability distribution over all possible paths. `Intelligence' is learnt by assigning higher probabilities to paths representing intelligent behaviors. No learning can take place \textit{in-context}; context only narrows the subset of paths sampled during decoding. 5) The Transformer is a self-mapping composition function, folding a context sequence into a context-vector such that it's proximity to a token-vector reflects its co-occurrence and conditioned probability. Thus, the physical arrangement of vectors in Embedding Space determines path probabilities. 6) Context vectors are composed by aggregating features of the sequence's tokens via a process we call the \textit{encoding walk}. \textit{Attention} contributes a - potentially redundant - \textit{association-bias} to this process. 7) This process is comprised of two principal operation types: \emph{filtering} (data independent) and \emph{aggregation} (data dependent). This generalization unifies Transformers with other sequence models. Building upon this foundation, we formalize a popular semantic interpretation of embeddings into a ``concept-space theory'' and find some evidence of it's validity.
\end{abstract}

\section{Introduction}
Transformers \citep{DBLP:journals/corr/VaswaniSPUJGKP17} which started as text sequence modelers \citep{devlin2019bert, raffel2020exploring} generalize to unseen tasks via zero and few shot learning \citep{GPT-3}. Transformer based instruction tuned  \citep{ouyang2022training} large language models (LLMs) exhibit intelligent abilities enabling them to be used as general purpose intelligence machines \citep{openai2023gpt4}. The zero and few shot learning abilities also extend across modalities \citep{reed2022gato, driess2023palme, girdhar2023imagebind}. Further, prompting techniques such as chain-of-thought (CoT) \citep{https://doi.org/10.48550/arxiv.2201.11903, kojima2023large} have been developed on top of LLMs and now freeform conversations are possible. Very impressive intelligent behaviors have been demonstrated e.g., instruction based source code \citep{https://doi.org/10.48550/arxiv.2107.03374} and image generation \citep{https://doi.org/10.48550/arxiv.2102.12092} plus a variety of capabilities deemed as the beginnings of AGI \citep{bubeck2023sparks, wei2022emergent}. These abilities make it appear as if Transformers can 1) understand and follow instructions, 2) think, plan, reason and explain themselves, 3) even possess a Theory of Mind \citep{kosinski2023theory} and 4) comprehend multiple modalities at the same time. And most recently, LLMs are being trained to act as autonomous agents, appearing to observe, plan, reason, reflect, use external tools \& APIs and act \citep{schick2023toolformer, yao2023react, karpas2022mrkl, park2023generative} even in a multimodal space \citep{liang2023taskmatrixai, lu2023chameleon}. However, as \citeauthor{GPT-3} note: \textit{understanding precisely how few-shot [and zero-shot] learning works is an important unexplored direction for future research}. In this work we attempt to answer this question and explain how intelligent behaviours arise from the underlying dynamics. Our contributions are: 1) We cast Transformer layers as self-mapping operators in embedding space 2) cast encoding and decoding processes as dynamic walks in this space 3) analyze the models in detail in this new light and develop a generalized architecture that unifies other sequence models and attention-free architectures 4) provide a grounded explanation of intelligent behaviours dispelling some prevailing misconceptions and 5) formalize and test a popular semantic interpretation of embeddings.

\section{Embedding Space Centric View of Transformers}
\fancyhead[L]{\mytitle}
\begin{figure}[htb!]
\begin{center}
\includegraphics[width=\textwidth]{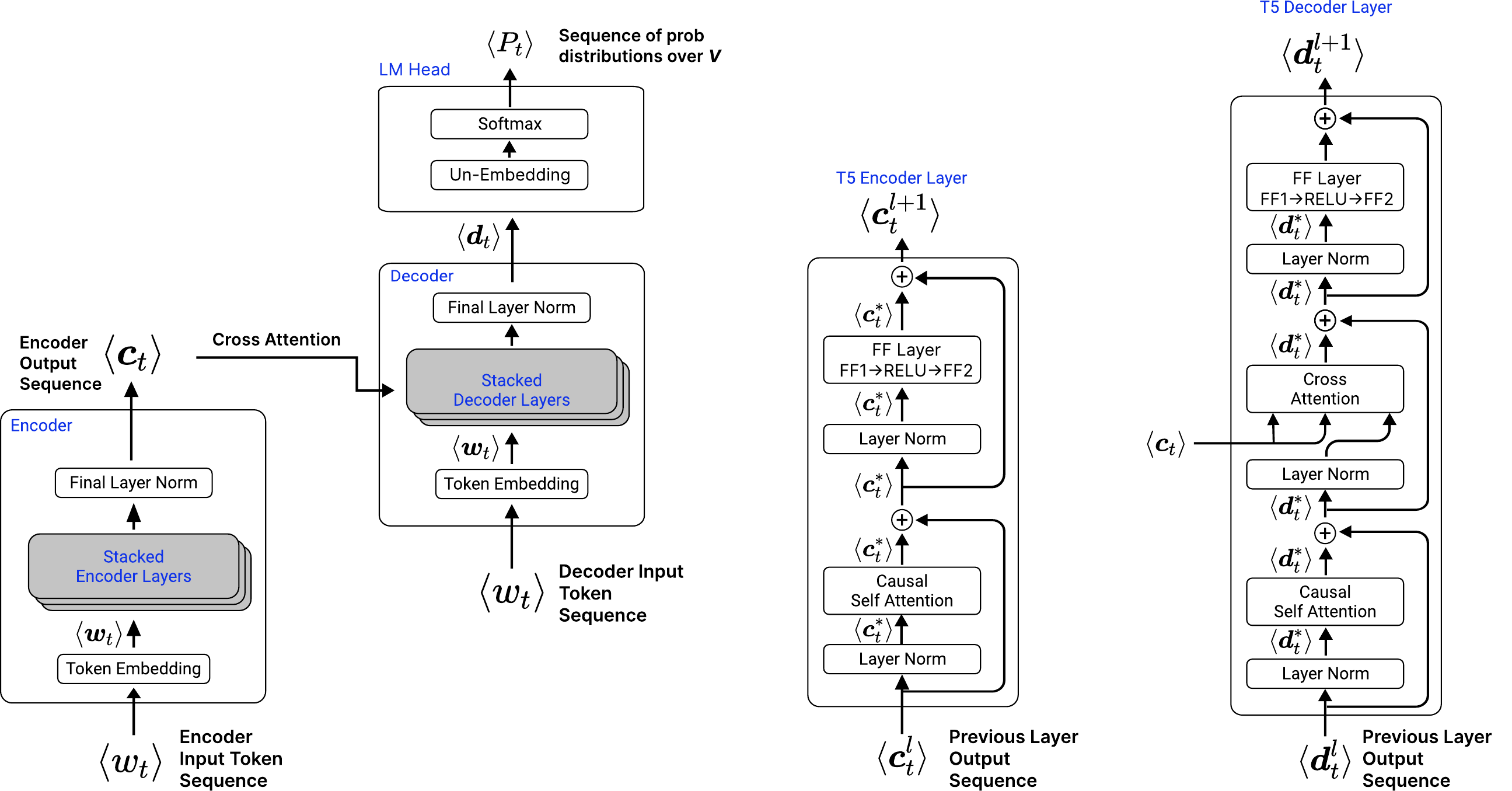}
\end{center}
\caption{ Architecture of a T5 Transformer. Encoder and decoder stacks (left), encoder layer (middle) and decoder layer (right).  $\vc_t^*$ and $\vd_t^*$ denote intermediate vectors.}
\label{fig-T5}
\end{figure}


While various Transformer architectures exist, they broadly categorize into three types: 1) The T5 \citep{raffel2020exploring} style encoder-decoder, 2) GPT3 \citep{GPT-3} style decoder only and 3) BERT \citep{devlin2019bert} style encoder only architecture. Since we're interested in generative decoding, we focus on the first two architectures only and since the encoder-decoder architecture is a superset of the other two, we shall refer to the diagram of Figure \ref{fig-T5}.
\begin{defn}[Embedding Vector]
We define an embedding vector (vector) as a hidden activation $\ve \in \sR^D$ where $D = d_{model}$ is the hidden size of the model. For convenience, activations produced inside the encoder and decoder stacks are denoted as $\vc$ (or $\vc_t$ or $\vc_i$) and $\vd$ (or $\vd_t$ or $\vd_i$) respectively. All token embeddings are vectors by definition. But in addition ...
\end{defn}
\begin{prop}
\label{prop-residual-stream}
All size $D$ hidden activations i.e., inputs \& outputs of the encoder \& decoder stacks, individual layers, summation points and layer norm i.e., those marked on Fig. \ref{fig-T5} as $\vw_t, \vc_t, \vc_t^*, \vd_t \text{ and } \vd_t^*$ are also embedding vectors. 
\end{prop}
\begin{pf}
If this was not true, the Transformer would not be able to coherently: 1) Sum vectors across (sub)layers via residual connections, 2) Compare input and output layer vectors; this happens implicitly when output $\vd_t$ of topmost layer is multiplied with the un-embedding matrix which is usually tied with the input embedding matrix (Fig. \ref{fig-T5}). This amounts to an inner-product - i.e., comparison since inner-product is the similarity metric - between $\vd_t$ and input (token) embeddings, 3) Export encoder activations into all layers of the decoder via. cross-attention or 4) Sum outputs across attention heads\footnote{Multi-headed attention can be trivially refactored as the sum of independent attention paths. See \ref{appendix-attention-head-math}.} 
For all of these operations to be coherent, the vector dimensions must have the same \emph{meaning} across the (sub) layers and stacks.
\cite{elhage2021mathematicalAnthropic} implicitly support this viewpoint via their concept of ``residual streams''. \cite{geva2022transformer} and \cite{dar2022analyzing} also posit that all layers operate in the same vector space although they are referring to a Vocabulary Space. RNNs similarly, update a residual state vector at each step and layer. This is one of the foundational tenets of this paper.
\end{pf}
\begin{cor}
It follows from prop. \ref{prop-residual-stream} that attention blocks, feedforward and layer-norm layers, individual Transformer layers, the encoder \& decoder stacks and the entire model are all self-mapping n-ary operators ($f: \sR^{D\times n} \rightarrow \sR^D$) mapping a sequence of one or more input vectors $\langle\ve_i\rangle$ to an output vector $\ve$. Note that since the attention layer aggregates its inputs, $\ve$ is \textit{composed} from $\langle\ve_i\rangle$. We call these \textit{composite} vectors. The above mentioned (sub)layers therefore, are \underline{n-ary composition functions}.
\end{cor}

\begin{figure}[htb]
\begin{center}
  \begin{subfigure}{0.475\textwidth}
    \includegraphics[width=\linewidth]{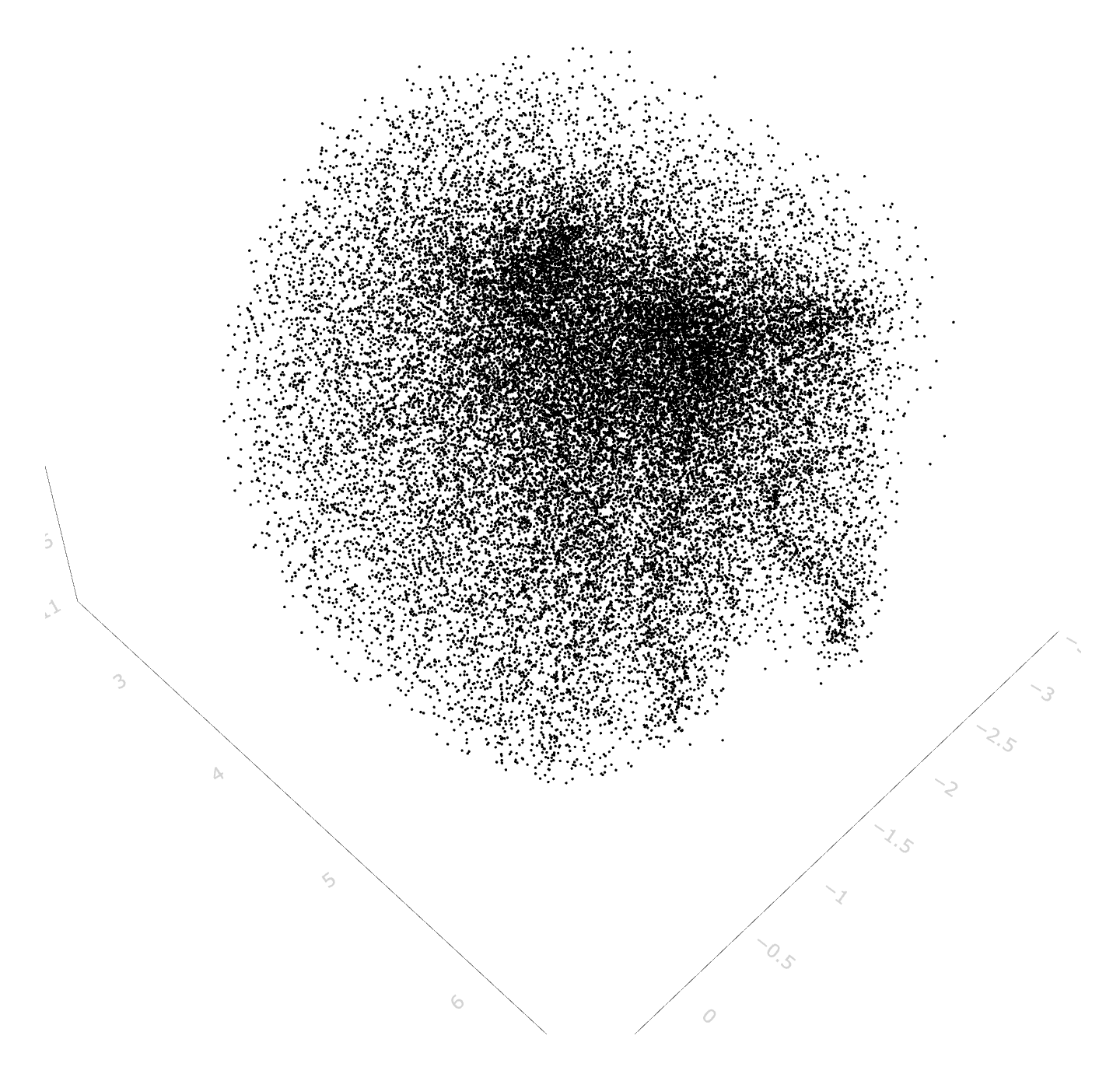}
    \caption{}
    \label{fig-embedding-sphere-a}
  \end{subfigure}
  \begin{subfigure}{0.475\textwidth}
    \includegraphics[width=\linewidth]{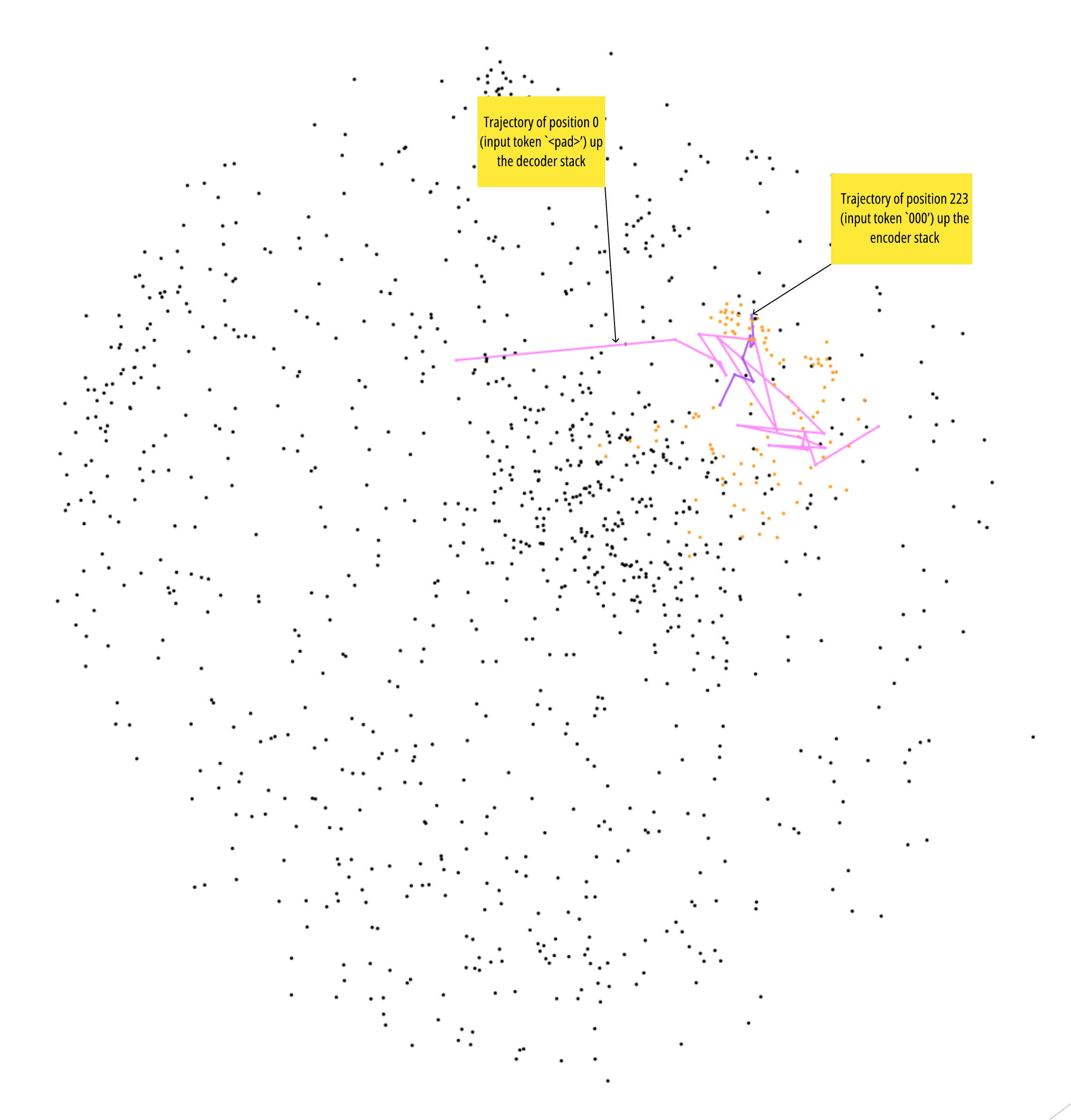}
    \caption{}
    \label{fig-trajectories}
  \end{subfigure}

\end{center}

\caption{ a) 3D UMAP projection of the set of token embedding vectors $\sV$ trained with negative inner-product as the distance metric. The vectors are obtained from a t5-small model fine-tuned with few-shot examples of freeform answer grading. The 3d space organizes in a roughly spherical shape because vector magnitudes are bounded and the distance between vectors is governed largely by their angular distance. (b) Encoding walks of figs. \ref{fig-encoding-walk-a} and \ref{fig-encoding-walk-b} mapped onto a 3D UMAP projection of $\sS$. Orange dots are input tokens.}
\label{fig-embedding-sphere}
\end{figure}

\begin{defn}
We define Embedding Space as the topological space $\sR^D$ where inner-product is the similarity metric (Figs. \ref{fig-trajectories} and \ref{fig-embedding-sphere}). Further, we define $\sS = \sV \cup \{\ve\}$ as the set of all possible embedding vectors.
\end{defn}

\begin{figure}[htbp]
\begin{center}
  \centering
  \begin{subfigure}{0.90\textwidth}
    \includegraphics[width=\linewidth]{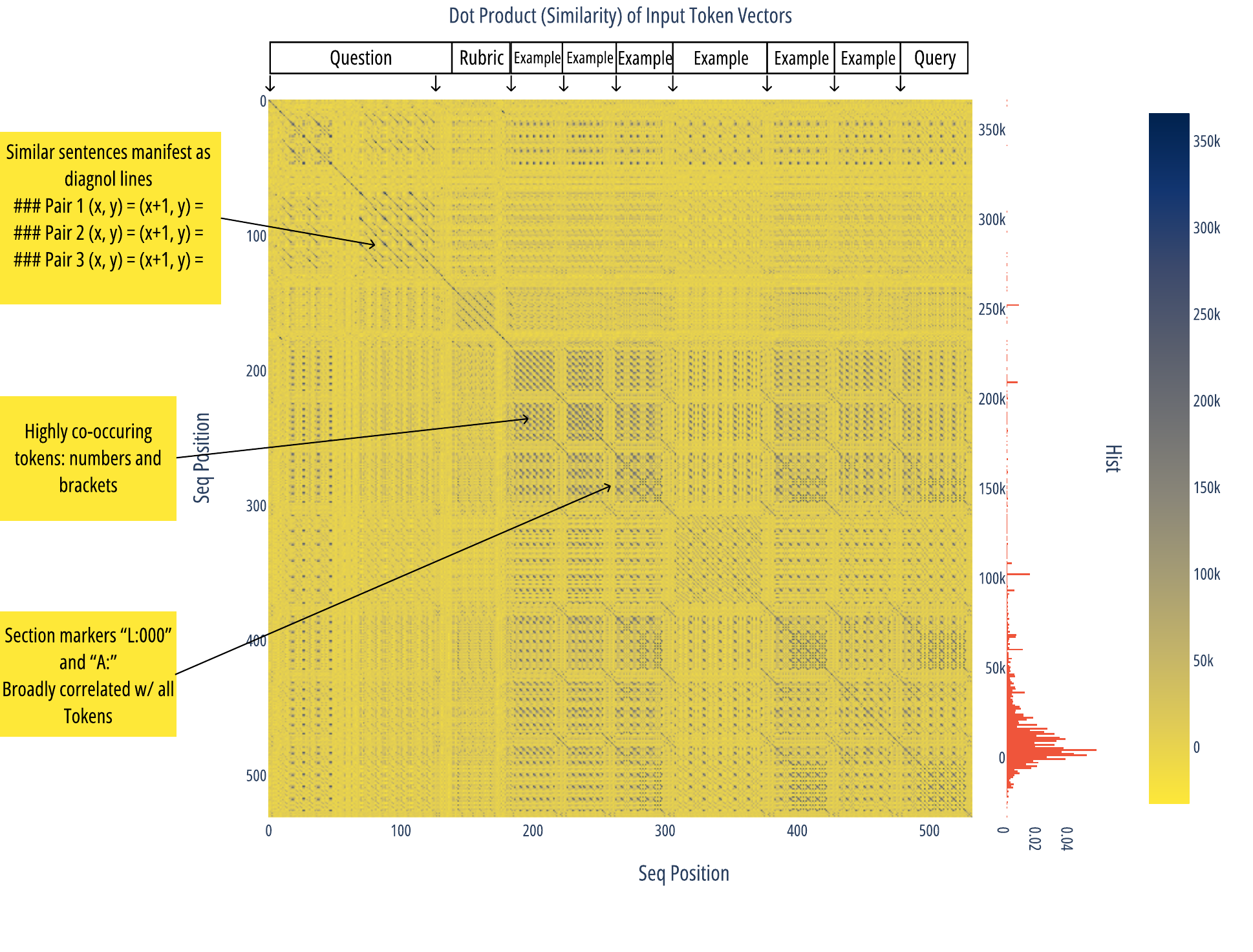}
    \caption{Input Layer}
    \label{fig-layer-heatmaps-a}
  \end{subfigure}
  \begin{subfigure}{0.475\textwidth}
    \includegraphics[width=\linewidth]{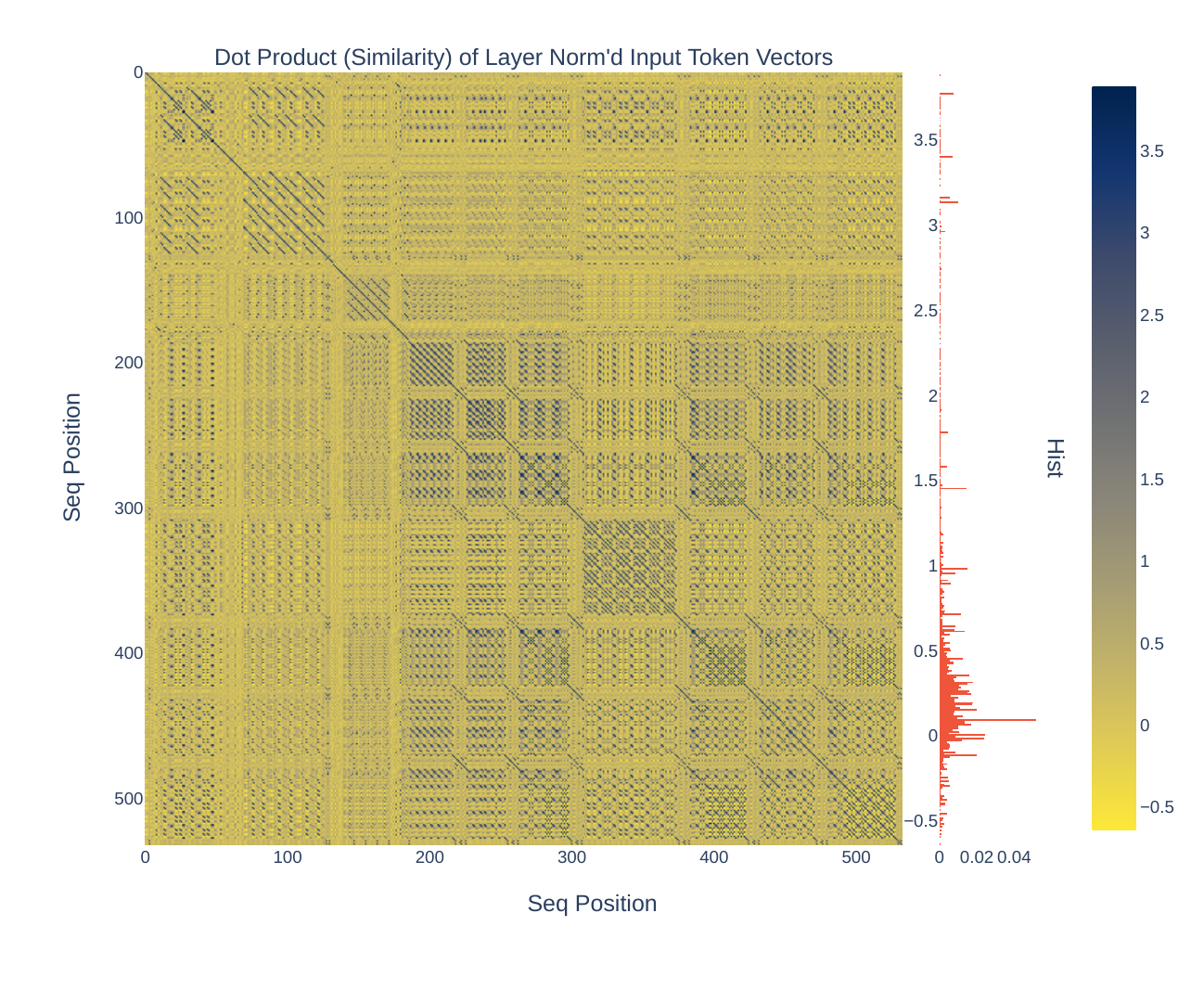}
    \caption{Layer-Norm'd Inputs}
    \label{fig-layer-heatmaps-b}
  \end{subfigure}
  \begin{subfigure}{0.475\textwidth}
    \includegraphics[width=\linewidth]{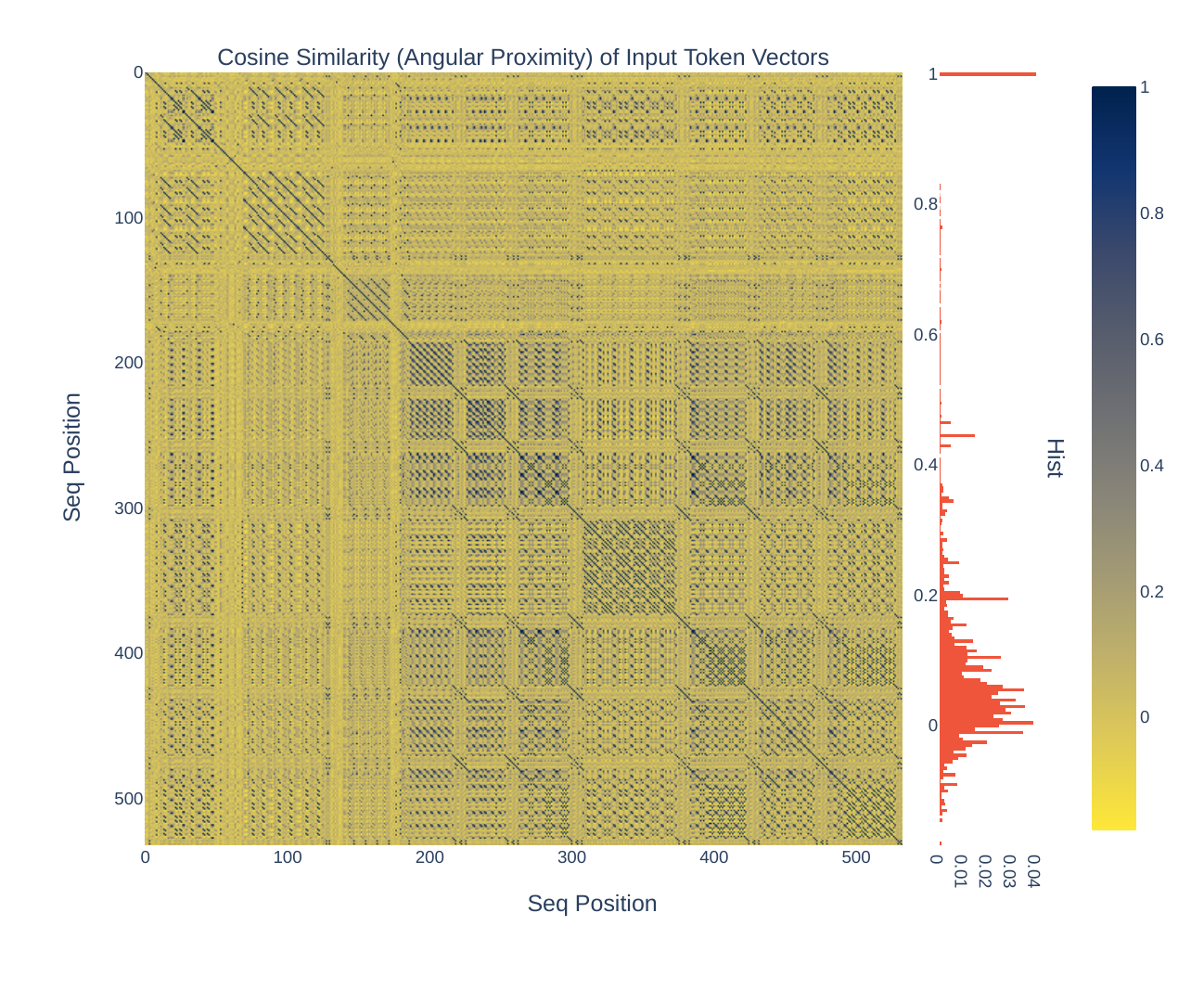}
    \caption{ Cosine Similarity i.e., Angular Proximity}
    \label{fig-layer-heatmaps-c}
  \end{subfigure}
\end{center}
\caption{Similarity maps of a vector sequences produced by the few-shot sample in section \ref{sec-appendix-fewshot-example} inside a T5-small model fine-tuned as a few-shot classifier. (a) Map of nput token embeddings. Formatting tokens that mark section boundaries (such as `Q:', `A:' and `L:000') co-occur weakly with all tokens and therefore form light bands. Highly co-occurring tokens like numbers and brackets form dark clusters. Thus the model appears to recognize the formatting pattern of the few-shot sample. (b) The pattern is more pronounced after vectors are layer-norm'd in layer 1 or (c) compared with cosine-similarity; indicating that co-occurring / associated vectors cluster together by angular distance. Additional maps of pretrained models are shown in appendix \ref{appendix-similarity-maps}.}
\label{fig-layer-heatmaps}
\end{figure}

\begin{disc}
The above findings are important because they imply that all layers of the Transformer operate at the same level of abstraction, contradicting the traditional view that higher layers of a neural network operate at progressively higher levels of abstraction. Importantly, this fact can be exploited to interpret hidden activations and weights via similarity comparisons (Fig. \ref{fig-layer-heatmaps}) and nearest neighbor analyses with token embeddings (Fig. \ref{fig-encoding-walk}).
\end{disc}
\subsection{The Decoding Walk}
\label{sec-decoding-semantics}
In this section we analyse causal decoding within this framework. Later sections delve into the layers. As a reminder causal decoding entails predicting the next token given an input token sequence, part of which may itself have been generated\footnote{In encoder-decoder architectures the input sequence is a concatenation of encoder and decoder inputs.}, . Sequence positions also called time steps, start at 1 and are denoted by $t$ or $i$. An analysis of causal decoding in this light yields equations \ref{eqn-defn-begin} - \ref{eqn-defn-end}.
\begin{IEEEeqnarray}{lCl}
    w_t \quad & \equiv & \quad \text{input token id at position } t \;; t >= 1 \label{eqn-defn-w_t} \label{eqn-defn-begin} \\
    \mV \quad&\equiv&\quad \text{token embedding matrix} \;; \text{size } V \times D \\
    \sV \quad&\equiv&\quad \{\vw_i\}_{i=1}^V \quad \subset \quad \sS \;; \text{set of all token embedding vectors} \label{eqn-defn-V}  \\
    \vw_t \quad&\equiv&\quad \text{word embedding vector of } w_t  \\
    \quad&\equiv&\quad \mV_{{w_t},:} \;; \text{row } w_t \text{ of } \mV \\
    {\langle \vw_i \rangle}_1^t \quad&\equiv&\quad  \text{embedded input sequence from position } 1 \text{ through } t \label{eqn-defn-vw-seq}\\ 
    \quad&=&\quad \text{concated encoder and decoder inputs in case of encoder-decoder arch} \IEEEnonumber\\
    f_{\Sigma;\theta} &:&\quad \sV^n \to \sS \;; \text{causal composition function characterizing the model} \label{eqn-defn-sigma} \\
    \vd_t, \vd_{{\langle \vw_i \rangle}^t_1} &\equiv&\quad \text{\emph{decoding vector} ; output vector of decoder at position } t \label{eqn-defn-d_t}  \label{eqn-defn-d_w} \\
    \quad&=&\quad f_{\Sigma;\theta}\left( {\langle \vw_i \rangle}_1^t \right) \;; \text{causal composition} \label{eqn-causal-composition}\\
    {\langle \vd_i \rangle}_1^{t} \quad&\equiv&\quad \text{\emph{decoding walk} ; the sequence of decoding vectors through position } t \label{eqn-defn-walk} \\
    P_t, P_{{\langle \vw_i \rangle}^t_1} \;\;&\equiv&\quad \text{probability distribution over } \sV \text{ output by the model at step } $t$  \\
    \quad&=&\quad softmax \left(\vd_t \mV^T \right) \;; \text{ the Logits Layer}  \\
    \quad&=&\quad softmax \left(\vd_t \odot \sV \right) \;; \odot \equiv \text{inner product} = \text{ similarity metric of } \sS \label{eqn-P_t}\\
    \quad&=&\quad softmax \left(  f_{\Sigma;\theta}\left( {\langle \vw_i \rangle}_1^t \right) \odot \sV \right)  \label{eqn-markovian-sampling}\\
    \vw_{t+1} \quad&\sim&\quad P_t \;; \text{the decoding process}  \label{eqn-decoding-sampling} \label{eqn-defn-end}
\end{IEEEeqnarray}
 
Equations \ref{eqn-defn-sigma} - \ref{eqn-causal-composition} show that the model is a composition function $f_{\Sigma;\theta}$ that composes a sequence of token vectors $\seqw_1^t$ into a single vector representation $\vd_t \equiv \vd_{\seqw_1^t} \in \sS$.
The generative decoding process generates a sequence, one token at a time, by sampling a token embedding $\vw_{t+1}$ from probability distribution $P_t$ which was produced at position t (eqn. \ref{eqn-decoding-sampling}). 
Since $P_t$ is a function of the previously traversed path ${\langle \vw_i \rangle}_1^t$ (eqn. \ref{eqn-defn-vw-seq} and \ref{eqn-markovian-sampling}), it follows that the decoding process is a non-Markovian random walk in $\sS$. We denote the sequence of output vectors produced by this process as ${\langle \vd_i \rangle}_1^t$ and term it the \emph{decoding walk} (eqn. \ref{eqn-defn-walk}).
Next, note that $P_t$ is a function of the inner-product between $\vd_t$ and all the token vectors $\vw_i$ in $\sV$ (equations \ref{eqn-defn-V} and \ref{eqn-P_t}) i.e., the relative location of $\vd_t$\footnote{i.e, relative w.r.t. words in the vocabulary} \underline{characterizes} the probability distribution of the next token which in turn determines the \emph{decoding walk} which in turn determines the generated sequence $\seqw$.
\begin{defn}
We capture this observation into a new term -  \textit{organization} of $\sS$ - which denotes the arrangement in $\sR^D$ of all vectors in $\sS$\footnote{\textit{Shape} of $\sS$ is another good name for this property. If on the other hand, $\sV$ were a continuous space - to account for soft-prompts for e.g. - then $\sS$ would be continuous too in which case, \textit{curvature} of $\sS$ would be a more appropriate term since it determines the decoding trajectories.}.
\end{defn}
\begin{result}[\textbf{Intelligence is Emergent}]
Therefore, given a fixed decoding procedure, the \textit{organization} of $\sS$ completely defines decoding walks. Since all the knowledge, intelligence \& skills expressed by the model are the result of \emph{decoding walks}, it follows that these abilities are embodied in the \textit{organization} of $\sS$. They are emergent in the classical sense since they are a property of the system, not intrinsic to specific parts.
\end{result}
Mechanically speaking, these abilities are determined by the same factors that determine the \textit{organization} of $\sS$ viz. 1) the information capacity of $\sR^D$; which increases with $D$, 2) the \textit{organization} of token embeddings $\sV$ and 3) the vector composition function $f_{\Sigma;\theta}$ which is defined by model architecture and parameters. 
This analyses unifies neural sequence models that have a residual stream i.e., $f_{\Sigma;\theta}$ could equally be implemented by a RNN, LSTM \citep{graves2014generating} and the newer post transformer architectures \citep{gu2022efficiently, fu2023hungry, poli2023hyena, sun2023retentive} . Finally, these analyses are equally applicable across modalities.

\subsection{Vector Composition: The Encoding Walk}
\label{sec-concept-composition}

\begin{IEEEeqnarray}{lCl}
    f_{C;t}^l \quad&:&\quad \sS^n \to \sS \text{ slice of encoder layer } l \text{ at position } t \text{ - a composition function} \label{eqn-defn-cc-begin}\\
    \vc^l_t &\equiv&\quad \text{vector output by encoder layer } l \text{ at position } t \IEEEnonumber \\
    \quad&=&\quad f_{C;t}^l\left( {\left< \vc_i^{l-1} \right>_1^T} \right) \\
    f_C^l \quad&:&\quad \sS^n \to \sS^n \text{ encoder layer } l \text{ as a seq-to-seq composition function} \IEEEnonumber \\
    \quad&=&\quad \text{concat} \left( {\left< \vc^l_t \right>}_1^T \right) = \text{concat} \left( f_{C;t}^l\left( {\left< \vc_i^{l-1} \right>_1^T} \right) \right) \IEEEnonumber \\
    f_C \quad&:&\quad  \sS^n \to \sS^n \text{ encoder stack as a seq-to-seq function} \IEEEnonumber \\
    \quad&=&\quad f_C^L \circ f_C^{L-1} \cdots \circ f_C^1 \label{eqn-chain-cc} \\
    \vc_t &\equiv&\quad \text{vector output by encoder stack (after final layer norm)} \text{ at position } t \IEEEnonumber \\
    \seqc_{\seqw_1^T} \;\;&\equiv&\quad \text{vector sequence composed by encoder stack from input } \seqw_1^T  \IEEEnonumber\\
    \quad&=&\quad f_C\left(\seqw_1^T \right) \\
    f_D^l \quad&:&\quad \sS^n \to \sS \text{ composition func of decoder layer } l \label{eqn-defn-sigma-d} \IEEEyesnumber \\
    \vd_t^l, \vd_{\seqw^t_1}^l \;\;&\equiv&\quad \text{vector composed by decoder layer } l \text{ at step } t \text{ given input sequence } \seqw^t_1 \IEEEnonumber\\
    \quad&=&\quad f_D^l\left( \seqc_{\seqw_1^T} \sqcup {\langle\vd^{l-1}_t\rangle}_{T+1}^t \right) \quad;\, \sqcup \text{ denotes concatenation} \IEEEnonumber \\
    \quad&=&\quad f_D^l\left( f_C(\seqw_1^T) \sqcup \langle\vd^{l-1}_{\seqw_1^t}\rangle_{T+1}^t \right) \\
    \quad&&\quad ;\,T = \text{ context length, } {\langle \vd^0_t \rangle}_{T+1}^t = \seqw^t_{T+1} \text{, the decoder input sequence}\IEEEnonumber \\
    f_{\Sigma;\theta} \quad&:&\quad \sV^n \to \sS \text{ the entire model framed as a composition function}\IEEEnonumber \\
    &&\quad ;\, \text{itself composed from } f_D^l \text{ and } f_C \label{eqn-defn-cc-end} 
\end{IEEEeqnarray}


Equations \ref{eqn-defn-cc-begin} through \ref{eqn-defn-cc-end} formalize the encoder-decoder Transformer architecture as an algebra. Each layer is cast as a composition function. Next, we analyze the layer wise functions $f_C^l$ and $f_D^l$ that are composed of the \textit{attention layer} and other functions that we simply generalize as \textit{filters}. We identify vectors by their position regardless of (sub)layer they may be in e.g., $\vc_i$ is the i'th vector. Traversal up through the (sub) layers is viewed as `movement' of the `same' vector through $\sS$ i.e., the \textit{encoding walk}.

\paragraph{Filters}
\label{sec-filters}
The layers include point-wise transforms that we call \textit{filters} ($filter: \sS \to \sS$). These are (see Figure \ref{fig-T5}) : 1) Layer norm 2) the key, query, and value-output projection matrices $\mW_{k,h}, \mW_{q,h}$ and $\mW_{vo,h}$\footnote{ $\mW_{vo,h} = \mW_{v,h} \times \mW_{o,h}$. See sec. \ref{appendix-attention-head-math} for definitions and derivations.} respectively in the attention heads and 3) the nonlinear feed-forward layers. They are \textit{static} functions in the sense that their output only depends on the input vector i.e., there is no interaction between vectors and therefore these are *not* composition functions. 

\begin{prop}
\label{result-filter-selects-features}
Filters are static feature selectors / extractors.
\end{prop}
\begin{pf}
Under prop. \ref{prop-residual-stream} self-mapping filters (layer-norm and feed-forward layers and $\mW_{vo,h}$) can only reshape (rotate) and / or amplify or attenuate (stretch) a vector. The low-rank matrices $W_{k,h}$ and $W_{q,h}$ can be viewed to perform the same operations but into a sub-space (\citet{elhage2021mathematicalAnthropic} support this view). The non-linear activation function (RELU or similar) in the feedforward layer is clearly a filter since it blocks (clamps to 0) all negative values and allows the rest through.
\end{pf}
Note that since linear transforms cause a mixing of dimensions of $\sR^D$  (also called channel mixing \citep{sun2023retentive}) this implies that features are redundantly represented across dimensions.

\subsubsection{Attention Layer}
\label{sec-attention}
\begin{figure}[htb!]
\begin{center}
  \centering
      \begin{subfigure}{0.35\textwidth}
        \includegraphics[width=\linewidth]{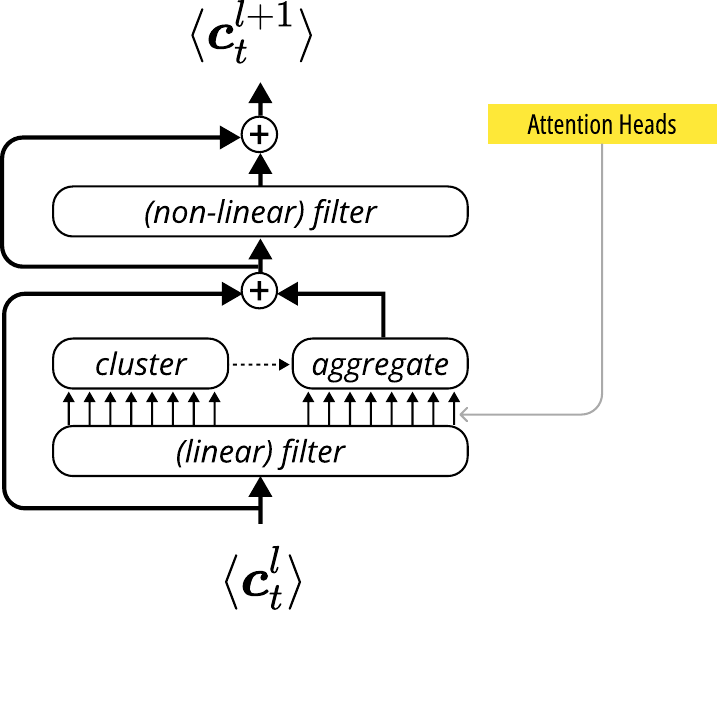}
        \caption{Encoder layer of T5}
        \label{fig-tta-a}
      \end{subfigure}
      \hfill
      \begin{subfigure}{0.25\textwidth}
        \includegraphics[width=\linewidth]{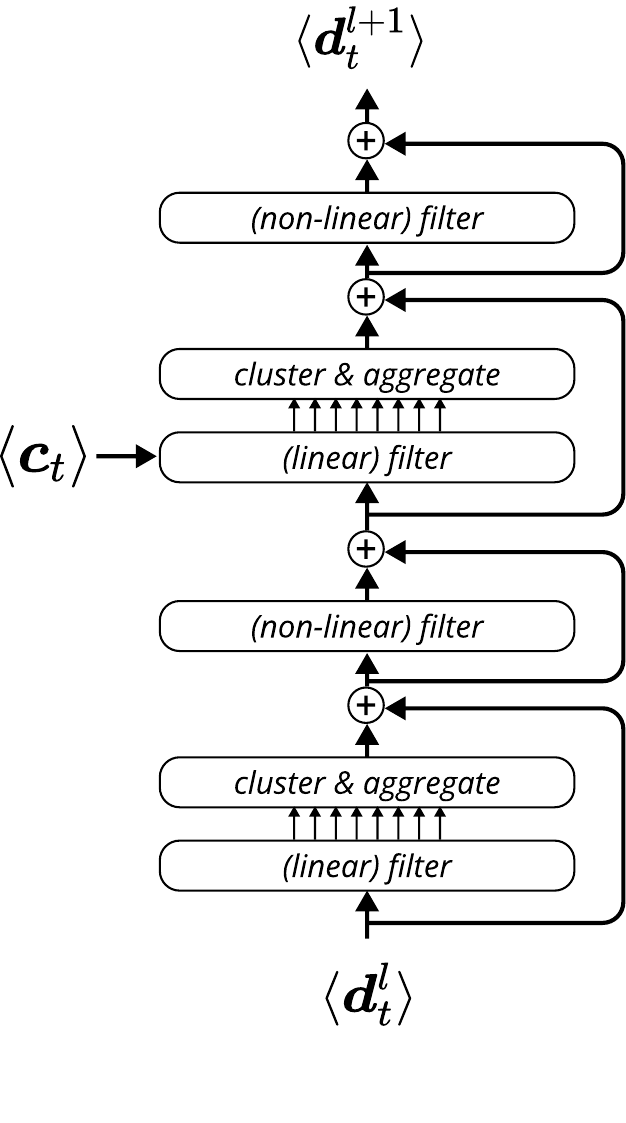}
        \caption{Decoder layer of T5}
        \label{fig-tta-b}
      \end{subfigure}
      \hfill
      \begin{subfigure}{0.35\textwidth}
        \includegraphics[width=\linewidth]{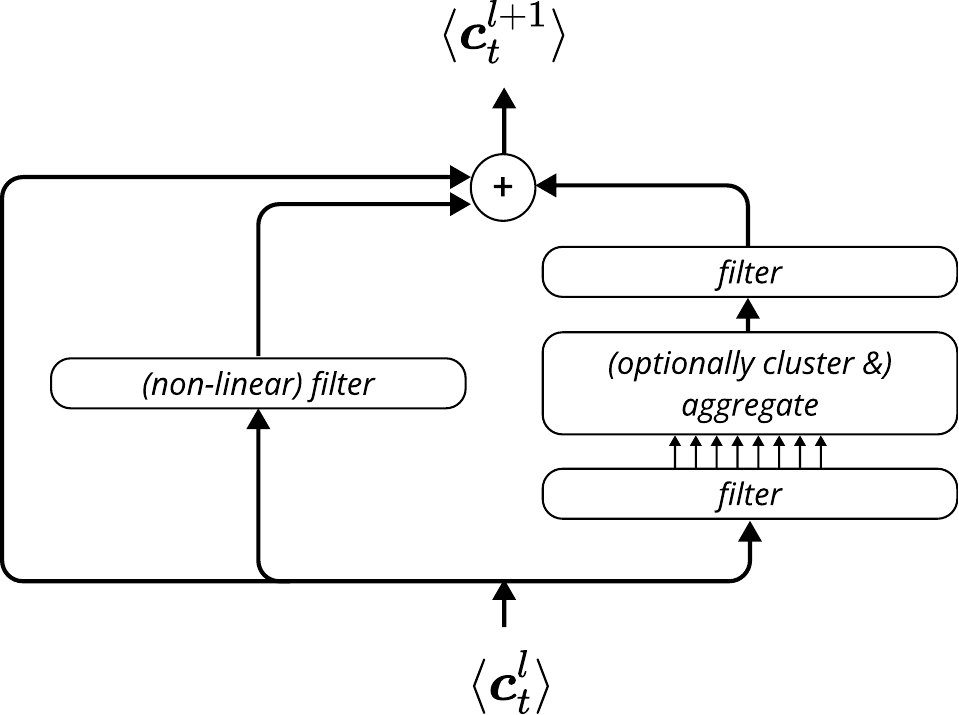}
        \caption{Generalized layer accounting for architectures with parallel feed-forward and attention layers (GPT-J6B, GPT-Neox 20B, Tiiuae/Falcon). Clustering denotes attention-based / associative clustering which is optional if we include attention-free Transformers.}
        \label{fig-tta-c}
      \end{subfigure}
  \end{center}
\caption{ The transformer architecture generalized in terms of filters and soft-clustering operations.}
\label{fig-tta}
\end{figure}

The \textit{multiheaded attention} operation can be refactored as independent attention paths per attention head by slicing $\mW_{o}$ into H slices $\mW_{o,h}$ one per head, who's outputs are summed (see section \ref{appendix-attention-head-math} for derivations). Each head therefore independently \emph{aggregates context} around a query position, with the (filtered) context vectors being weighted by their normalized similarity (softmaxed inner product) to the (filtered) query and by their position in the sequence (via relative position biases).
Therefore, attention can be generalized into the following two steps. 1) \textbf{Cluster}: compute a \emph{soft-cluster} for each query vector $\vw_t$; use attention weights as the degree of cluster membership i.e., ``gather similar stuff from selected locations'' and 2) \textbf{Aggregate}: aggregate the cluster by membership score i.e., compute the soft cluster centroid. In encoding walk parlance, the vector at query position moves to the centroid of its soft-cluster.
\begin{result}
\label{result-att-is-feature-aggregation}
This amounts to the following sequence of operations: \emph{filter} $\rightarrow$ \emph{cluster \& aggregate}$\rightarrow$ \emph{filter}. The two linear filters $\mW_{v,h}$ and $\mW_{o,h}$ can be merged together into $\mW_{vo,h}$ (section \ref{appendix-attention-head-math}) to yield \emph{filter} $\rightarrow$ \emph{cluster \& aggregate}.
\end{result}
\begin{result}[Generalized Layer Architecture]
\label{result-generalized-arch}
Adding feed-forward layers and residual connections into the picture we can generalize the Transformer architecture as shown in (Figure \ref{fig-tta-c}).    
\end{result}

\begin{figure}[htb!]
\begin{center}
  \centering
      \begin{subfigure}{0.475\textwidth}
        \includegraphics[width=\linewidth]{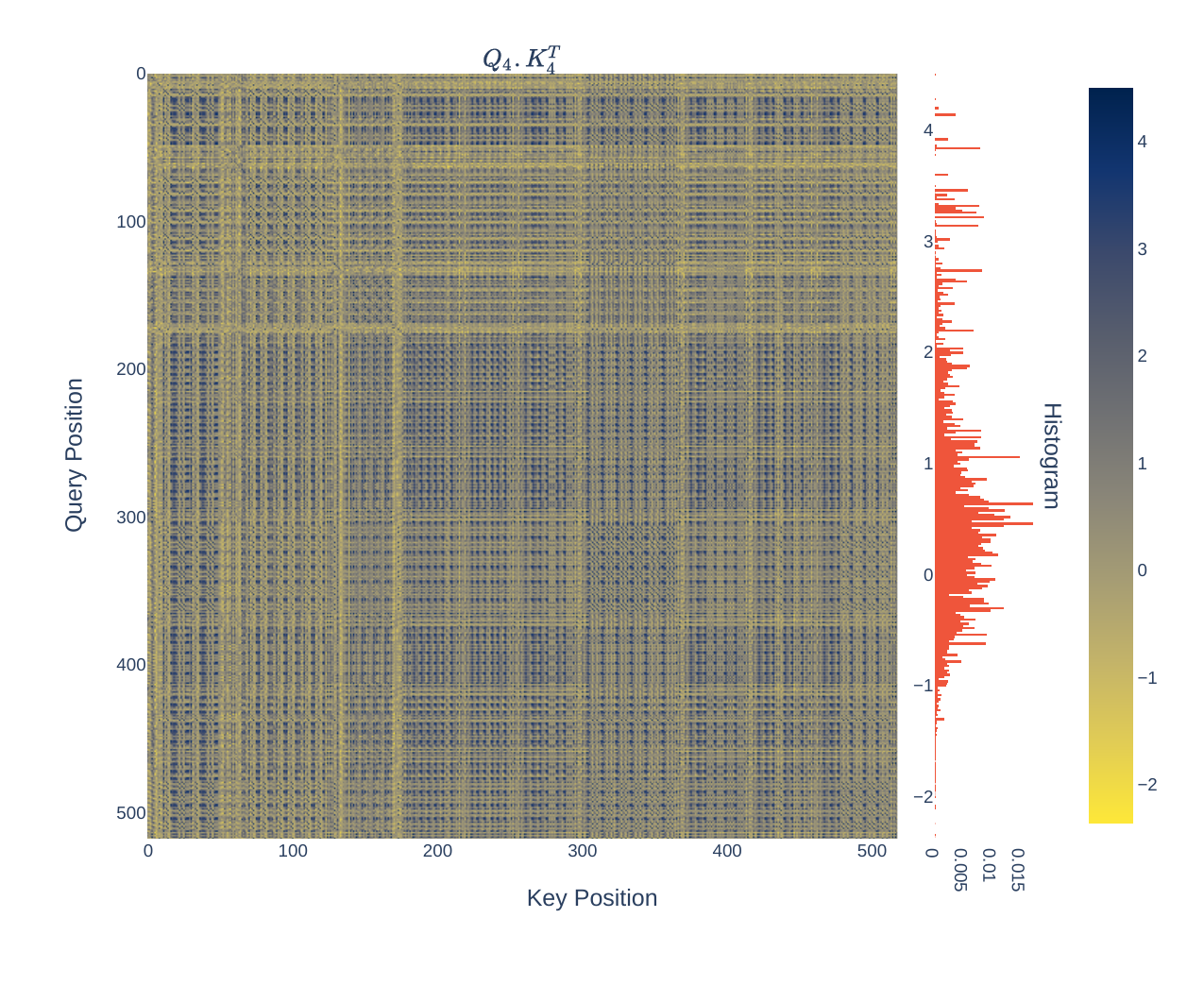}
        \caption{Attention Scores: Inner product of query and key vectors without bias}
        \label{fig-pos-enc-a}
      \end{subfigure}
      \hfill
      \begin{subfigure}{0.475\textwidth}
        \includegraphics[width=\linewidth]{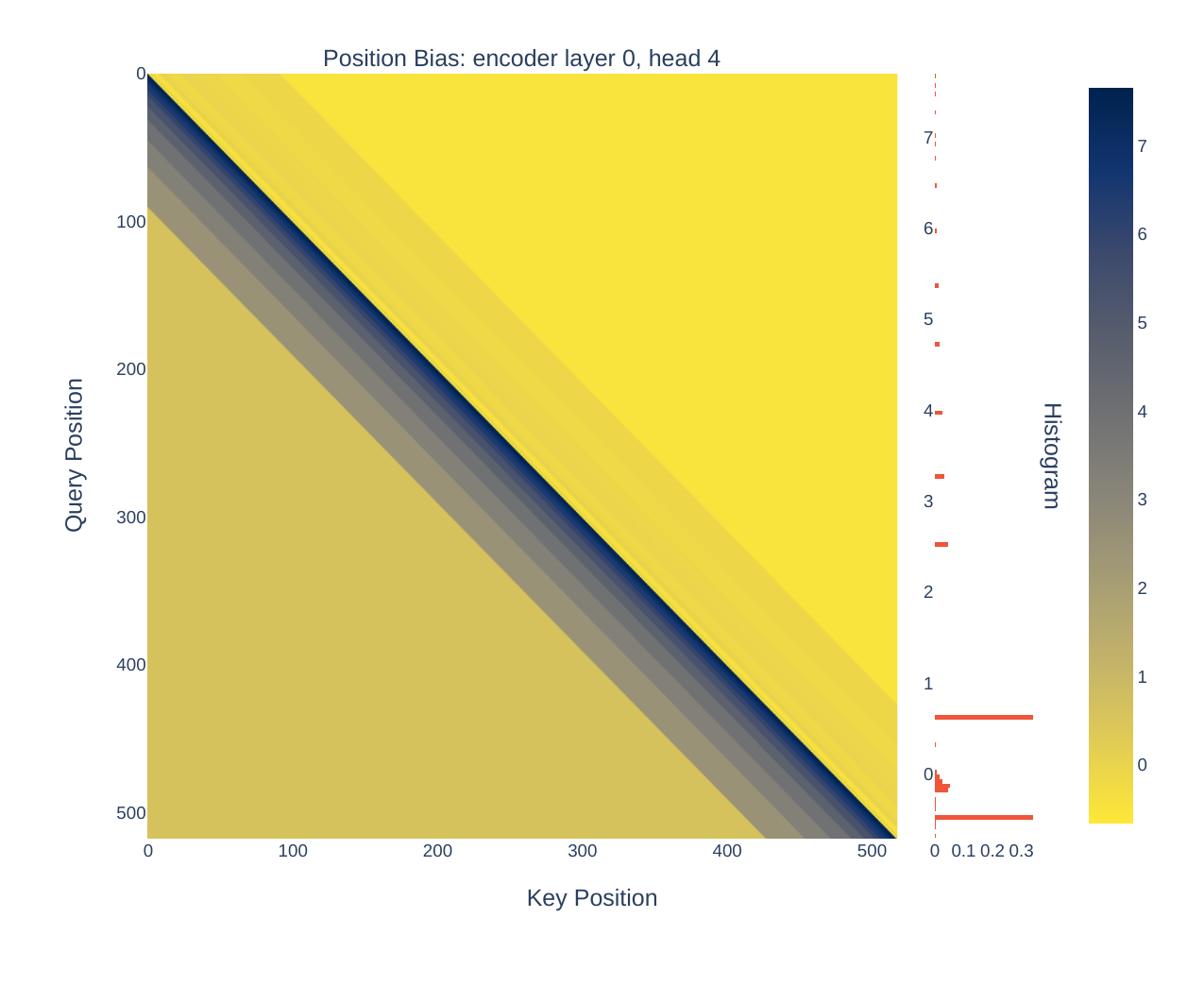}
        \caption{Position biases with negative self bias (-0.66)}
        \label{fig-pos-enc-b}
      \end{subfigure}
      \begin{subfigure}{0.475\textwidth}
        \includegraphics[width=\linewidth]{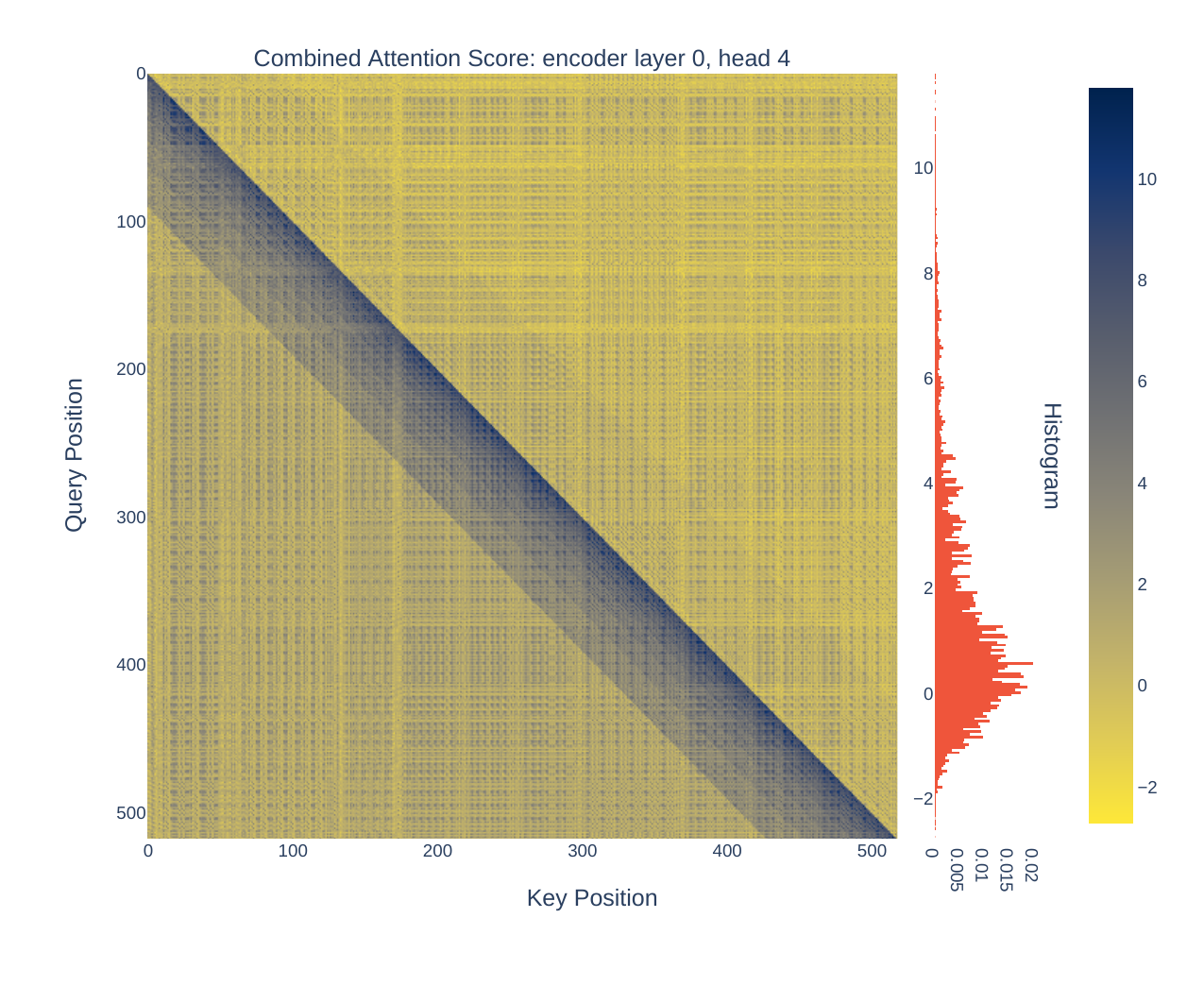}
        \caption{Attention weights before softmax = Attention scores + position biases. Position bias significantly influences attention scores.}
        \label{fig-pos-enc-c}
      \end{subfigure}
      \hfill
      \begin{subfigure}{0.475\textwidth}
        \includegraphics[width=\linewidth]{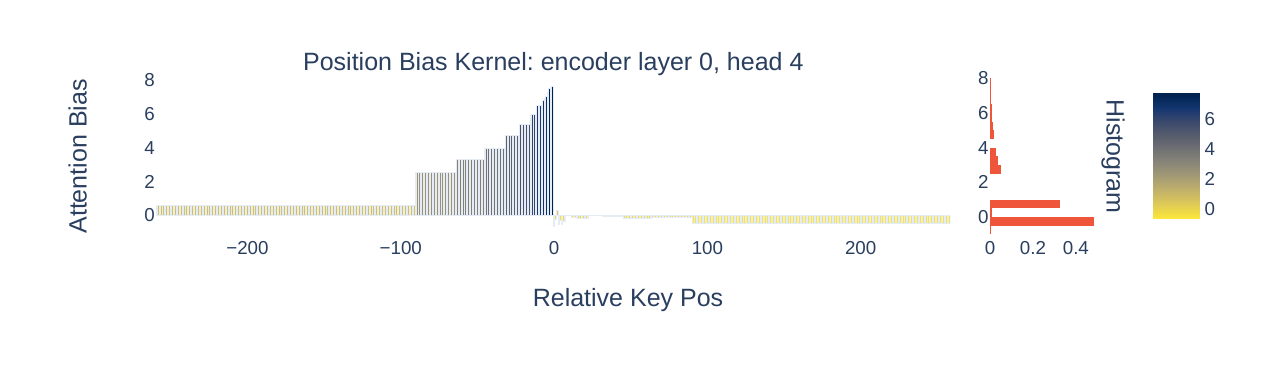}
        \caption{Position kernel corresponding to \ref{fig-pos-enc-b} and \ref{fig-pos-enc-c}. Position zero (self) is in the middle. Amplifies a band on the left side. Right hand side positions including the middle (self) are suppressed.}
        \label{fig-pos-enc-d}
      \end{subfigure}
  \end{center}
\caption{ Pretrained Flan-T5-XL encoder attention layer activation maps.}
\label{fig-pos-enc}
\end{figure}

\begin{figure}[htb!]
\begin{center}
  \centering
      \begin{subfigure}{0.475\textwidth}
        \includegraphics[width=\linewidth]{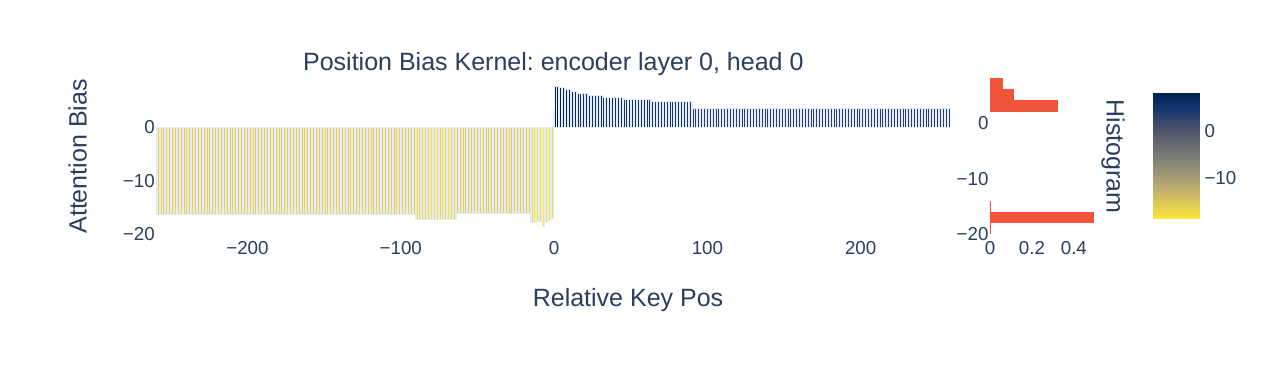}
      \end{subfigure}
      \hfill
      \begin{subfigure}{0.475\textwidth}
        \includegraphics[width=\linewidth]{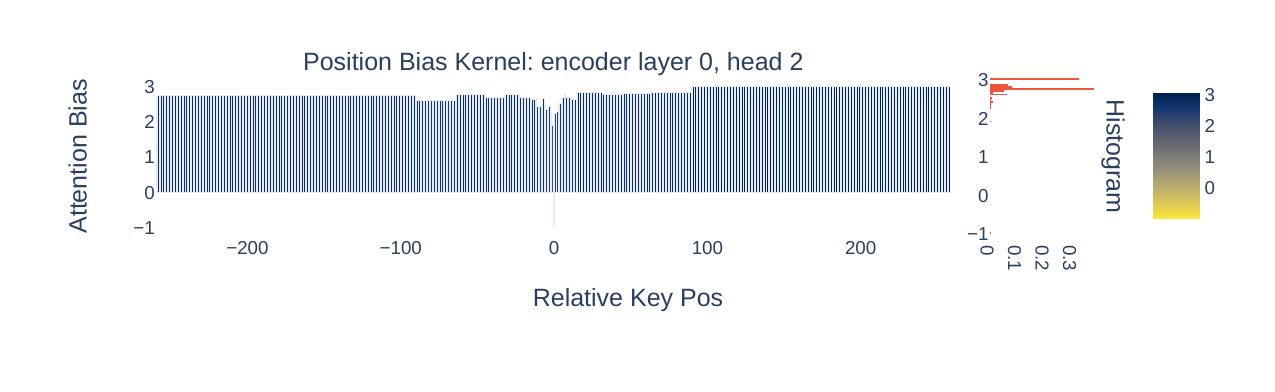}
      \end{subfigure}
      \begin{subfigure}{0.475\textwidth}
        \includegraphics[width=\linewidth]{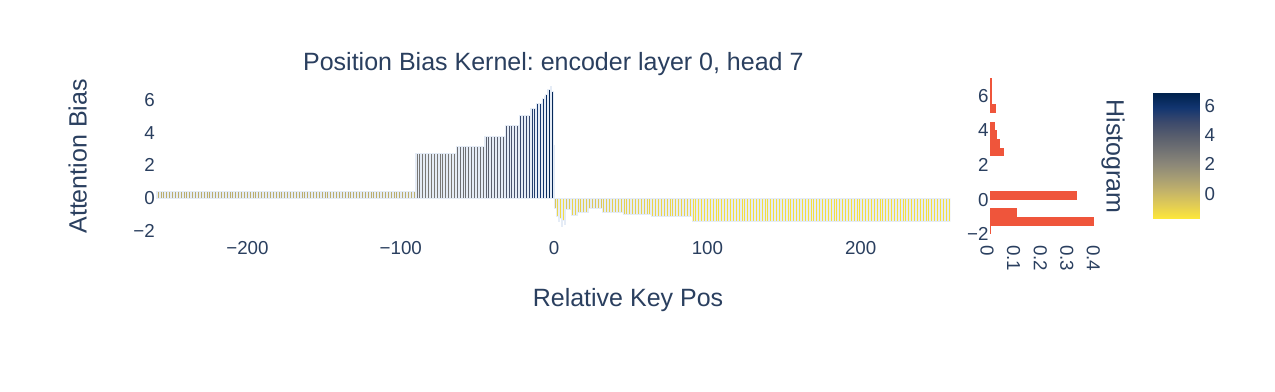}
      \end{subfigure}
      \hfill
      \begin{subfigure}{0.475\textwidth}
        \includegraphics[width=\linewidth]{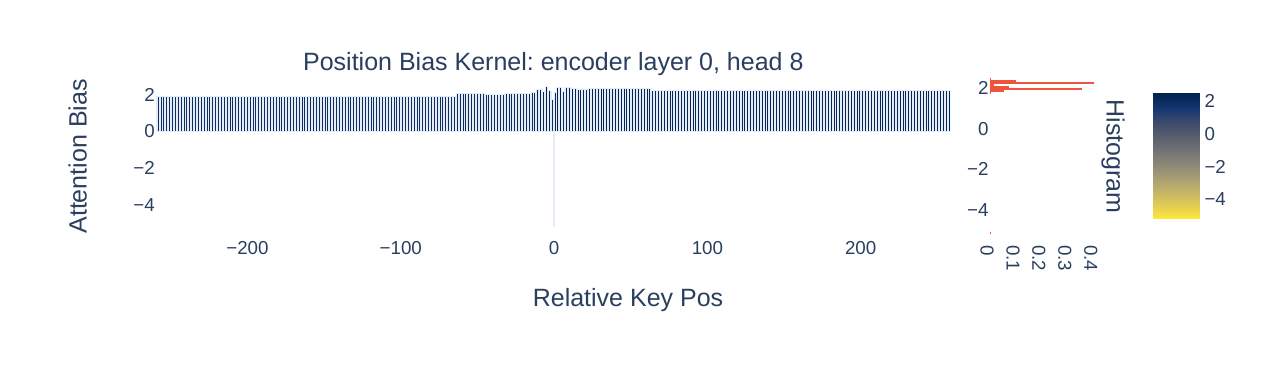}
      \end{subfigure}
      \begin{subfigure}{0.475\textwidth}
        \includegraphics[width=\linewidth]{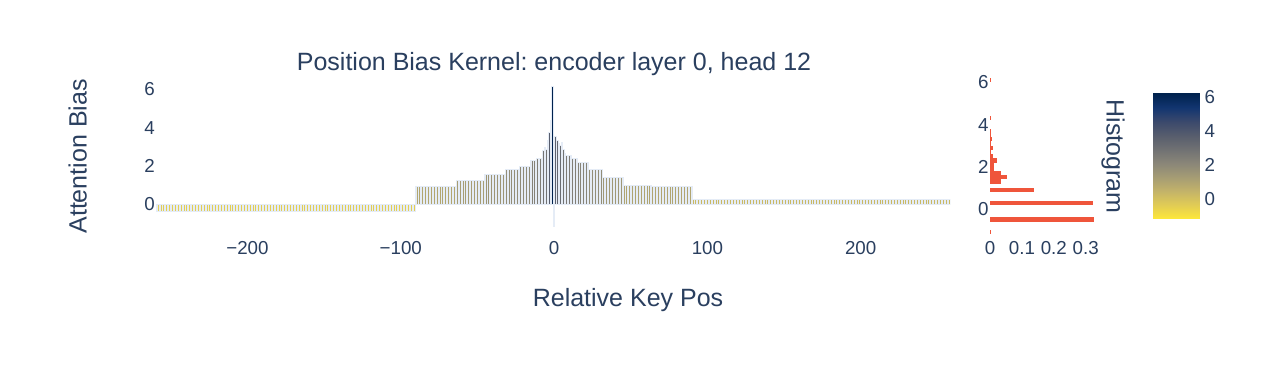}
      \end{subfigure}
      \hfill
      \begin{subfigure}{0.475\textwidth}
        \includegraphics[width=\linewidth]{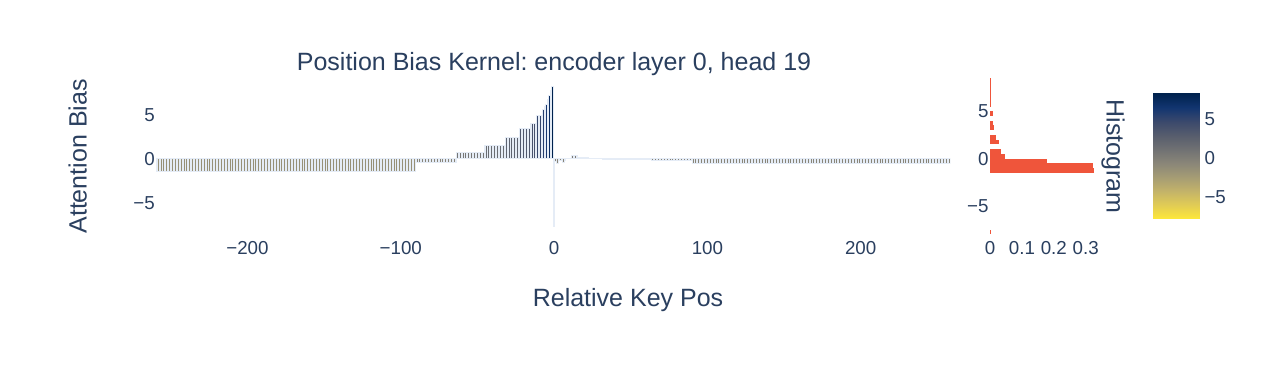}
      \end{subfigure}
      \begin{subfigure}{0.475\textwidth}
        \includegraphics[width=\linewidth]{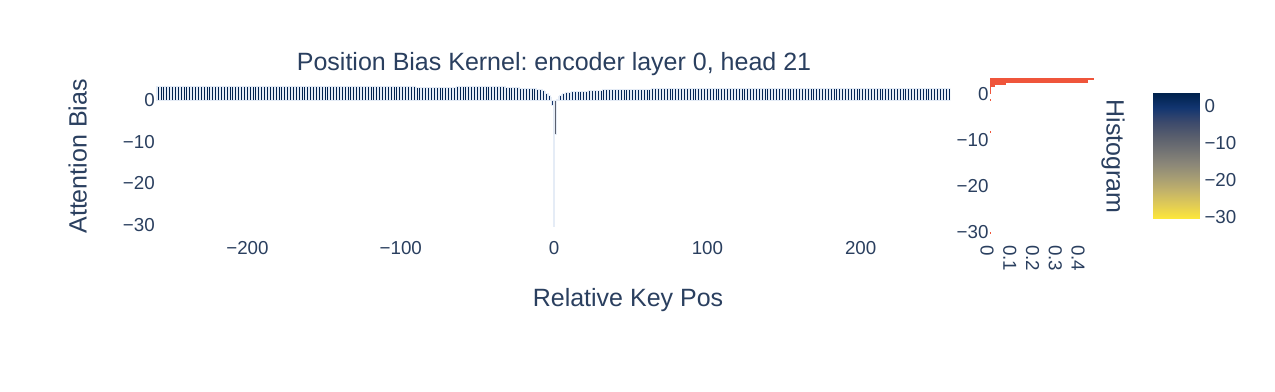}
      \end{subfigure}
      \hfill
      \begin{subfigure}{0.475\textwidth}
        \includegraphics[width=\linewidth]{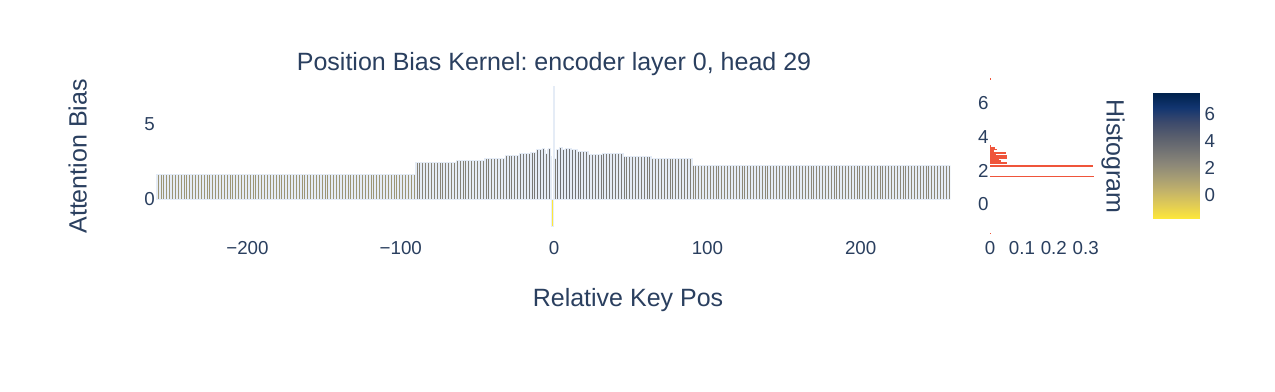}
      \end{subfigure}
  \end{center}
\caption{Some position kernels of a pretrained Flan-T5-XL encoder.}
\label{fig-pos-kernel}
\end{figure}

\begin{figure}[htb!]
\begin{center}
  \centering
      \begin{subfigure}{0.475\textwidth}
        \includegraphics[width=\linewidth]{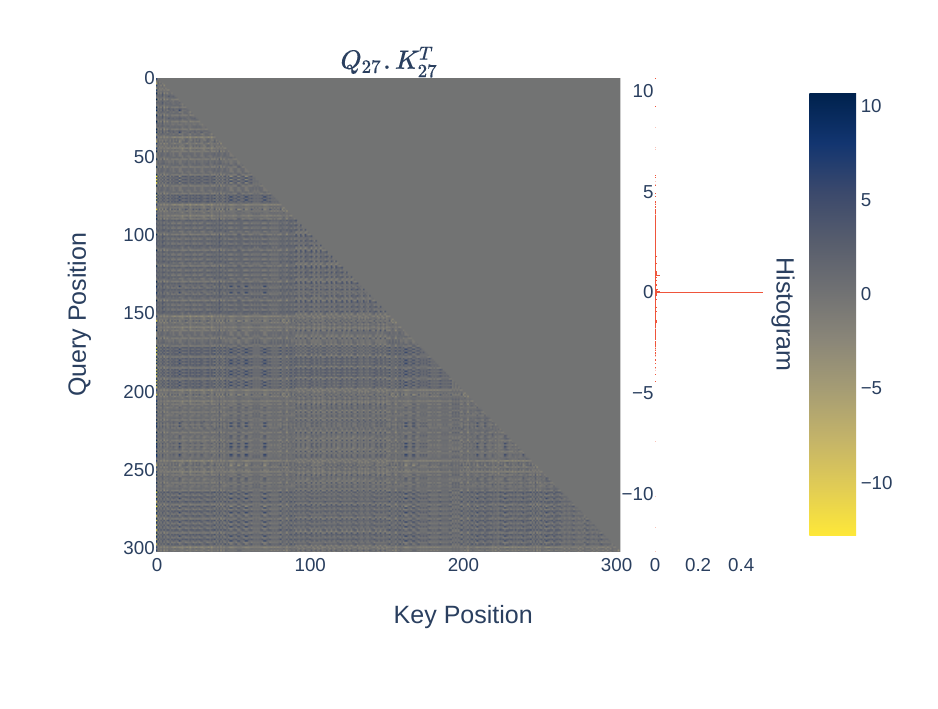}
        \caption{Attention Scores: Inner product of query and key vectors without position bias}
        \label{fig-pos-dec-a}
      \end{subfigure}
      \hfill
      \begin{subfigure}{0.475\textwidth}
        \includegraphics[width=\linewidth]{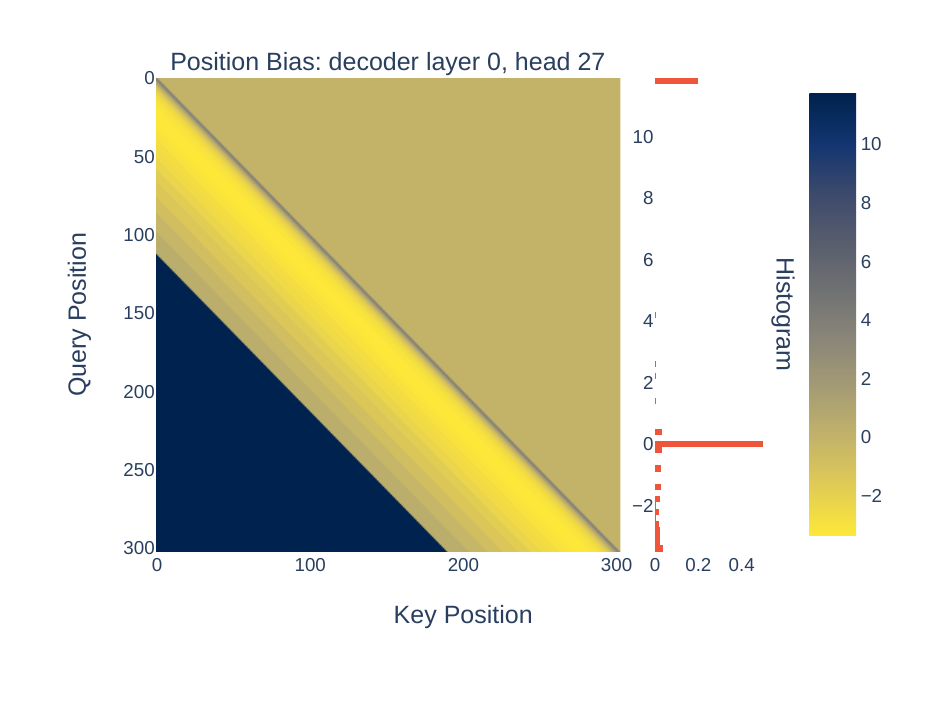}
        \caption{Position biases with positive self bias (2.12). Position kernel is in Fig. \ref{fig-pos-kernel-decoder}}
        \label{fig-pos-dec-b}
      \end{subfigure}
      \begin{subfigure}{0.475\textwidth}
        \includegraphics[width=\linewidth]{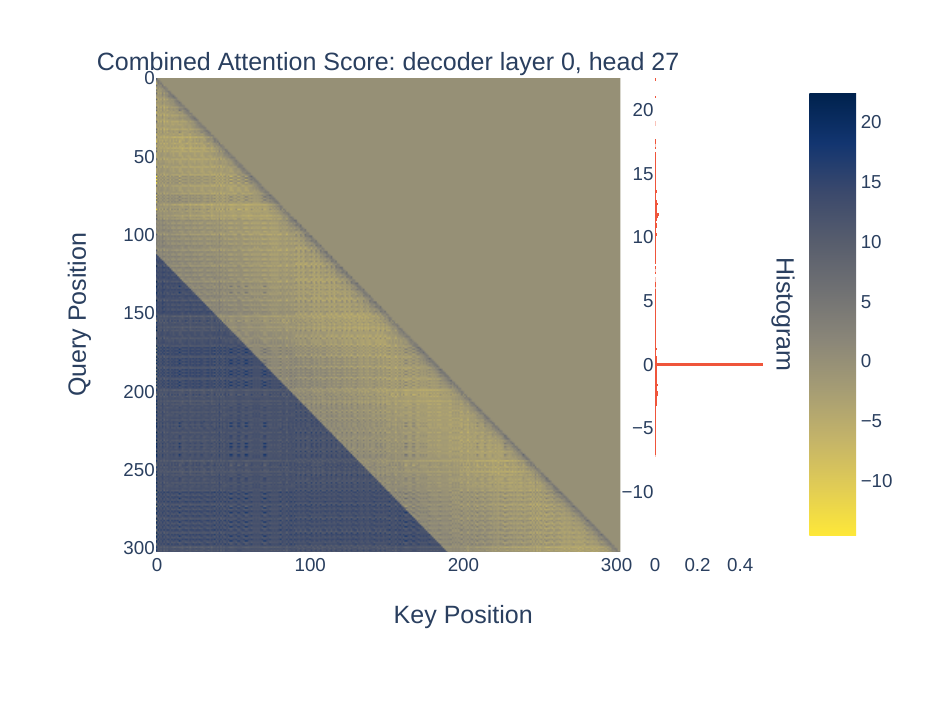}
        \caption{Attention weights before softmax = attention scores + position biases. Position bias significantly influences attention scores.}
        \label{fig-pos-dec-c}
      \end{subfigure}
      \hfill
      \begin{subfigure}{0.475\textwidth}
        \includegraphics[width=\linewidth]{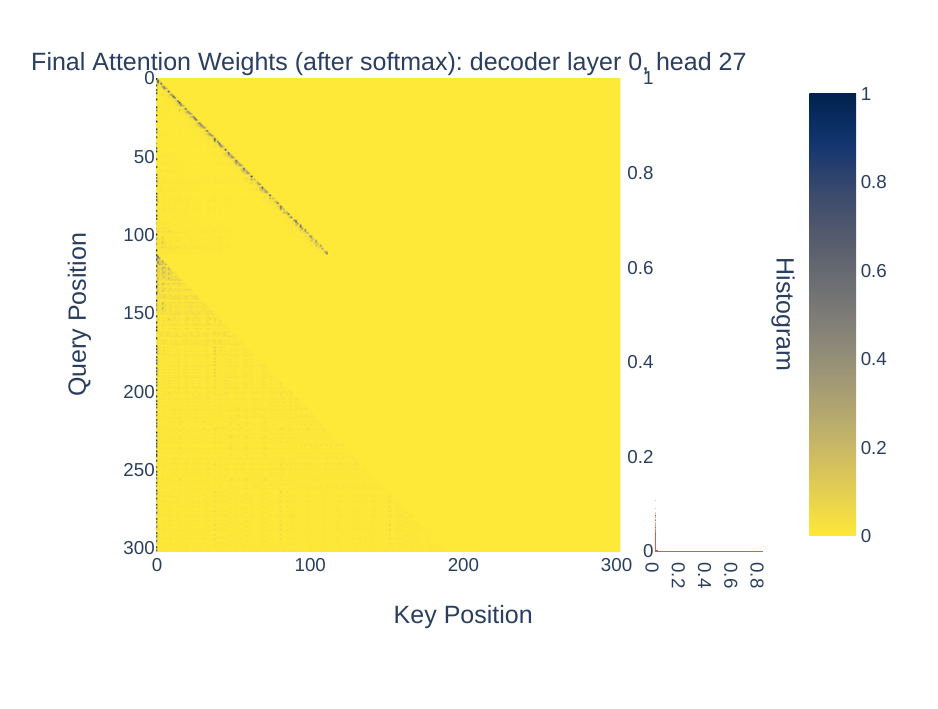}
        \caption{Attention weights (after softmax). Self bias (dark diagonal line) quickly dissipates as soon as the impact of foreign context imposed by position bias (bottom left dark triangle in Fig.
        \ref{fig-pos-dec-c}) kicks in.}
        \label{fig-pos-dec-d}
      \end{subfigure}
  \end{center}
\caption{ Pretrained Flan-T5-XL decoder attention layer activation maps.}
\label{fig-pos-dec}
\end{figure}

\begin{figure}[htb!]
\begin{center}
  \centering
      \begin{subfigure}{0.475\textwidth}
        \includegraphics[width=\linewidth]{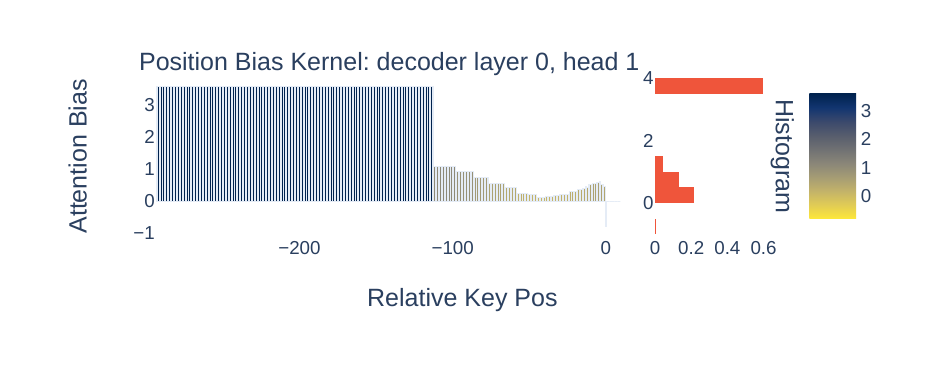}
      \end{subfigure}
      \hfill
      \begin{subfigure}{0.475\textwidth}
        \includegraphics[width=\linewidth]{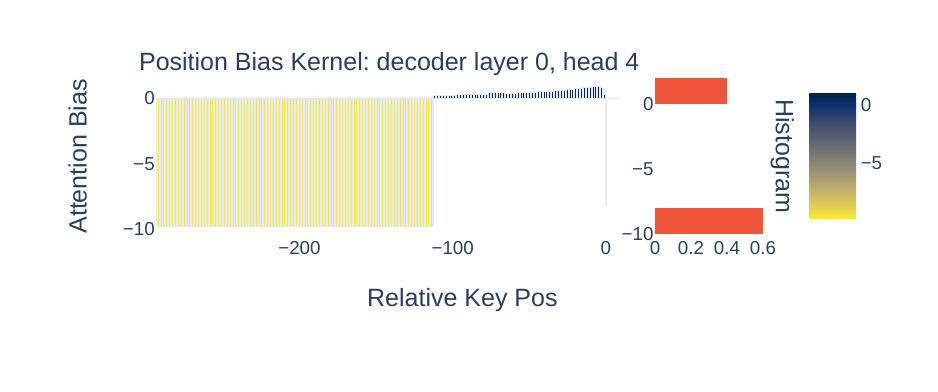}
      \end{subfigure}
      \begin{subfigure}{0.475\textwidth}
        \includegraphics[width=\linewidth]{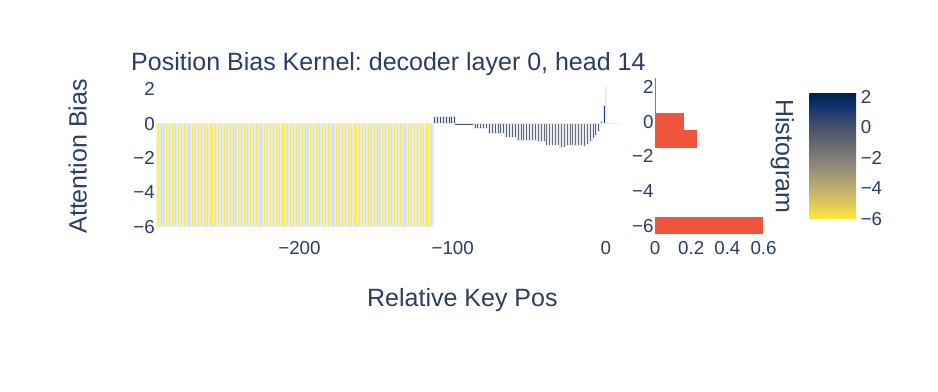}
      \end{subfigure}
      \hfill
      \begin{subfigure}{0.475\textwidth}
        \includegraphics[width=\linewidth]{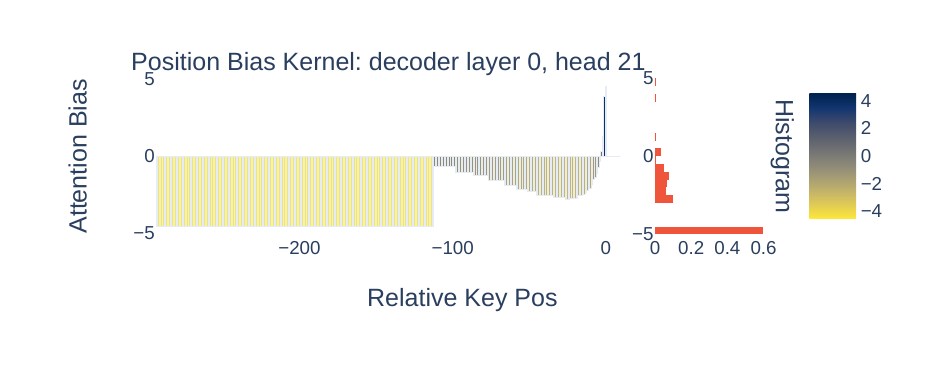}
      \end{subfigure}
      \begin{subfigure}{0.475\textwidth}
        \includegraphics[width=\linewidth]{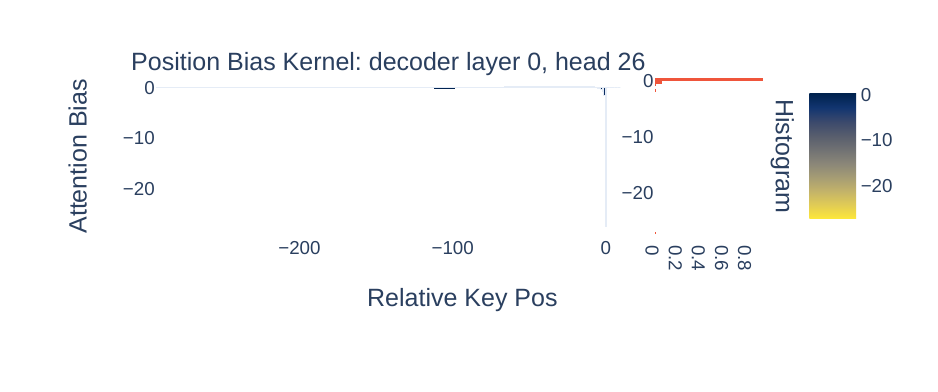}
      \end{subfigure}
      \hfill
      \begin{subfigure}{0.475\textwidth}
        \includegraphics[width=\linewidth]{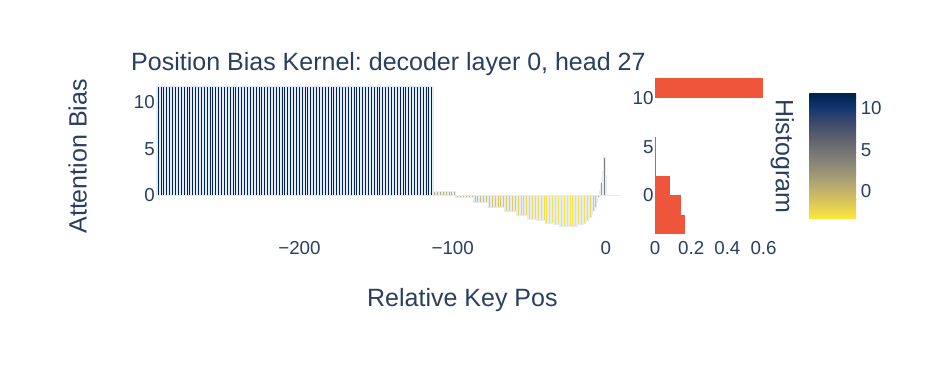}
      \end{subfigure}
  \end{center}
\caption{Some position kernels of a pretrained Flan-T5-XL decoder. In the bottom right is position kernel for head 27, corresponding to \ref{fig-pos-dec-b} and \ref{fig-pos-dec-c}. Position zero (self) is the right most position. Amplifies the far left positions and self. An intermediate band is suppressed.}
\label{fig-pos-kernel-decoder}
\end{figure}

\paragraph{Position Weighted Bag of Words}
Note that while relative positions influence the aggregation scores (i.e. attention weights) they are not directly stored in the aggregated vector. Position biases can only influence it by changing the composition of the cluster. In the T5 architecture, each head learns a distinct shape of relative position bias which influences the attention weight distribution over the sequence. We call the relative position biases the \emph{position kernel} since they are similar to a 1D long convolution kernel (see figs. 
\ref{fig-pos-kernel} and \ref{fig-pos-kernel-decoder} for examples). In T5-small and Flan-T5-XL models, we found that the magnitudes of the biases are large enough to significantly influence attention weights (Figs. \ref{fig-pos-enc} and \ref{fig-pos-dec}). After aggregation however, the sequence information is lost and hence the aggregation procedure can be thought of as a weighted summation of a bag of words. \underline{There is no other mechanism besides position bias}, to incorporate sequence information.

\paragraph{Similarity, Co-Occurrence and Angular Distance in Embedding Space}
\label{sec-similarity-associativity-proximity}
\emph{Cluster \& aggregate} is the only operation that involves interaction amongst vectors (also called \emph{data dependence} in literature). As analyzed, the effect of this operation is to aggregate features from across the context. Consequently, the composite decoding vector $\vd_t$ ultimately produced at the top of the stack, contains aggregated features from the context $\langle \vw_i \rangle_1^t$. At training time the denoising objective (e.g. MLM or LM) nudges it to come closer to the target token $\vw_{t+1}$. This effect is same as that explicitly sought by distributed word representation models such as Word2Vec \citep{Word2Vec} except that aggregation of context here is performed implicitly (by the attention layer) and that the aggregation weights are influenced by vector similarity and position. This should cause the features and consequently token embeddings that co-occur frequently to come closer together in embedding space thereby further increasing their similarity score and so on until either an equilibrium is reached\footnote{There's possibility of a feedback loop here.} or training stops. It follows therefore, that similarity is a proxy for co-occurrence / association. Notably, denoising training objectives will also cause context representations $d_{(\langle \vw_i \rangle)_1^t}$ to come closer to the target co-occurring words $\vw_{t+1}$. Also, as stated before, since the vector magnitudes are normalized due to layer-norm, similarity should imply angular proximity too. Therefore one would expect highly co-occurring tokens to cluster together angle-wise in $\sR^D$. We observed all of these effects empirically. For e.g., in figure \ref{fig-layer-heatmaps} we see closely associated tokens such as numbers and brackets exhibit high similarity scores. Broadly co-occurring tokens such as section separators of formatted samples exhibit low similarity across the board. The effects are even more pronounced when the vectors are normalized by layer-norm or L2-norm (i.e., cosine-similarity). More such patterns are shown in the appendix for pretrained flan-t5-large (encoder-decoder) (Fig. \ref{fig-flant5-heatmaps}) and falcon-7b (decoder only) \citep{refinedweb} (Fig. \ref{fig-falcon-heatmaps}) models. 
\begin{result}
\label{result-three-way-correspondence}[\textbf{Associative Organization of $\sS$}]
Thus a three way correspondence between Similarity (inner-product), association (co-occurrence) and angular proximity (cosine similarity) is established in embedding space. We call this the \textit{associative organization} of $\sS$ and as shown above it applies across token and context representations (composite vectors).
\end{result}
Notably the \textbf{role that attention plays} in co-locating co-occurring token embeddings \textbf{during training has been largely overlooked so far}.
\begin{result}
\label{result-att-conclusion}
    Putting results \ref{result-att-is-feature-aggregation} and \ref{result-three-way-correspondence} together, the attention layer implements position biased associative aggregation (of features) i.e., it ``gathers similar stuff from selected locations''.
\end{result}
\paragraph{The Role of Attention}
We hope to have demystified attention at this point. In particular, the model is not ``attending to relevant parts of the context'', because it has no mechanism to decide what's relevant. Attention's role is to provide an association bias to feature aggregation via similarity based soft-clustering\footnote{This could cause a feedback loop during training causing similar / co-occurring tokens to get more similar over time. But this doesn't happen in practice perhaps because of competing 'forces of association' between the tokens.}. Attention-free architectures \citep{Hochreiter1997LongSM, gu2022efficiently, fu2023hungry, poli2023hyena} on the other hand do organize $\sS$ associatively but purely based on the stochastic co-occurrence of features during training. They do not have the similarity bias that the Attention layer contributes and perhaps that's why they do not perform as well as Transformers \citep{vardasbi2023state}. However, these models are catching up quickly and it may turn out that Attention is not necessary after all.

\subsubsection{The Encoding Walk}
\label{sec-encoding-walk}

\begin{figure}[htbp!]
\begin{center}
  \centering
      \begin{subfigure}{0.32\textwidth}
        \includegraphics[width=\linewidth]{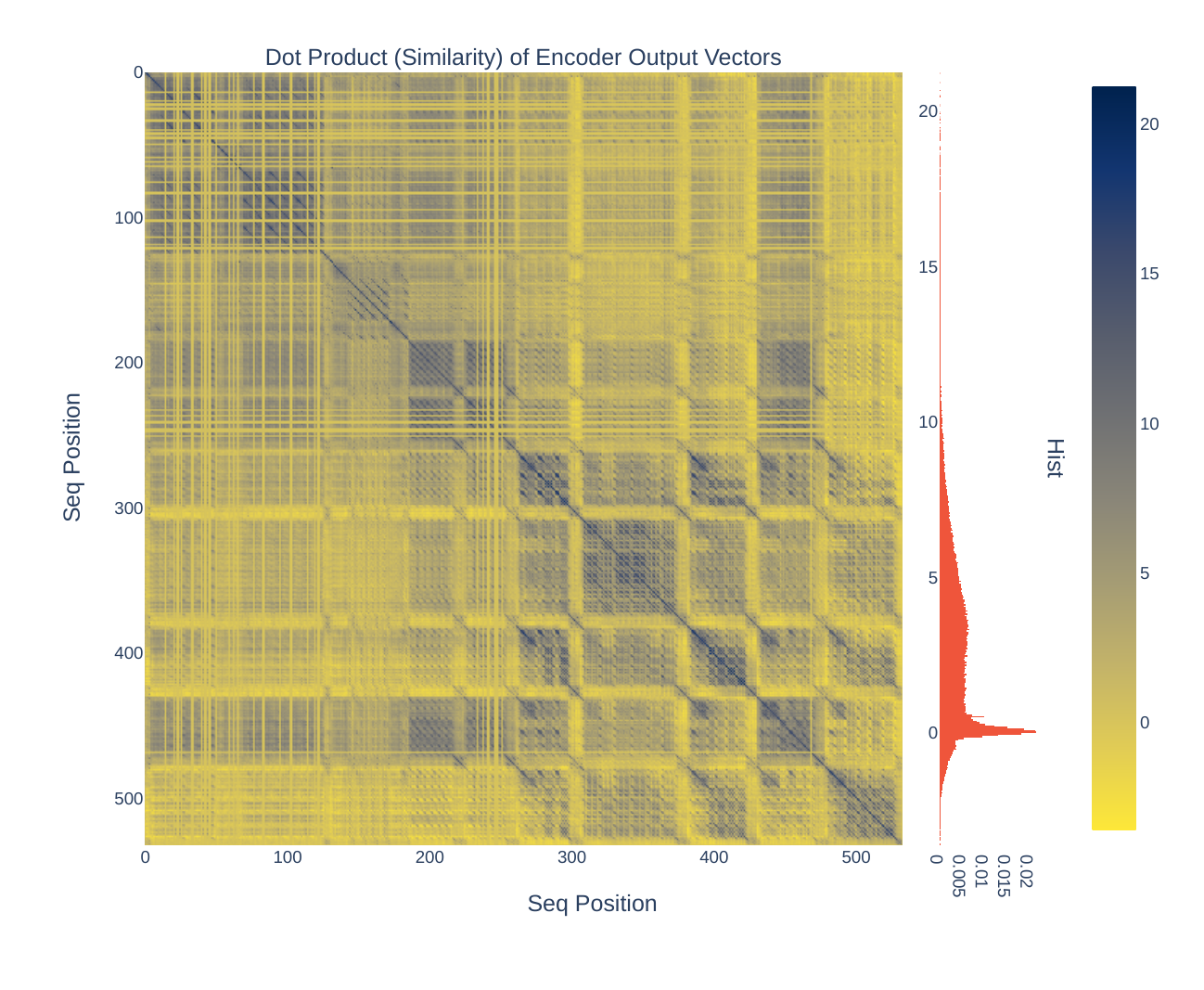}
        \caption{Output Layer}
        \label{fig-encoding-walk-d}
      \end{subfigure}
      \begin{subfigure}{0.32\textwidth}
        \includegraphics[width=\linewidth]{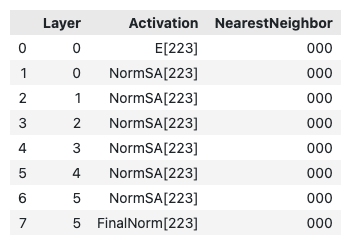}
        \caption{}
        \label{fig-encoding-walk-a}
      \end{subfigure}
      \begin{subfigure}{0.32\textwidth}
        \includegraphics[width=\linewidth]{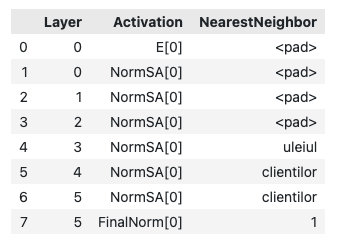}
        \caption{}
        \label{fig-encoding-walk-b}
      \end{subfigure}
  \end{center}
\caption{ (a) Continued from fig. \ref{fig-layer-heatmaps}. Similarity map of encoder output. The pattern gets fuzzy here because of aggregation of context into token vectors. However the broad pattern persists because the encoder vectors generally stay in the neighborhood of the original input. (b) Encoding walk of position 223 (input token `000') up a T5 encoder stack. The composed vector stays within the vicinity of the input token. E = token vector, NormSA = layer-norm'd layer input, FinalNorm = stack output. (c) Encoding walk of position 0 up the decoder stack: input token \textless pad\textgreater moves into neighborhood of the predicted token 1.}
\label{fig-encoding-walk}
\end{figure}

If you follow the path of a vector $\ve^l_t$ as it moves up a encoder or decoder stack (henceforth \emph{encoding vector}), it starts off as $\vw_t$ in layer 0 and, with every layer it traverses upwards, it ``attracts'' and aggregates ``similar context from selected locations'' (result \ref{result-att-conclusion}) until it finally emerges at the top as either $\vc_t$ or $\vd_t$. Note that with every such step the encoding vector itself gets ``pulled'' in the direction of the context that it just pulled, which in turn influences what it attracts the next time around. We call this traversal the \emph{encoding walk} (fig. \ref{fig-encoding-walk}).
\begin{result}
\label{result-encoding-walk}
The encoding walk is steered by 1) the features (of vectors) in the context, 2) ``the pull of association'' between them, 3) the feature-sequence-pattern, indirectly enforced via relative position-biases and 4) the static filters $\mW_{vo,h}$ that ultimately select the features to aggregate. Since $\mW_{vo,h}$ are fixed, they get marginalized w.r.t. dynamic behaviour, leaving only the first three factors to govern in-context learning. 
\end{result}
In the decoder each instance of encoding walk constitutes one step of the decoding walk. The walk ends at the final aggregated vector $\vd_t$ representing the context's feature sequence (results \ref{result-att-conclusion} and \ref{result-encoding-walk}). The next predicted token $\vw_{t+1}$ is picked from it's neighborhood. The LM training objective in turn forces this neighborhood to be populated with tokens that are correlated (in expectation) with $\vd_t$ (result \ref{result-three-way-correspondence}). 
\begin{result}
\label{result-encoding-walk-organizes-S}
Thus, the encoding walk, which is just $f_{\Sigma;\theta}$ unrolled, \textit{organizes} $\sS$ such that context feature-sequences (representations) lie in proximity of co-occurring next tokens. In other words the organization of $\sS$ attempts to capture the joint co-occurrence of feature-sequences presented during training\footnote{This property aligns very well with the pattern recognition task because a pattern can be defined as a sequence of features.}. The model generalizes at inference time by walking on higher probability paths through Embedding Space.
\end{result}
Sec. \ref{appendix-sec-self-bias} has a deeper analyses of the encoding walk.

\section{The Mechanics of Intelligence}
\label{sec-mechanics-of-intelligence}
\paragraph{In Context Learning?}
Our analyses shows that: 1) \emph{Cluster \& aggregate} is the only operation that involves interaction between positions and that happens via position-biased associative aggregation. Without it the Transformer would degrade to a point wise FF stack with no interaction between positions, completely incapable of recognizing patterns in the context\footnote{Unless the training objective was altered to cause such interaction.}. 2) Aggregation on the other hand, results in loss of explicit sequence information as explained in section \ref{sec-attention}. Hence, unlike a CNN which is imbued with spatial kernels, \emph{there is no mechanism in the transformer to even learn static sequential patterns, let alone dynamically as it may appear it does} when ``learning from context". Intelligent behaviour therefore occurs only by walking on high probability paths that resemble intelligent behaviour. However, since these paths are fixed after training no in-context learning or pattern recognition can take place. Context - which can be viewed as a type of momentum - merely places the model at a specific node in the DAG of (predetermined) paths which have predetermined probabilities. These are the mechanics of in-context learning.

Take for e.g. the patterns observed in figures \ref{fig-layer-heatmaps}, \ref{fig-encoding-walk-d}\footnote{More in the appendix (figs. \ref{fig-flant5-heatmaps} \& \ref{fig-falcon-heatmaps})} which clearly demarcate 9 sections of the few-shot input: one question, one rubric, six examples and one query sample. Each of the sections is of varying length, which may lead one to believe that the model `recognizes' the pattern and therefore is able to complete it. However a much simpler and mechanical explanation is that the section separators are broadly correlated with all tokens and therefore have weak similarity with them resulting in the light colored bands. Next, the answers themselves are made up of very similar tokens - brackets and numbers, which being highly correlated cause the appearance of dark square patterns.

\paragraph{It's always hallucinating}
Decoding walks are random walks regardless of whether the generated sequence may depict fact or fiction. Both facts and fiction are high probability paths through $\sS$.

\section{Related Work}
\citep{TTA_slides} introduced the embedding space centric framework that our work expands upon. Around the same time \citet{elhage2021mathematicalAnthropic} and \citet{olsson2022context} introduced a mathematical framework which included the notion of a ``residual stream'', a working memory / bandwidth from which attention heads read and write into, thereby implying the existence of an common vector space, same as ours.  However, they view it's purpose as a communication channel between layers and as temporary storage of information as opposed to our viewpoint wherein it plays a much more central and encompassing role: an abstract space that not only embodies information but also intelligence and skill as paths. In their formulation the job of attention heads is to move information between positions, whereas in our theory heads perform associative aggregation - the mechanics that give rise to the ``forces of association'' that shape encoding and decoding walks through Embedding Space. 
While their approach of isolating circuits is certainly a useful and necessary one, our work takes a more system-level approach attempting to answer the same questions via the encoding and decoding walk analyses. 
\cite{geva2022transformer, dar2022analyzing} espouse a vocabulary centric viewpoint. They view the token embedding vectors as an overcomplete basis for representing activation and weights matrices and apply it for steering inference and transferring knowledge across models. Our framework on the other hand just reuses the $d_{model}$ dimensions of the token embeddings as the basis of the vector space and our objective is to explain intelligent behavior in general. \cite{dai2023gpt} and \cite{vonoswald2023transformers} propose that self-attention equations when simplified to linear transforms only, under autoregressive decoding, can be reinterpreted as performing gradient descent across sequence positions. While this is a great orthogonal viewpoint - albeit it applies only to linear attention - it is still unclear what loss function is being optimized here. Also, training of token vectors $\sV$ is absent from this picture. While the approach of carefully studying the mathematics of small linear-models is certainly a useful and necessary one, we chose to study and experiment on real-world models - about 15 - ranging from 70 million to 13 billion parameters in size.

\section{Concept Space Theory}
\label{sec-experiments}
Since the objective of this work is to bridge the underlying mechanics of sequence models with their high level semantic behaviors, we tested the prevalent semantic interpretation of embedding vectors.
In recent literature embeddings are implicitly viewed as embodying meaning. This notion is especially prevalent in multimodal models \citep{https://doi.org/10.48550/arxiv.1612.09161, CLIP, lu2022unifiedio, lerf2023, reed2022gato, girdhar2023imagebind} where different modalities share a common embedding space. This is also the case with multilingual models or machine translation language models wherein the same embedding space is shared by different languages. Formalizing this intuition, we redefine \textit{embedding vector} to \textit{concept vector} $\vk$ and \textit{embedding space} to \textit{concept space}. Concepts serve as a building blocks of the semantic understanding exhibited by the model. Concept space is populated with concepts; coherent / incoherent, complete / incomplete and often themselves composed from other concepts in a part-whole hierarchy. Zero-shot instructions, few-shot prompts, chat history as well as the model's CoT reasoning, plans, observations and reflections are all (composite) concepts. Composite concepts are first class citizens of the \textit{concept space} alongside token-vectors.
\subsection{Experiments}
We test the following implied properties of this theory:
\begin{property}
    \label{property-distributed-composition}
    Like text documents, concepts may be composed from other sub-concepts in a pyramidal part-whole hierarchy. In other words,  composition (sec \ref{sec-concept-composition}) is additive and distributive.
    Therefore it should be possible to predict tokens by replacing the input context's token sequence with a sequence of (pre-aggregated) concepts each representing a segment of the context i.e., 
    $f_\Sigma\left({\langle \vw_i \rangle}_{T_0}^{T_n}\right) \; = \; f_\Sigma \left( {\left< \vk_{\seqw_{T_i}^{T_{i+1}}} \right>}_0^{n-1} \right)$ where $\vk_{\seqw_{T_i}^{T_j}}$ is the concept vector (embedding) representing the segment $\seqw_{T_i}^{T_j}$.\footnote{In practice property \ref{property-distributed-composition} can be applied to 1) compact long contexts by replacing them with preaggregated concepts and 2) cache frequently used concepts.}
\end{property}
\begin{property}
    \label{property-concept-similarity}
    Since composite concepts are first-class citizens of embedding space, they should organize associatively like tokens (sec. \ref{sec-similarity-associativity-proximity}).
    Therefore, it should be possible to perform a decoding walk over concepts i.e., predict $\vk_{\seqw_{T_n}^{T_{n+1}}}$ given ${\left \langle \vk_{\seqw_{T_i}^{T_{i+1}}} \right \rangle}_0^{n-1}$.\footnote{In practice, property \ref{property-concept-similarity} can be used to semantically evaluate the output of an LLM.
}
\end{property}
\paragraph{Hierarchical Concept Composition}
Intuitively one would think that $f_{\Sigma;\theta}$ should compose $\vk_\seqw$. However, that would be incorrect because $f_{\Sigma;\theta}$ is trained to map to individual tokens not aggregate concepts.
Instead, we employ a pyramidal aggregation scheme - $f_\oplus: \sR^{n\times D} \to \sR^D$ - wherein we encode a segment into a sequence of composite vectors $\left< \ve_t \right>$ and then aggregate them into a single segment-level embedding vector. This is optionally followed by aggregation of segment vectors within an example and then the optional aggregation of all example vectors. The argument here is that a semantically sound pyramidal partitioning i.e. sentences $\to$ sections $\to$ examples $\to$ full-context, should yield the best representation from a model that understands semantics of language and it's part-whole hierarchy. For encoder-decoder models, $\ve_t = \vc_t$ and for decoder-only models, $\ve_t = \vd_t$. The segment aggregation function for encoder-decoder models was (element-wise) \textit{mean}, while for decoder-only models we used a weighted mean (denoted \emph{w1mean}) where the $i^{th}$ vector in a sequence of $N$ vectors is weighted by $i/N$. Segment and example vectors were always aggregated via \emph{mean}. We shall call this the \textit{standard concept composition scheme}. For select models, we also tested several variations of this scheme, as explained later.

We tested properties 1 and 2 on 14 pretrained models on the Hellaswag benchmark \citep{zellers2019hellaswag}. The Hellaswag task is to choose one of 4 completion sentences given a few-shot context comprising of 0 or more example passages followed by a query passage. We tested both zero and 5-shot versions of this test with the following segmentation schemes: 1) \textit{segment\_sentences}: each sentence is considered a segment, 2) \textit{segment\_each\_example}: each example is segmented into its context and completion parts, 3) \textit{concat\_each\_example}: each zero/few-shot example is considered one segment and 4) \textit{concat\_all\_examples}: all examples of the context are concatenated together to form exactly one segment.

\subsubsection{Test 1: Token Generation Conditioned on Pre-aggregated Concepts}
To test property 1, we compute segment vectors as described above, yielding a sequence of vectors representing the context.  Next (without any further aggregation of the context), we concatenate token vectors of a completion sentence (one of the 4 choices) to this sequence and the resulting vector-sequence is fed to the model as input. Joint probability of the completion sentence is then obtained from the model's output probabilities and used to pick the winning choice. Except for the preaggregation of context segments, this is the standard procedure used to evaluate language models against Hellaswag. We modified the Language Evaluation Harness \citep{eval-harness} for this purpose. Our code is available as open source\footnote{Source code: \href{https://github.com/Turnitin-AI-Research/lm_eval_harness}{Experiments}, \href{https://github.com/Turnitin-AI-Research/tta-paper}{Open-source model visualizations}}.

\begin{figure}[htbp]
\begin{center}
  \centering
      \begin{subfigure}{.85\textwidth}
        \includegraphics[width=\linewidth]{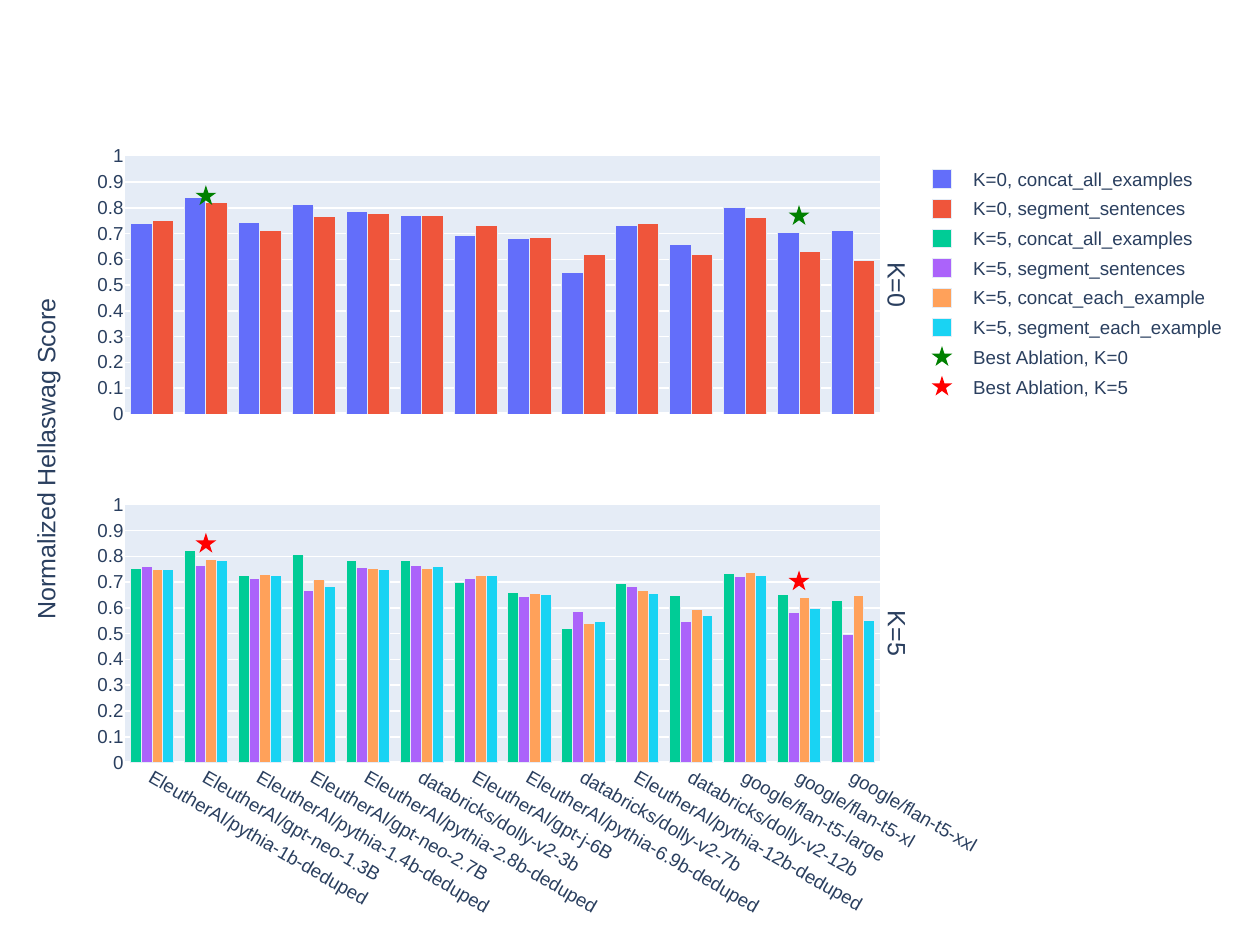}
        \caption{Pyramidal Generation: Test 1}
        \label{fig-dist-gen}
      \end{subfigure}
      \begin{subfigure}{.85\textwidth}
        \includegraphics[width=\linewidth]{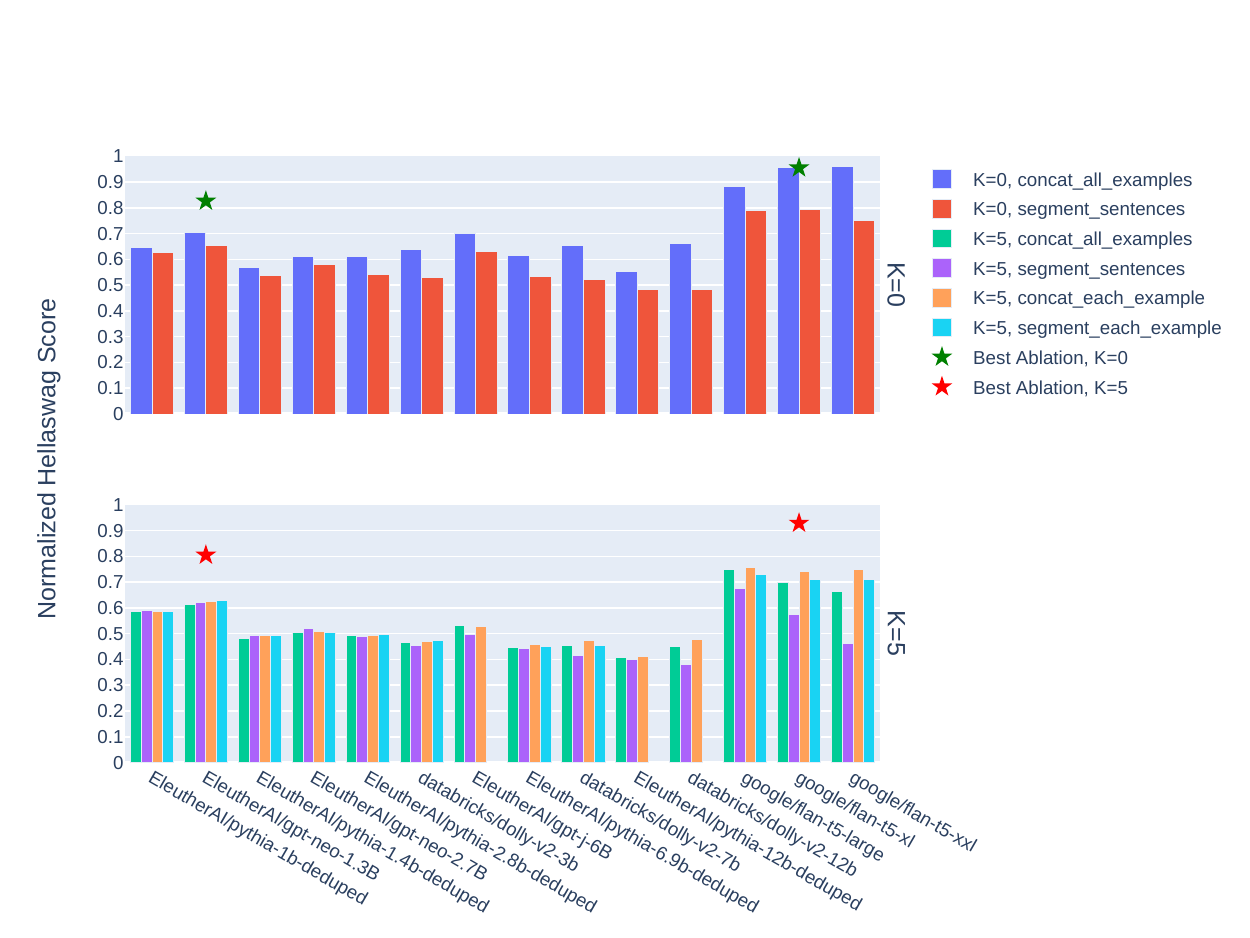}
        \caption{Concept Similarity: Test 2}
        \label{fig-dist-sim}
      \end{subfigure}
  \end{center}
\caption{Test 1\& 2 results: a) Pyramidal generation and b) Concept similarity test. }
\label{fig-dist-enc}
\end{figure}

Figure \ref{fig-dist-gen} shows the scores of 11 decoder-only and 3 encoder-decoder models. All of these models were obtained from the Huggingface transformers library \citep{wolf2020huggingfaces} . The scores shown are normalized by the model's baseline hellaswag score computed as usual with the full token-sequence as input. If property 1 holds and if the composition scheme was correct, then we expect this method to do as well or better than the baseline i.e. a normalized score $>= 1.0$. However as we see in Fig. \ref{fig-dist-gen}, none of the models achieve the baseline. 

This could mean that either the hypotheses are not true, are partially true and/or the standard composition scheme was incorrect. For e.g. one could argue that the top-layer of the stack is perhaps not the best place to pick token embeddings for all models - that perhaps it should be an intermediate layer for decoder-only models because since it has only one Transformer stack, that may have partitioned itself into an encoder stack followed by a decoder, thereby mimicking an encoder-decoder model. Or that perhaps the $f_\oplus$ should be machine-learnt. Therefore we tried several versions of the segmentation and aggregation functions and attempted to find the best combination i.e., the best concept composition scheme. However, the best performing scheme turned out to be unique per model, much like a machine-learnt function would have. Therefore we consider it a proxy for the ideal machine-learnt scheme and consequently rely on that score as a true test score. A few top performing schemes for gpt-neo-1.3B (decoder-only) and flan-t5-xl (encoder-decoder) models are reported in section \ref{sec-appendix-ablations} and all other results are available in our code repository (about 1200). Scores of the top performing schemes are shown with a $\star$ marker in Fig. \ref{fig-dist-gen}.

Decoder-only models generally performed better than encoder-decoder models of the same size in this test, the best scores being 83.9\% and 82.3\% for zero and 5-shot respectively for gpt-neo-1.3b. Its best composition scheme scores (marked with a $\star$) were 84.5\% and 84.9\% respectively. Encoder-decoder model results are generally lower with the best value being around 80\% for zero-shot and 73.8\% for 5-shot for flan-t5-large. 

\subsubsection{Test 2: Concept Similarity}
Here we test property 2 of the concept space theory. We embed the Hellaswag context into a single vector using the pyramidal aggregation scheme described earlier and do the same for each choice sequence (considered as exactly one segment). Then we compare the context and completions via dot-product and pick the winning choice i.e., we predict $f_\oplus ({\langle \vw_i \rangle}_{T_{n}}^{T_{n+1}})$ given $f_\oplus ({\langle \vw_i \rangle}_{T_{1}}^{T_n} )$. This procedure is reminiscent of asymmetric semantic search \citep{reimers-2019-sentence-bert} commonly employed for retrieval augmented generation \citep{lewis2021retrievalaugmented}. In those cases however, the embedding models are fine-tuned or trained for embedding passages but we are using vanilla generative models instead. As with test 1, we also test various permutations of composition schemes for gpt-neo-1.3B and flan-t5-xl models amounting to a total of over 6000, all of which are available in our code repository. This includes schemes that produce a concept vector sequence instead of a single vector for the context. In such cases each vector of the sequence is compared with the choice vector and the best match is used for picking the choice.

Fig. \ref{fig-dist-sim} shows normalized Hellaswag scores of this test. One thing that stands out is that all three encoder-decoder models perform much better than decoder-only models, achieving up to 96\% score. This is perhaps since they have a separate encoder stack which we used to encode the sequences. Decoder-only models do much worse, with the best composition scheme on the decoder-only side (for gpt-neo-1.3b) being 82.6\% and 80.5\% for zero-shot and 5-shot respectively. This could be because lack of an encoder stack makes finding the ideal $f_\oplus$ harder. 

\subsubsection{Summary of Results}
Since the scores are all under 100\%, concept space theory is clearly not proven. If properties 1 \& 2 had held we should've seen 100+\% scores on both tests because our segmentation schemes were carefully aligned with the the underlying semantics. That said, scores close to 85\% for test 1 and 95+\% for test 2 despite a rudimentary aggregation scheme does keep the possibility open that perhaps Transformers at least partially learn to map complex concepts in embedding space. But until that's proven, Transformers do seem to primarily map sequence → token rather than sequence → sequence. Sequence $\rightarrow$ sequence mapping subsumes sequence $\rightarrow$ token mapping and is a harder task since it requires mapping all partial sequences as well.

\section{Conclusion}
Through analyses we've shown that generative Transformers learn behaviors in the form of abstract paths in Embedding Space. Training sets conditional probabilities of all possible paths and stochastic decoding samples from the subset of paths that contain the context as prefix (candidate paths). ``In-context learning'' is a misnomer since the context only determines the set of candidate paths. Well performing LLMs place higher probability on paths representing intelligent behaviors.

Sequence $\rightarrow$ token probabilities are modeled by folding i.e., self-mapping, the context sequence into a single representation vector and co-locating it (per the dot-product similarity metric) with highly co-occurring next tokens (and doing the opposite for rarely co-occurring ones) i.e. by associatively organizing $\sS$. Context sequences are folded into vectors by aggregating their tokens' features. The aggregation process - the encoding walk - is steered by association and position biases in Transformers whereas attention-free architectures employ only position bias. The only role of \textit{attention} is to contribute this association-bias. While the above describes associative mapping of sequence $\rightarrow$ token, we also found some evidence of associative mapping from concept $\rightarrow$ concept although the results are far from conclusive.

These analyses are equally applicable across modalities and other neural sequence models that have a residual stream e.g., generative RNNs \citep{graves2014generating} and the newer post transformer architectures \citep{gu2022efficiently, fu2023hungry, poli2023hyena, sun2023retentive}. Thereby we arrive at a generalized architecture that's based on two very simple primitives - filtering and aggregation. We hope that these insights will guide the design of better neural sequence models, spur new research at the intersection of embedding space geometries and cognitive science, and ultimately lead to a deeper understanding of artificial intelligence.

\paragraph{Future Work}
\textit{Organization} of $\sS$ as a mathematical concept needs a more mathematically rigorous study around how many and what kind of patterns, meta-patterns and meta-meta-patterns and so on can be packed into a vector space in the form of paths and the factors governing that. This will reveal an upper bound of how intelligent such machines can get. Related to the above, self-bias in the encoding-walk needs more study. Finally, to what extent intelligence can be represented as behaviours i.e abstract paths needs to be studied from a cognitive science point of view. This is a fundamental question about the nature of intelligence.


\bibliography{main}
\bibliographystyle{tmlr}

\appendix
\section{Appendix}

\subsection{Token Neighborhoods and Target Path}
In this section we define \textit{neighborhoods} for referencing later. Consider a scenario where the model predicts a sequence $\seqw_{T+1}^{T+P}$ of length $P$ given a context $\seqw_1^T$. Under greedy decoding, $\seqw_{T+1}^{T+P}$ consists of words closest to the points $\seqd_{T+1}^{T+P}$ on the decoding walk. Therefore if we wanted to produce a target sequence of words - say $\seqdotw_{t=T+1}^{T+P}$ - we would need to ensure that every predicted vector $\vd_{\seqw^t_1}$ fell within a neighborhood $\mathcal{N}_{\bm{\dot{w}}_i}$ of $\bm{\dot{w}}_i$ such that $\dot{\vw}_i$ was its closest token. We shall denote the sequence of these neighborhoods of $\seqw_{t_1}^{t_2}$ as $\mathcal{N}_{\seqw_{t_1}^{t_2}}$ and we shall say that the decoding walk \underline{lies within} $\mathcal{N}_{\seqw_{t_1}^{t_2}}$ iff $\vd_i \in \mathcal{N}_{\vw_i} ; \forall i \in [t_1,t_2]$. Finally, we define $\mathcal{N}_{\seqw_{T+1}^{T+P}}$ as the \emph{target path} corresponding to the target sequence $\seqw_{T+1}^{T+P}$.

\subsection{Impact of Position Bias on the Encoding Walk}
\label{appendix-sec-self-bias}
In this section we discuss the impact of position bias on the encoding walk

\paragraph{Negative Self Bias}
While we haven't yet performed an exhaustive survey of learnt position biases across all popular models, analyses of T5 and Flan-T5 models gives us confidence that position biases play a significant role in vector composition for the T5 and perhaps for other architecture flavors as well. In particular about 80\% of the position kernels (one per head) in the T5-Small and Flan-T5-XL encoders significantly attenuate the query's own attention weight (Fig. \ref{fig-pos-kernel}). In its decoder stack we found about 30\% self-attention heads that amplify far away locations (Fig. \ref{fig-pos-kernel-decoder}). In both these cases, context is aggregated primarily from foreign locations i.e. locations excluding the query itself, thereby suppressing its own membership in its cluster. This also happens in cross-attention where the aggregated context does not include the query at all. We call this phenomenon \emph{negative self-bias}. This amounts to a copy-in from foreign locations. In our framing however, instead of saying that context has been copied into the query location, we say that the query vector moves into the center (i.e., centroid) of the context cluster. Note that if this context was obtained entirely from foreign locations, then the centroid could be significantly far from the query. Negative self-bias goes counter to popular intuition that position bias must be greatest at position 0 (query position) and then decrease gradually from there. However, negative self-bias is necessary to allow the query vector to escape its ``gravitational pull of self association'' i.e. the degenerate scenario where the query token is predicted over and over because it is most similar to itself \citep{holtzman2020curious}. Without negative self-bias, the query vector would dominate its cluster since it is likely to have the highest inner-product with itself. The subsequent summation with the residual vector - an unfiltered version of itself - would further exacerbate this situation. 

\paragraph{Encoder vs Decoder Walks}
On the mechanical side of things, we observe empirically that in the encoder stack $\vc^L_t$ stays within vicinity of $\vw_t$ i.e, $\mathcal{N}_{\vw_{t}}$, which is expected since the target token is the input itself (Fig. \ref{fig-encoding-walk-a}). In the decoder stack where the output un-embedding matrix is tied to the input embedding matrix however, the output vector strays away from $\vw_t$ (Fig. \ref{fig-encoding-walk-b}) since its training objective is to predict the next token, not the input token. However, for a model like T5 which has both an encoder and a decoder, it is instructive to examine how the stacks behave differently. The behaviour of the encoder stack is understandable since the ``pull of self association'' of the primary vector $\vw_t$ is expected to be strong given that it is highly correlated with itself (because it's mutual angle with itself is zero) and that residual connections should further help keep it in it's own neighborhood $\mathcal{N}_{\vw_{t}}$. In the decoder stack however, $\vw_t$ must move out of it's own neighborhood $\mathcal{N}_{\vw_{t}}$ and into the ``foreign'' neighborhood $\mathcal{N}_{\vw_{t+1}}$. In order to do this it would need to overcome the strong ``self-association pull'' by aggregating sufficient amount of foreign context as it moves up through the layers. The only way that can happen is if the cumulative mass of attention placed on foreign locations i.e. on vectors other than the query, was much higher than in the encoder stack. This could happen via the context pulled in from the encoder by cross-attention in addition to negative self bias exerted by position biases. We see this happening in Fig. \ref{fig-pos-dec-c} where the self bias (dark diagonal) dissipates the moment the far left context kicks in. For decoder-only architectures\footnote{... where the in/out embedding matrices are tied, which is the majority case} however, there is no cross-attention and therefore $\vw_t$ must overcome self-association pull purely through context aggregated by self-attention. 
Negative self-bias exerted by position biases as discussed in section \ref{appendix-sec-self-bias} is the only mechanism we have found that can accomplish this in decoder-only models. However, the GPT series of models only has global attention which, being inserted at the input layer would quickly dissipate after aggregation in layer 1. So, how negative self-bias is exerted in those models is a mystery to us. Perhaps an interplay between filters and the organization of $\sS$? Plus the fact that generative decoding is performed via. stochastic sampling \citep{holtzman2020curious} not greedy decoding (which is known to cause degenerate repetition), relaxes this constraint? We haven't done enough investigation into this topic to have the answers. Specifically, the relative impact of the following four factors across various model flavors needs further investigation: 1) vector similarity, 2) position biases 3) filters and 4) decoding method.


\subsection{Attention Layer Refactored}
\label{appendix-attention-head-math}
First we show that the attention layer output can be recast as a sum of outputs of individual heads i.e., each attention head works independently and in parallel with other heads. All vectors are represented as rows.
\begin{IEEEeqnarray*}{lCl}
    H   \quad&:&\quad \text{number of heads} \\
    d   \quad&:&\quad \text{vector dimension of heads } = D/H \\
    A_h \quad&:&\quad n \times m \text{ attention matrix of head h, rows sum to 1} \\
    V_h \quad&=&\quad m \times d \text{ value vectors of head } h \\
    \mW_{o} \quad&:&\quad D \times D \text{ output projection matrix } \\
    \mW_{o,h} \quad&:&\quad d \times D\;;\, h^{th} \text{ of } H \text{ row-wise slices of } \mW_o \text{ where }\\
    O   \quad&:&\quad n \times D \text{ attention layer output vectors} \\
        \quad&=&\quad \begin{bmatrix}
        A_1 V_1 & \cdots & A_H V_H
    \end{bmatrix} \mW_o\\
        \quad&=&\quad \begin{bmatrix}
        A_1 V_1 & \cdots & A_H V_H
    \end{bmatrix} \begin{bmatrix}
        \mW_{o,1} \\
        \cdots \\
        \mW_{o,H} 
        \end{bmatrix}\\
        \quad&=&\quad \sum_h A_h V_h \mW_{o,h} \\
    \text{defining } O_h \quad&=&\quad A_h V_h \mW_{o,h} \;;\, n \times d \text{ output of head } h \\
    \text{we get } O \quad&=&\quad \sum_h O_h
\end{IEEEeqnarray*}

We show next that each head can be viewed as an independent linear transform of input vectors $x_i$ without any interaction with other vectors i.e, filtering, followed by soft-clustering i.e., matrix multiplication with $A_h$.

\begin{IEEEeqnarray*}{lCl}
    \mX \quad&:&\quad \begin{bmatrix}
        x_1 \\
        x_2 \\
        \cdots \\
        x_n
    \end{bmatrix} \;;\, n \times D \text{ matrix of row vectors (inputs); } n = \text{ sequence length} \\
    \mW_{v,h} \quad&:&\quad D \times d \text{ value projection matrix of head } h \\
    V_h \quad&=&\quad \mX\mW_{v,h}\\
    O_h \quad&=&\quad A_h V_h \mW_{o,h} \\
      \quad&=&\quad A_h \mX \left(\mW_{v,h} \mW_{o,h} \right) \\
      \quad&=&\quad A_h \mX \left(\mW_{vo,h} \right) \;;\, \mW_{vo,h} = \mW_{v,h} \mW_{o,h} \in R^{D \times D} \\
      \quad&=&\quad A_h \begin{bmatrix}
        x_1 \mW_{vo,h} \\
        x_2 \mW_{vo,h} \\
        \cdots \\
        x_n \mW_{vo,h}
      \end{bmatrix} \\
      \quad&=&\quad A_h \; filter_{W_{vo,h}}\left(\mX\right) \\
\end{IEEEeqnarray*}

For completeness sake, derivations of $A_h$ are shown below.
\begin{IEEEeqnarray*}{lCl}
    \mW_{q,h} \quad&:&\quad D \times d \text{ query projection matrix of head } h \\
    \mW_{k,h} \quad&:&\quad D \times d \text{ key projection matrix of head } h \\
    Q_h \quad&=&\quad \mX\mW_{q,h} \;;\, n \times d \text{ query vectors of head } h \\
    K_h \quad&=&\quad \mX\mW_{k,h} \;;\, m \times d \text{ key vectors of head } h \\
    B   \quad&:&\quad n \times m \text{ position bias; absolute or relative, including ROPE} \\
    A_h \quad&:&\quad n \times m \text{ attention matrix for head h, rows sum to 1} \\
        \quad&=&\quad softmax\left(Q_h K_h^T + B\right) \;;\, \text{softmax over columns i.e., } dim=1 \\
        \quad&=&\quad softmax\left(\mX\mW_{q,h}\mW_{k,h}^T\mX^T + B\right) \\
        \quad&=&\quad softmax\left(\mX\mW_{qk,h}\mX^T + B\right)\;;\, \mW_{qk,h} = \mW_{q,h}\mW_{k,h}^T \\
        \quad&=&\quad softmax\left(filter_{W_{qk,h}}(\mX)\mX^T + B\right)
\end{IEEEeqnarray*}

\subsection{Few Shot Prompt Example}
\label{sec-appendix-fewshot-example}
Below is a listing of the sample used to produce all the similarity maps shown in this paper. Its a formatted prompt containing few-shot examples on an answer grading task, starting with a question, then a scoring rubric, followed by 6 graded answers and finally a query answer whose grade the model is expected to generate. Section headers `Q:', `R:',  `A:', `G:' and `L:' demarcate the beginning of the question, rubric, answer, grade and label sections respectively.  The model is expected to generate the label of the query answer.
\begin{verbatim}
Grade answers

Q:
Give three pairs of pixel values (x, y) = [[?, ?, ?]] and (x+1, y) = [[?, ?, ?]] in the input image, for which this code **will** produce the correct result.

### Pair 1
(x, y) =
(x+1, y) =
### Pair 2
(x, y) =
(x+1, y) =
### Pair 3
(x, y) =
(x+1, y) =

R:
 1: All Correct - Average of first array must be >= second (5.0)
 2: Example 1 incorrect (0.0)
 3: Example 2 incorrect (0.0)
 4: Example 3 incorrect (0.0)
 5: Impossible pixel values, >1 (0.0)

A:
[5, 6, 7]
[2, 3, 4]
[4, 5, 6]
[2, 3, 4]
[7, 8, 9]
[3, 4, 5]

G: 00001
L: 000

A:
[3,4,5]
[2,3,4]
[4,5,6]
[3,4,5]
[5,6,7]
[4,5,6]

G: 00001
L: 000

A:
[2,2,2]
[1,1,1]
[2,2,2]
[0.2,0.2,0.2]
[2,2,2]
[0.3,0.3,0.3]

G: 10000
L: 1

A:
[.8, .8, .8]
[.3, .3, .3]
[.8, .8, .8]
[.8, .8, .8]
[.7, .7, .1]
[.3, .3, .3]

G: 10000
L: 1

A:
[2,2,2]
[1,1,1]
[0.6,0.6,0.6]
[0.3,0.3,0.3]
[0.7,0.7,0.7]
[0.4,0.4,0.4]

G: 10000
L: 1

A:
[2, 1, 1]
[1, 1, 1]
[24, 46, 61]
[2, 2, 2]
[22, 51, 2]
[21, 3, 1]

G: 00001
L: 000

A:
[2, 2, 2]
[1, 1, 1]
[0.6, 0.8, 0.9]
[0.2, 0.3, 0.4]
[0.8, 0.5, 0.6]
[0.3, 0.3, 0.4]

G:
L:
\end{verbatim}

\subsection{Additional Similarity Maps}
\label{appendix-similarity-maps}
\begin{figure}[htb!]
\begin{center}
  \centering
  \begin{subfigure}{0.475\textwidth}
    \includegraphics[width=\linewidth]{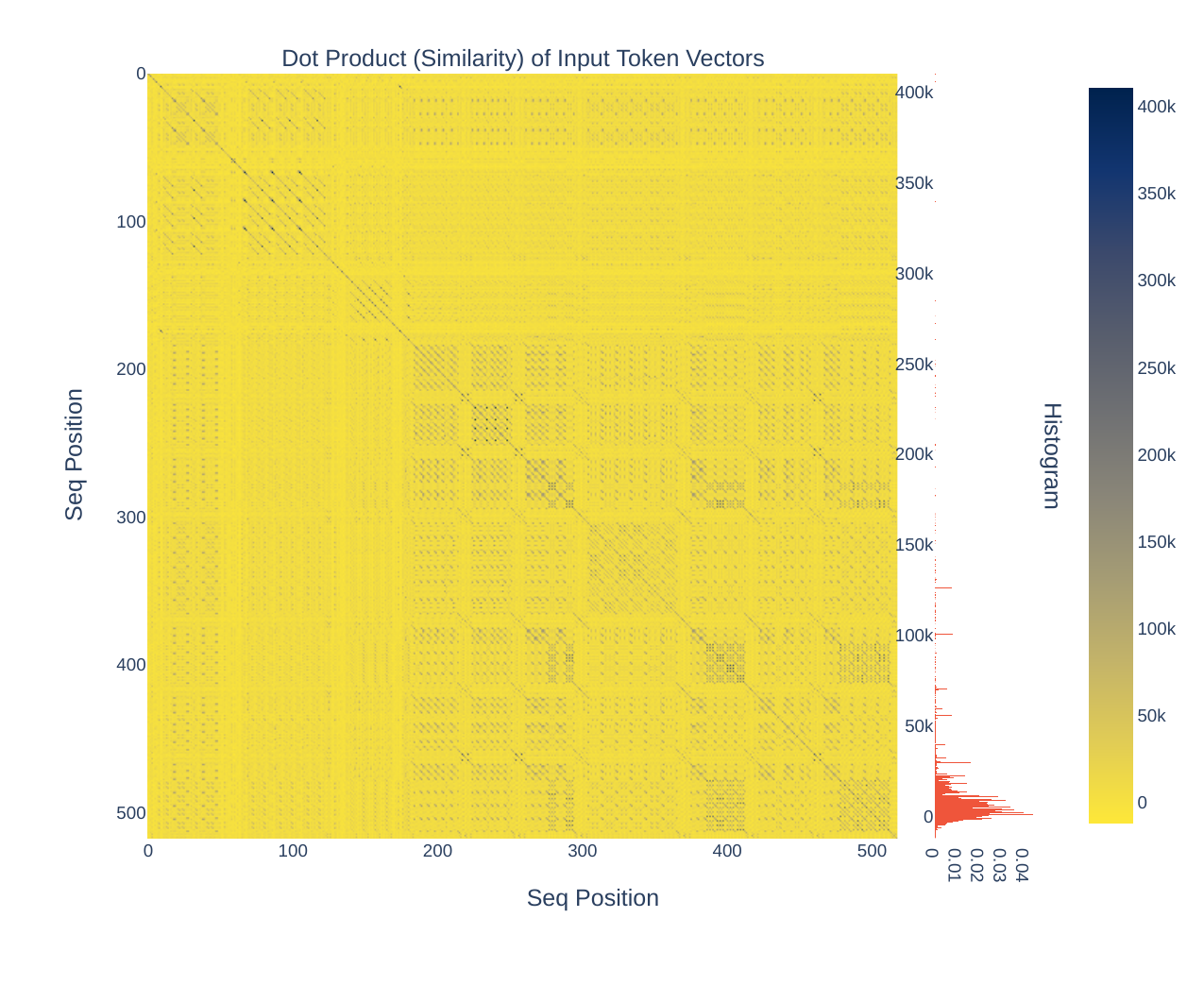}
    \caption{ Inner-product of encoder input embeddings}
    \label{fig-flant5-heatmaps-a}
  \end{subfigure}
  \hfill
  \begin{subfigure}{0.475\textwidth}
    \includegraphics[width=\linewidth]{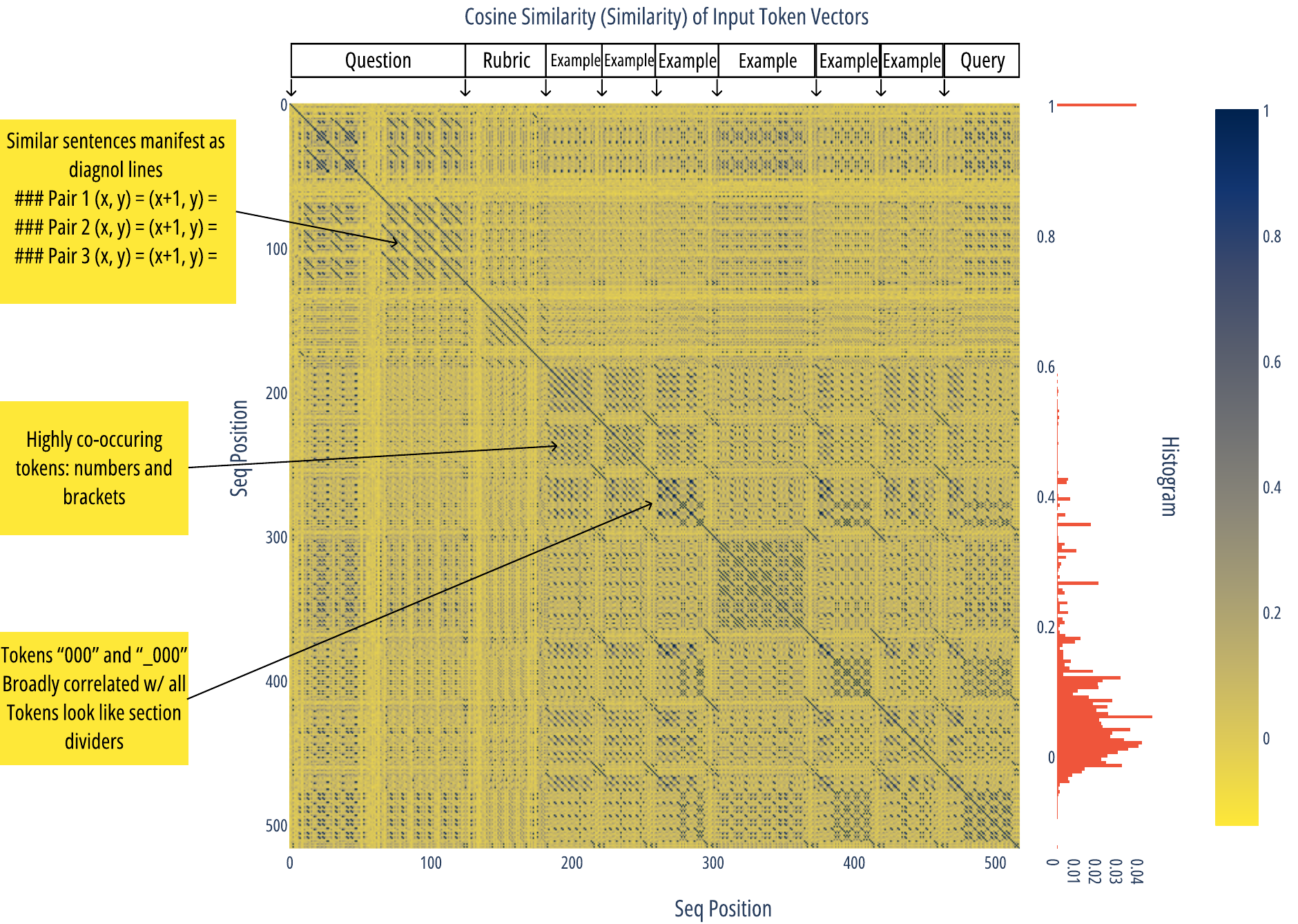}
    \caption{ Cosine similarity i.e., angular proximity of encoder input embeddings}
    \label{fig-flant5-heatmaps-b}
  \end{subfigure}
  \\
  \begin{subfigure}{0.475\textwidth}
    \includegraphics[width=\linewidth]{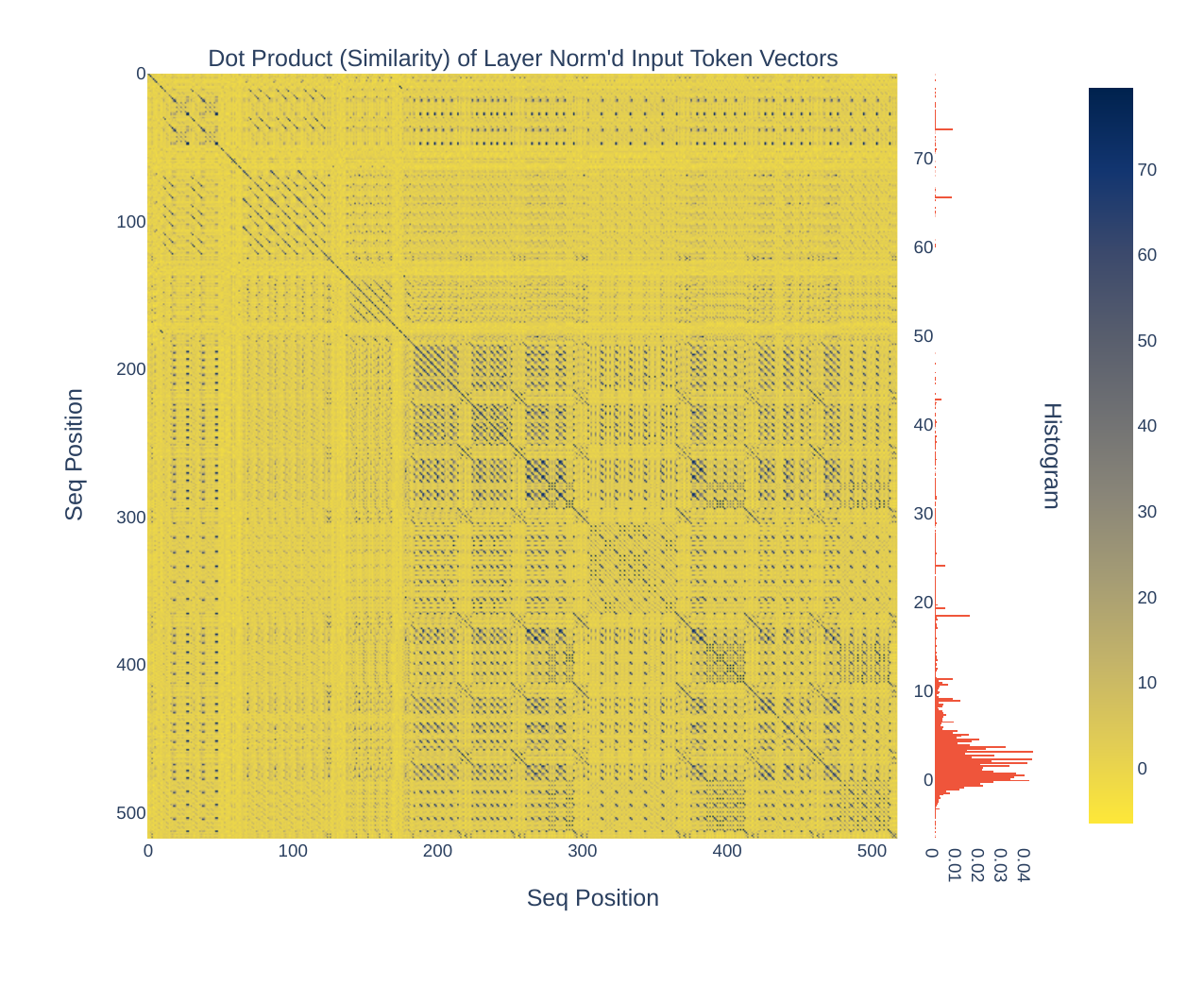}
    \caption{ Inner-product of tokens after input layer-norm of first encoder layer}
    \label{fig-flant5-heatmaps-c}
  \end{subfigure}
  \hfill
  \begin{subfigure}{0.475\textwidth}
    \includegraphics[width=\linewidth]{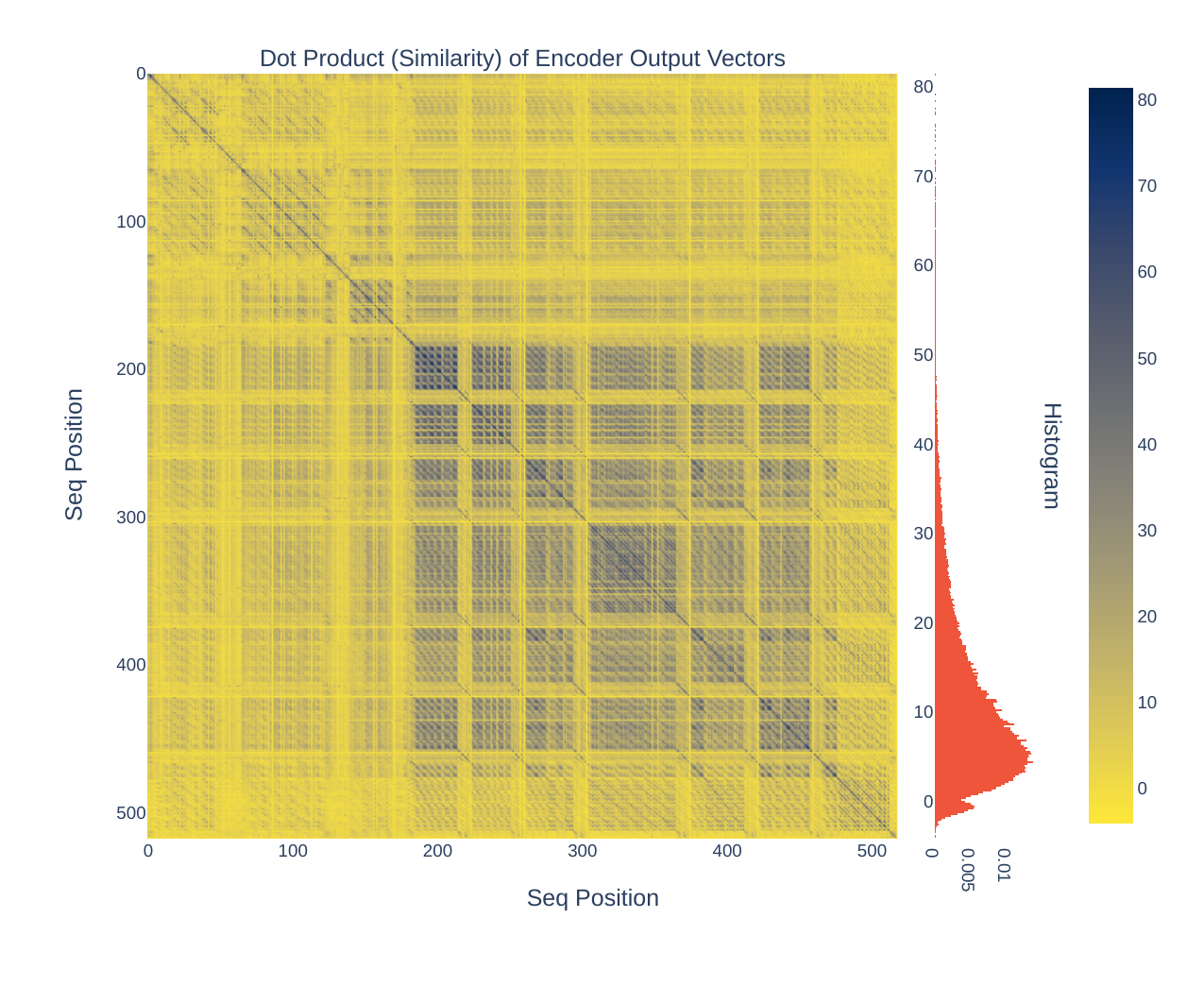}
    \caption{ Inner-product of encoder output vectors $\langle\vc_i\rangle$}
    \label{fig-flant5-heatmaps-d}
  \end{subfigure}
\end{center}
\caption{Vector similarity maps for the same example as in Fig. \ref{fig-layer-heatmaps}, but passed through a frozen google/flan-t5-xxl model. The structure of the document and similarity of associated tokens is still clearly visible despite the fact that the model is a generic pretrained model. The fact that the patterns are more visible by cosine-similarity indicates that associated tokens (such as brackets and numbers) cluster together in Embedding Space based on angular separation. Subfigure (d) shows a similar map of the encoder output. The patterns are fuzzier but still match up with the first layer indicating that encoded composite vectors still retain their original identity which is not surprising since it's trained to predict the input token.}
\label{fig-flant5-heatmaps}
\end{figure}

\begin{figure}[htb!]
\begin{center}
  \centering
  \begin{subfigure}{0.475\textwidth}
    \includegraphics[width=\linewidth]{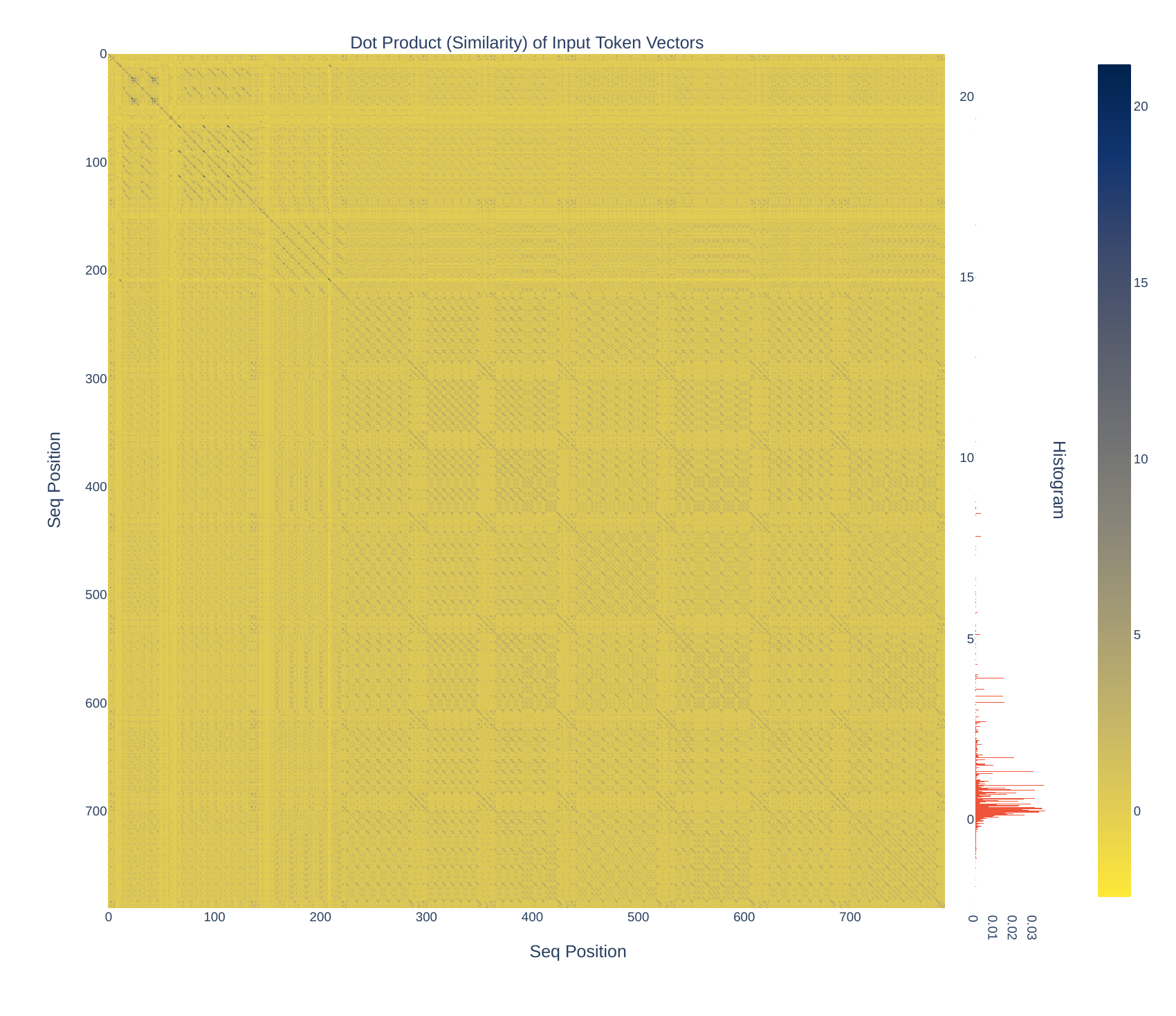}
    \caption{ Inner-product of input token embeddings}
    \label{fig-falcon-heatmaps-a}
  \end{subfigure}
  \hfill
  \begin{subfigure}{0.475\textwidth}
    \includegraphics[width=\linewidth]{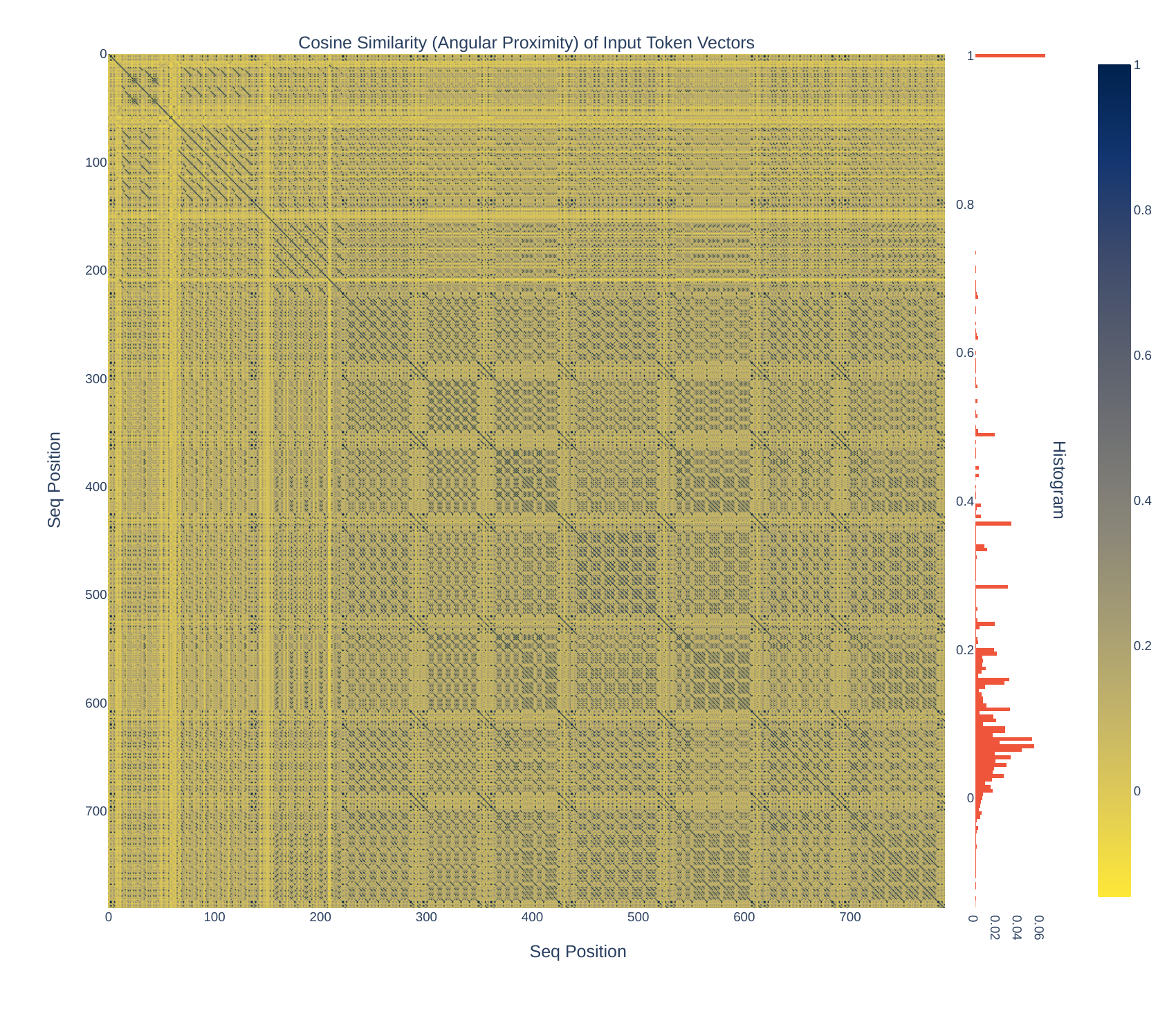}
    \caption{ Cosine similarity i.e., angular proximity of input token embeddings}
    \label{fig-falcon-heatmaps-b}
  \end{subfigure}
  \\
  \begin{subfigure}{0.475\textwidth}
    \includegraphics[width=\linewidth]{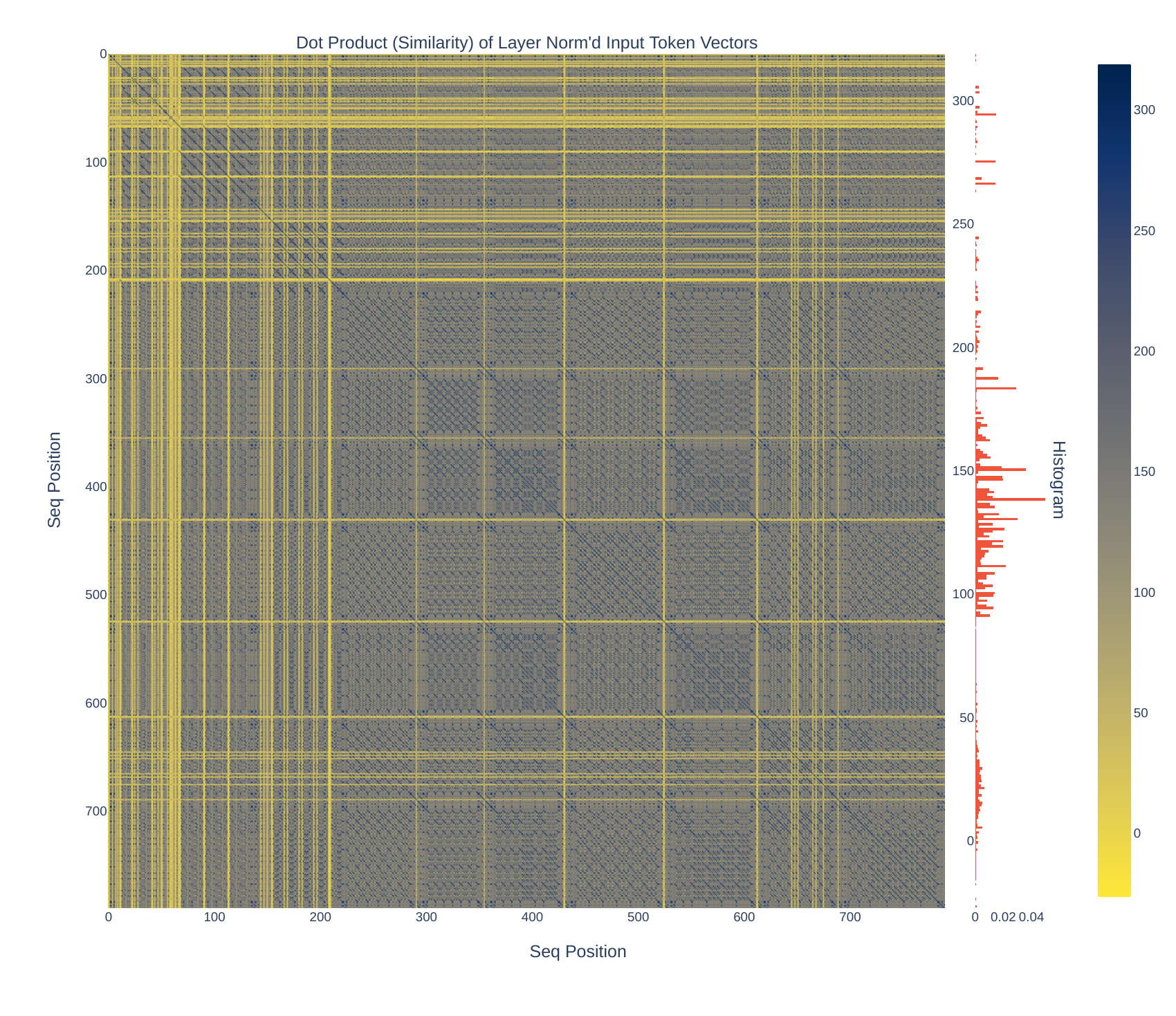}
    \caption{ Inner-product of vectors after input layer-norm of the first layer}
    \label{fig-falcon-heatmaps-c}
  \end{subfigure}
  \hfill
  \begin{subfigure}{0.475\textwidth}
    \includegraphics[width=\linewidth]{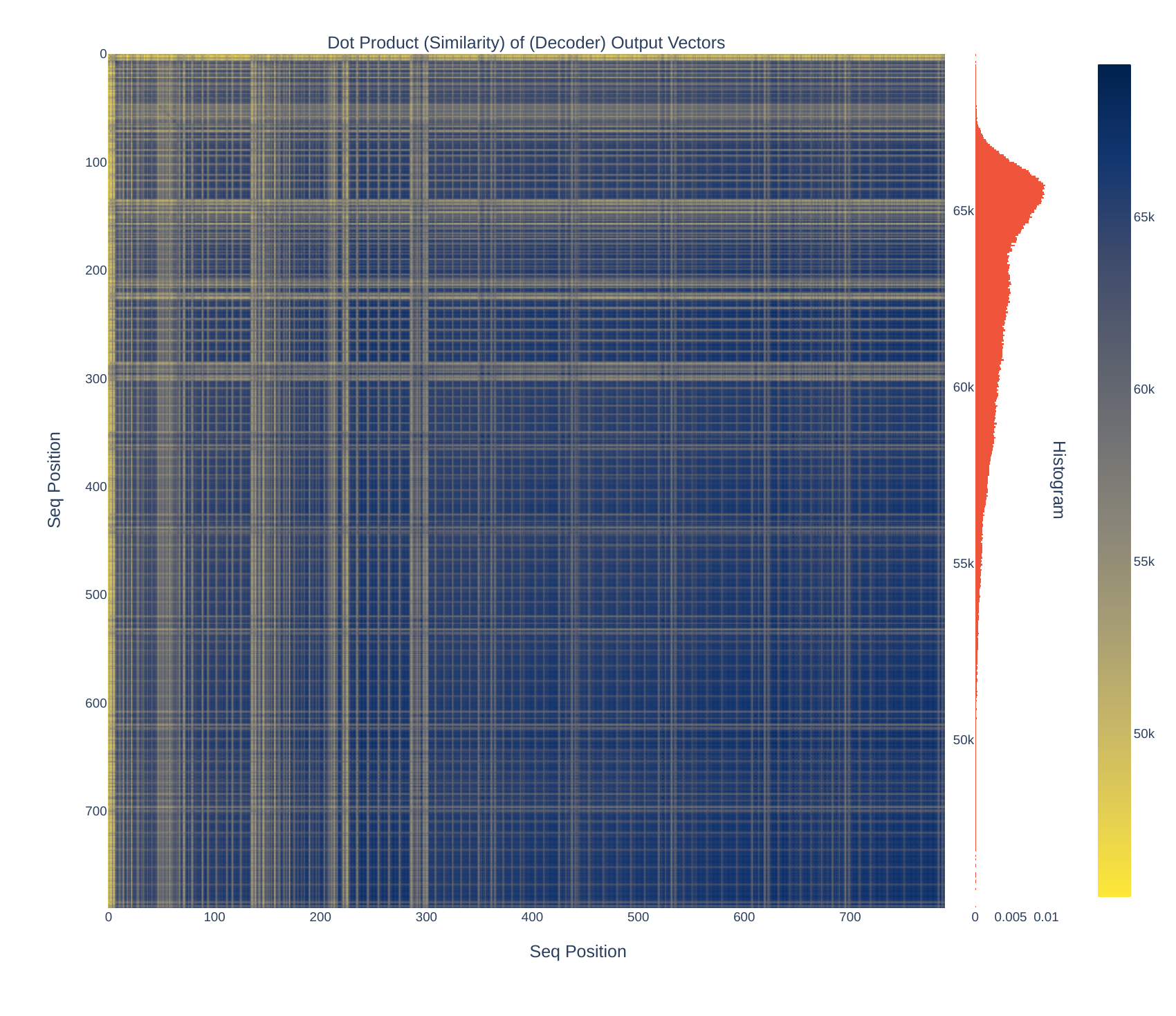}
    \caption{ Inner-product of decoder output vectors ${\langle\vd_i\rangle}$}
    \label{fig-falcon-heatmaps-d}
  \end{subfigure}
\end{center}
\caption{Vector similarity maps for the same input as Fig. \ref{fig-layer-heatmaps}, but passed through a frozen tiiuae/falcon-7b model. The structure of the document and similarity of associated tokens is still clearly visible despite the fact that the model is a generic pretrained model. The fact that the patterns are more visible by cosine-similarity indicates that associated tokens (such as brackets and numbers) cluster together in Embedding Space based on angular separation. Subfigure (d) shows a similar map of the decoder output. The patterns are fuzzier than its encoder stack (T5 and FlanT5) counterparts indicating that there's more mixing in the decoder stack which is not surprising since it's trained to predict a different token than the input.}
\label{fig-falcon-heatmaps}
\end{figure}

Additional similarity maps for the following sample are shown in figures \ref{fig-flant5-heatmaps} and \ref{fig-falcon-heatmaps}.

\subsection{Concept Composition Schemes}
\label{sec-appendix-ablations}
Permutations of the following parameters were used to generate the distributed concept composition schemes that we swept over:
\begin{enumerate}
    \item Encoding Scheme, denoted \emph{encoding\_scheme}. The overal scheme of distributed encoding. Possible values:
    \begin{enumerate}
        \item \emph{sentence\_level\_segmentation}: Segment the context into individual sentences. These then get encoded via the transformer stack based on the ENCODING\_LAYER parameter, yielding a sequence of embedded tokens which are then aggregated (or not) based on the SEGMENT\_AGG\_SCHEME parameter and so on.
        \item \emph{segment\_each\_example}: Segment each example of the context into it's individual sections. For Hellaswag this means, 1) the context passage and 2) the completion passage.
        \emph{merge\_all\_segments}: Each example is segmented same as in \emph{segment\_each\_example}, then all segments across all examples are merged into one single `example'. Doing so circumvents EXAMPLE\_AGG\_SCHEME (but not SEGMENT\_AGG\_SCHEME), since now there's only one example.
        \item \emph{concat\_each\_example}: Each example's text is considered a single segment.
        \item \emph{concat\_all\_examples}: All examples of the context are concatenated into a single sequence which is considered as the one segment of the entire context.
        \item \emph{cross-encoding}: This is the standard way of running a transformer, with no aggregation at any level. A token-sequence is input and a token-sequence is output.
    \end{enumerate}
    \item \emph{ENCODING\_LAYER}: The Transformer stack layer from where to extract the token embedding vector. In case of encoder-decoder models, this refers to the encoder stack. In case of decoder-only models, there's only one stack. Possible values:
    \begin{enumerate}
        \item \emph{E}: Token embedding vector obtained from the input embedding matrix.
        \item A positive integer $l$: Implies the layer whose hidden state is used as token embedding. For an $N$ layer stack, there are $N+1$ possible values, one per layer in the range $0-(N-1)$ and $N$ denoting the top of the stack (after final layer norm). Hidden state of a layer is typically it's input. In the case of decoder-only models like gpt2 that use a single global position encoding, layer 0's (first layer) hidden activation = token vectors $\langle\vw_i\rangle$ (i.e., `E') + global position encodings. 
        \item A negative integer $-l$: Layer number from the top, e.g. -1 means the the top of the stack i.e., the final hidden activation.
        \item \emph{middle}: The middle layer of the stack.
    \end{enumerate}
    \item \emph{OUT\_ENCODING\_LAYER}: Applicable only to Experiment 2: Concept Similarity. The ENCODING\_Layer for the choice sequences.
    \item Normalization, denoted \emph{NORM}: Applied to vectors after an aggregation operation. Possible values:
    \begin{enumerate}
        \item \emph{None}: No normalization.
        \item \emph{L2}: L2 vector-norm.
        \item \emph{varNorm}: normalize vector to have unit variance.
        \item \emph{zNorm}: Normalize to zero mean and unit-variance.
    \end{enumerate}
    \item Word aggregation scheme (denoted WORD\_AGG\_SCHEME) used to aggregate (possibly contextualized) embedded token vectors. Possible values:
    \begin{enumerate}
        \item \emph{mean}: elementwise mean across aggregated vectors.
        \item \emph{w1mean}: weighted elementwise mean across aggregated vectors. Weight of vector at position $t$ is $t/N$ where $N$ is the sequence length.
        \item \emph{relu\textbar mean} - pass the vector through the feed-forward layer's activation function and then perform \emph{mean}. Usually the activation function is a relu, gelu or similar function depending on the model.
        \item \emph{-relu\textbar mean}: same as above except the activation function is reflected about the y-axis i.e., -relu(-x). Now it let's the negative values pass instead of the positive ones.
        \item \emph{relu+\textbar *}, \emph{-relu+\textbar *}: similar to \emph{relu\textbar *} and \emph{-relu\textbar *} respectively except that the original vector is added to the filtered vector (i.e., relu($\ve$)) before being sent on to the next step (mean or w1mean).
        \item \emph{relu\textbar mean}
        \item \emph{last}: Take the last vector $\vd_T$ of the sequence ${\langle\vd_t\rangle}_1^T$ as the aggregated vector. This applies to decoder-only models which apply causal masking.
    \end{enumerate}
    \item \emph{OUT\_WORD\_AGG\_SCHEME}: Applicable only to Experiment 2: Concept Similarity. WORD\_AGG\_SCHEME for the choice sequences.
    \item Segment aggregation scheme, denoted SEGMENT\_AGG\_SCHEME. How to aggregate embedded segment vectors of an example. Applies when there are more than one segments in an example. Possible values are:
    \begin{enumerate}
        \item \emph{mean}: Vector mean (elementwise).
        \item \emph{None}: No aggregation. The segment vector sequence stays untouched and is then passed on to the EXAMPLE\_AGG\_SCHEME.
    \end{enumerate}
    \item EXAMPLE\_AGG\_SCHEME: Example aggregation scheme. How to aggregate embedded example vectors. Applies when there are more than one examples in the context. Possible values:
    \begin{enumerate}
        \item \emph{mean}: Vector mean (elementwise).
        \item \emph{None}: The input vector sequence is passed on unaltered.
        \item \emph{soft\_cluster}: A weighted vector-mean with weights = soft-maxed of dot-product similarity with a choice vector. This implies a different set of weights per choice sequence.
    \end{enumerate}
    \item Similarity Function, denoted SIMILARITY\_FUNC. The function used to compare vectors in Experiment 2. Possible values:
    \begin{enumerate}
        \item \emph{dot-product}: Inner product of vectors.
        \item \emph{cosine-similarity}: Cosine similarity of vectors.
        \item \emph{None}: Where it's not applicable, i.e., in Experiment 1: Pyramidal Generation.
    \end{enumerate}
\end{enumerate}

\begin{figure}
\begin{center}
  \centering
  \begin{subfigure}{1.0\textwidth}
    \includegraphics[width=\linewidth]{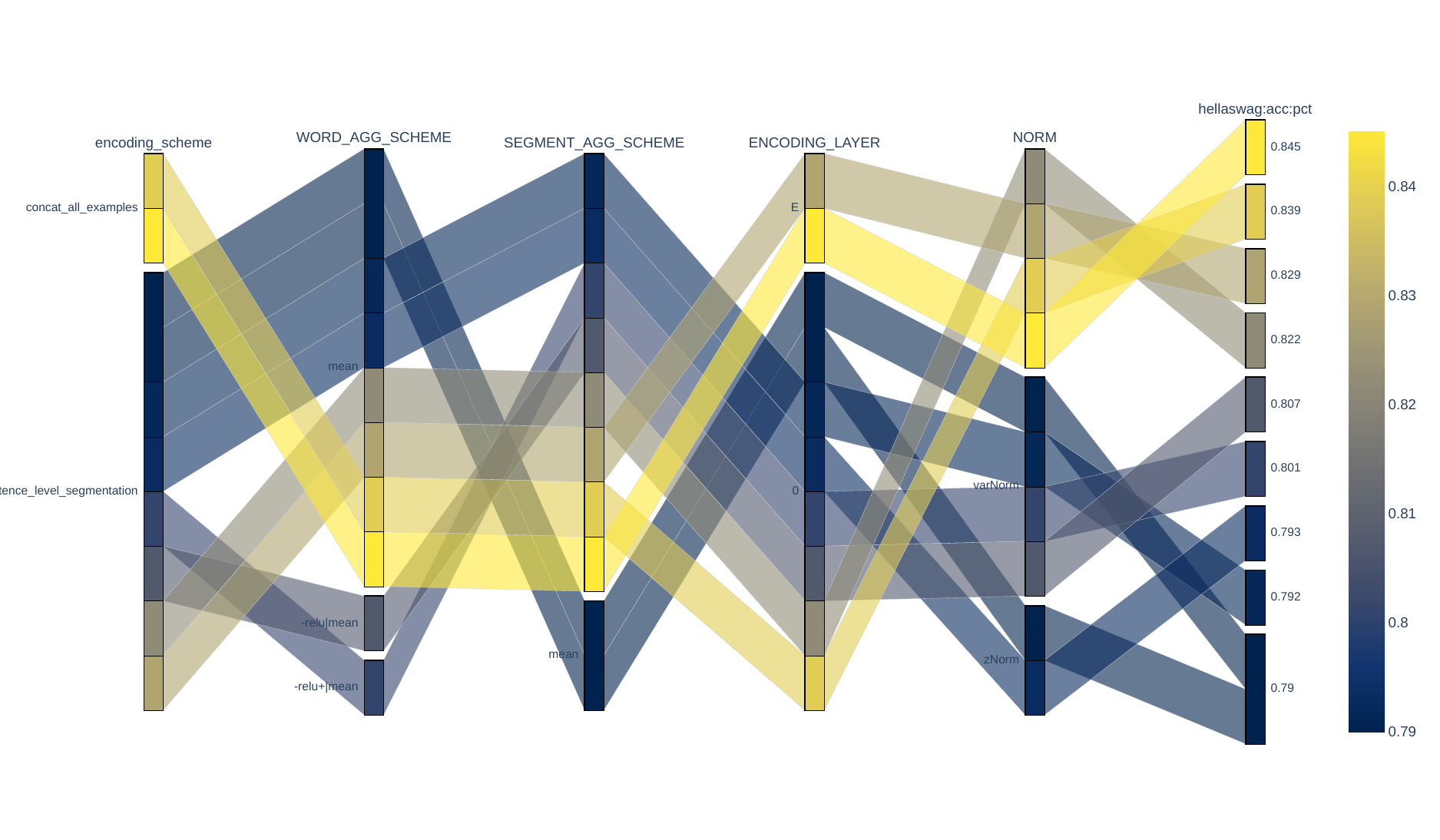}
    \caption{zero-shot}
  \end{subfigure}
  \begin{subfigure}{1.0\textwidth}
    \includegraphics[width=\linewidth]{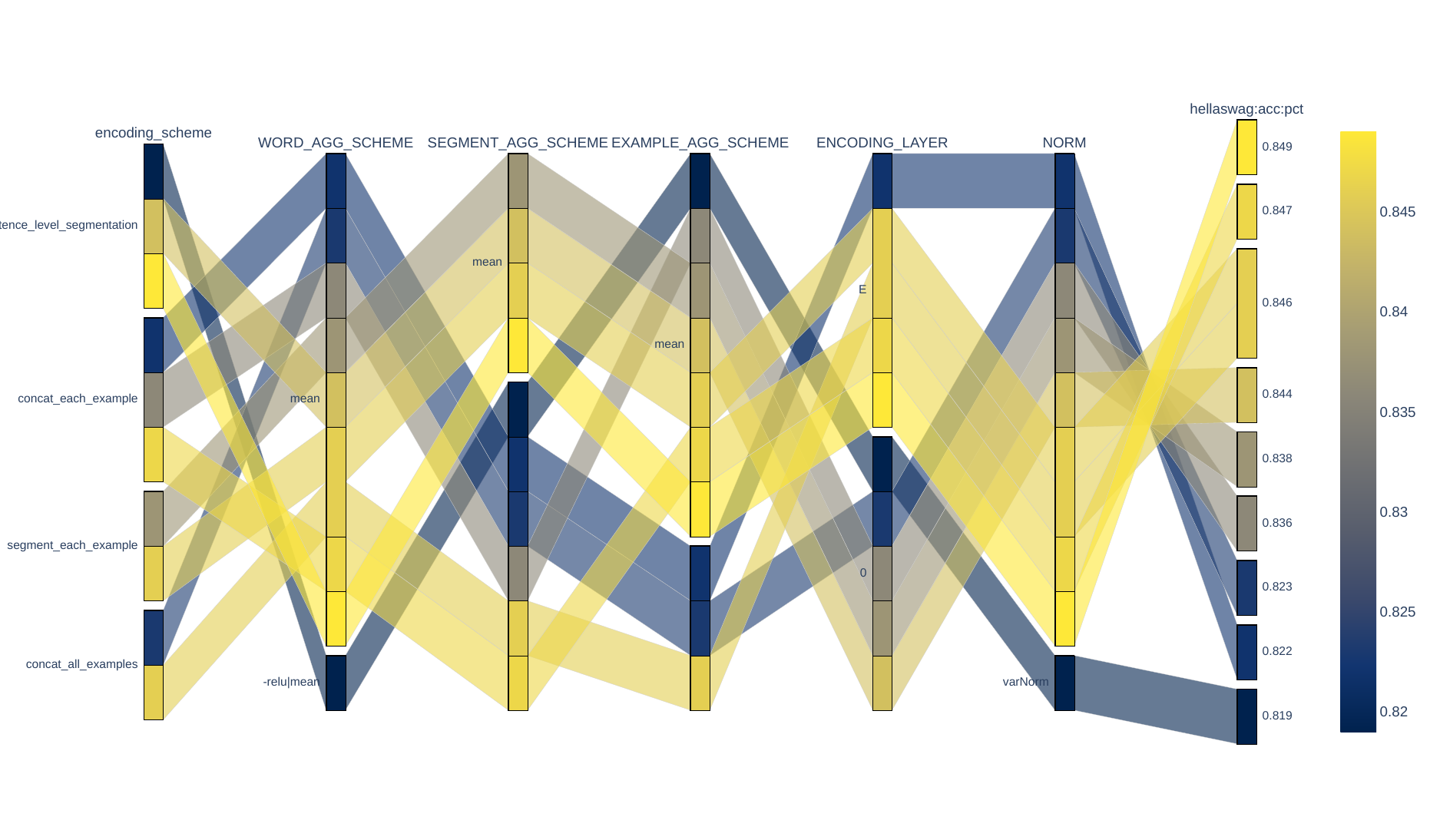}
    \caption{5-shot}
  \end{subfigure}
\end{center}
\caption{Top performing 10 distributed composition schemes respectively for zero-shot (K=0) and few-shot (K=5) Pyramidal Generation experiments (test 1) for model EleutherAI/gpt-neo-1.3B. The top values are around 85\% normalized Hellaswag accuracy score (denoted hellaswag:acc:pct).}
\label{fig-dist-gen-gpt-top-configs}
\end{figure}

\begin{figure}
\begin{center}
  \centering
  \begin{subfigure}{1.0\textwidth}
    \includegraphics[width=\linewidth]{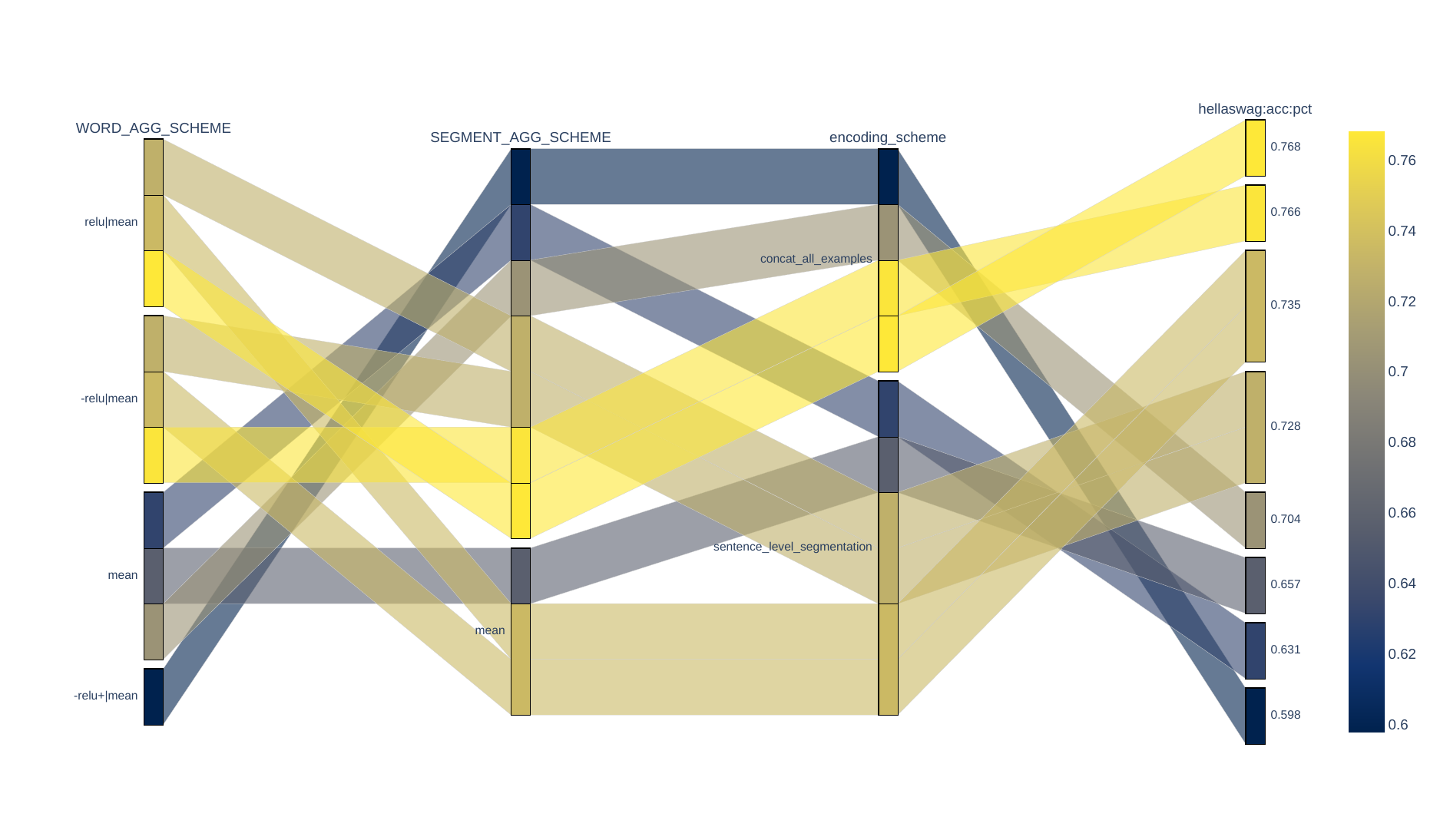}
    \caption{zero-shot, EXAMPLE\_AGG\_SCHEME=None, ENCODING\_LAYER=-1, NORM=None}
  \end{subfigure}
  \begin{subfigure}{1.0\textwidth}
    \includegraphics[width=\linewidth]{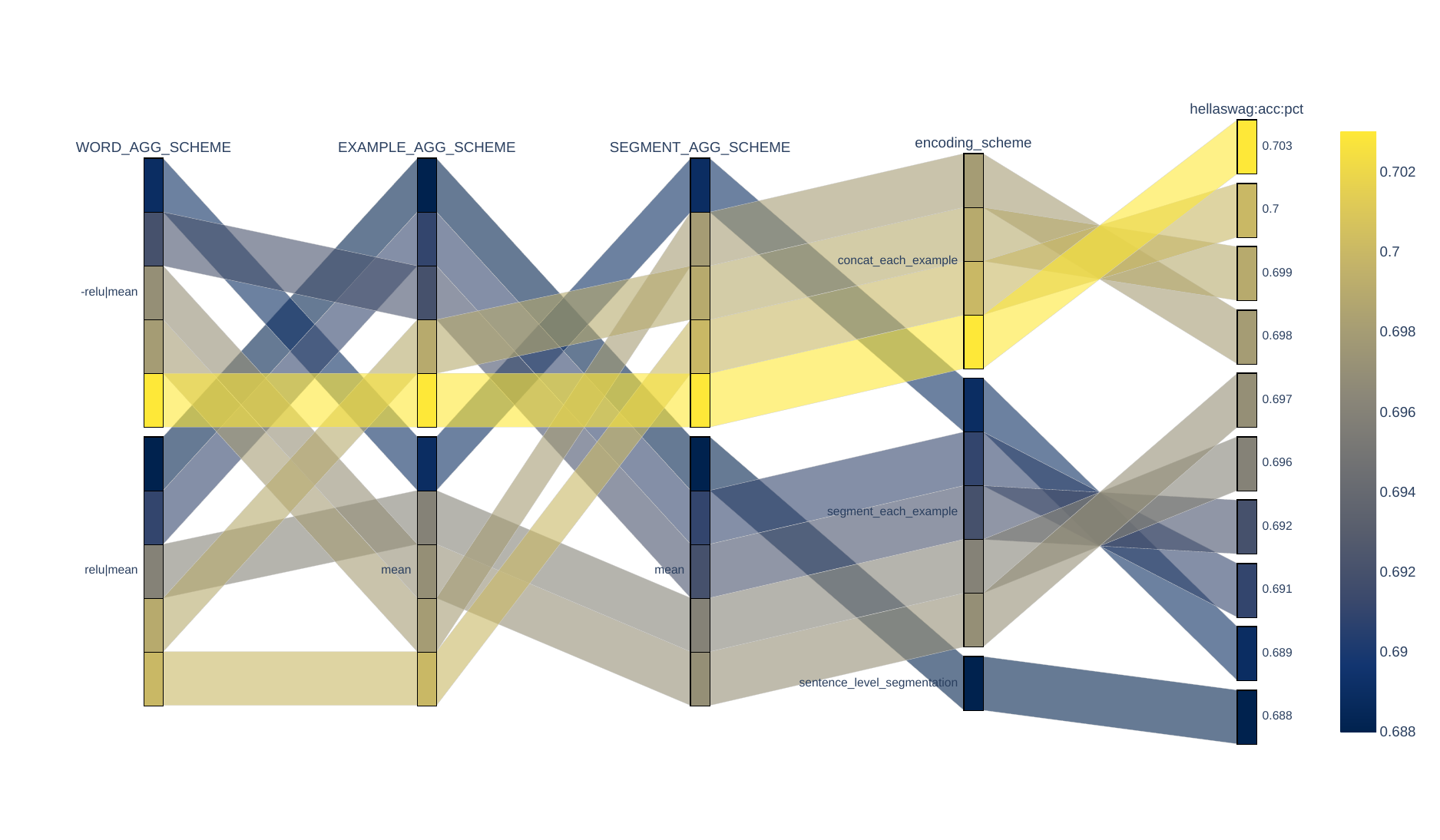}
    \caption{5-shot, ENCODING\_LAYER=-1, NORM=None}
  \end{subfigure}
\end{center}
\caption{Top performing 10 distributed composition schemes respectively for zero-shot (K=0) and few-shot (K=5) Pyramidal Generation experiments for model google/flan-t5-xl. hellaswag:acc:pct stands for the normalized Hellaswag score reported by the test run.}
\label{fig-dist-gen-t5-top-configs}
\end{figure}

\begin{figure}
\begin{center}
  \centering
  \begin{subfigure}{1.0\textwidth}
    \includegraphics[width=\linewidth]{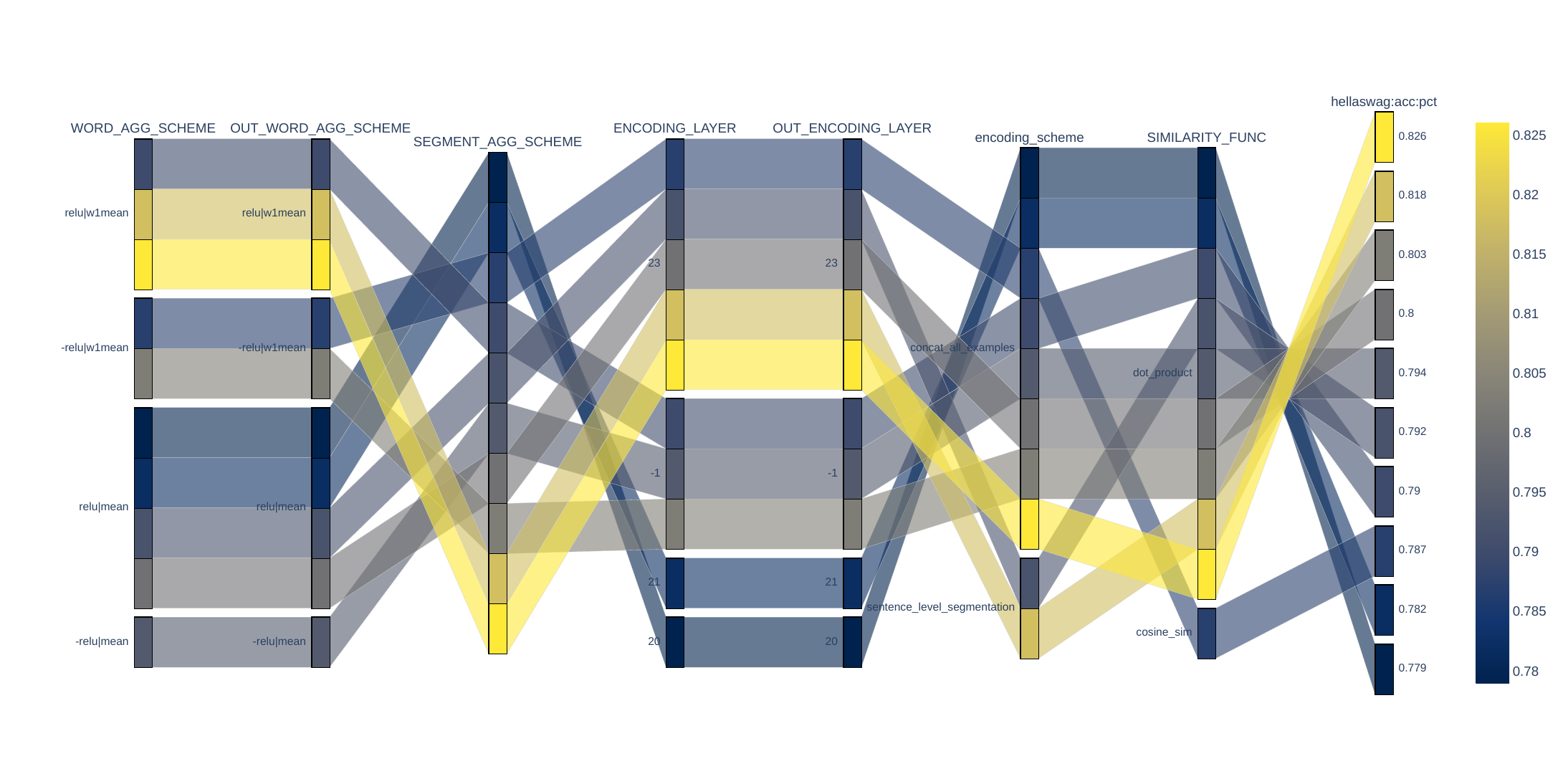}
    \caption{zero-shot. NORM=varNorm,EXAMPLE\_AGG\_SCHEME=None}
  \end{subfigure}
  \begin{subfigure}{1.0\textwidth}
    \includegraphics[width=\linewidth]{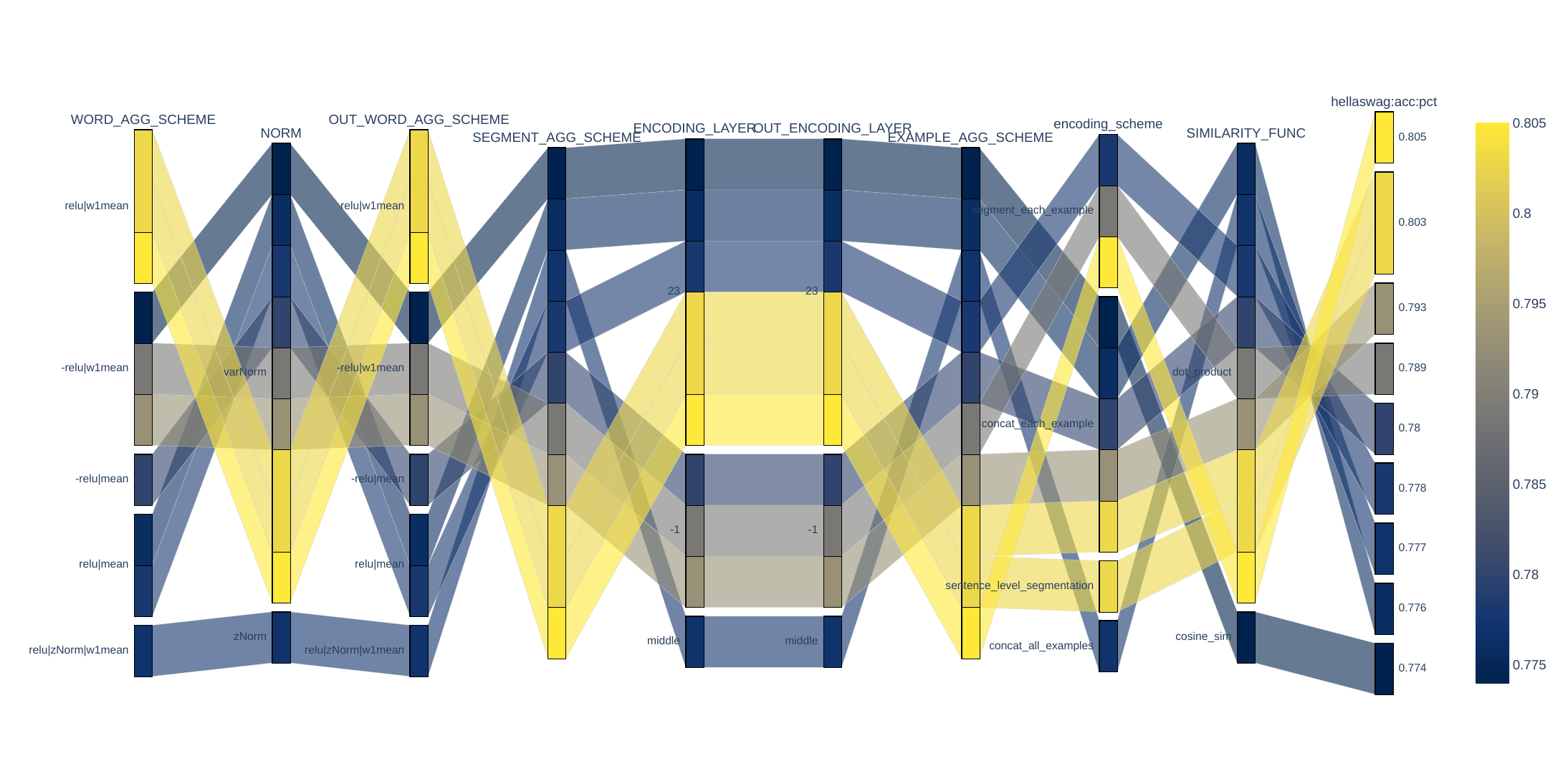}
    \caption{5-shot}
  \end{subfigure}
\end{center}
\caption{Top performing 10 distributed composition schemes of test 2 for model EleutherAI/gpt-neo-1.3B. hellaswag:acc:pct stands for the normalized Hellaswag score reported by the test run.}
\label{fig-dist-sim-gpt-top-configs}
\end{figure}

\begin{figure}
\begin{center}
  \centering
  \begin{subfigure}{1.0\textwidth}
    \includegraphics[width=\linewidth]{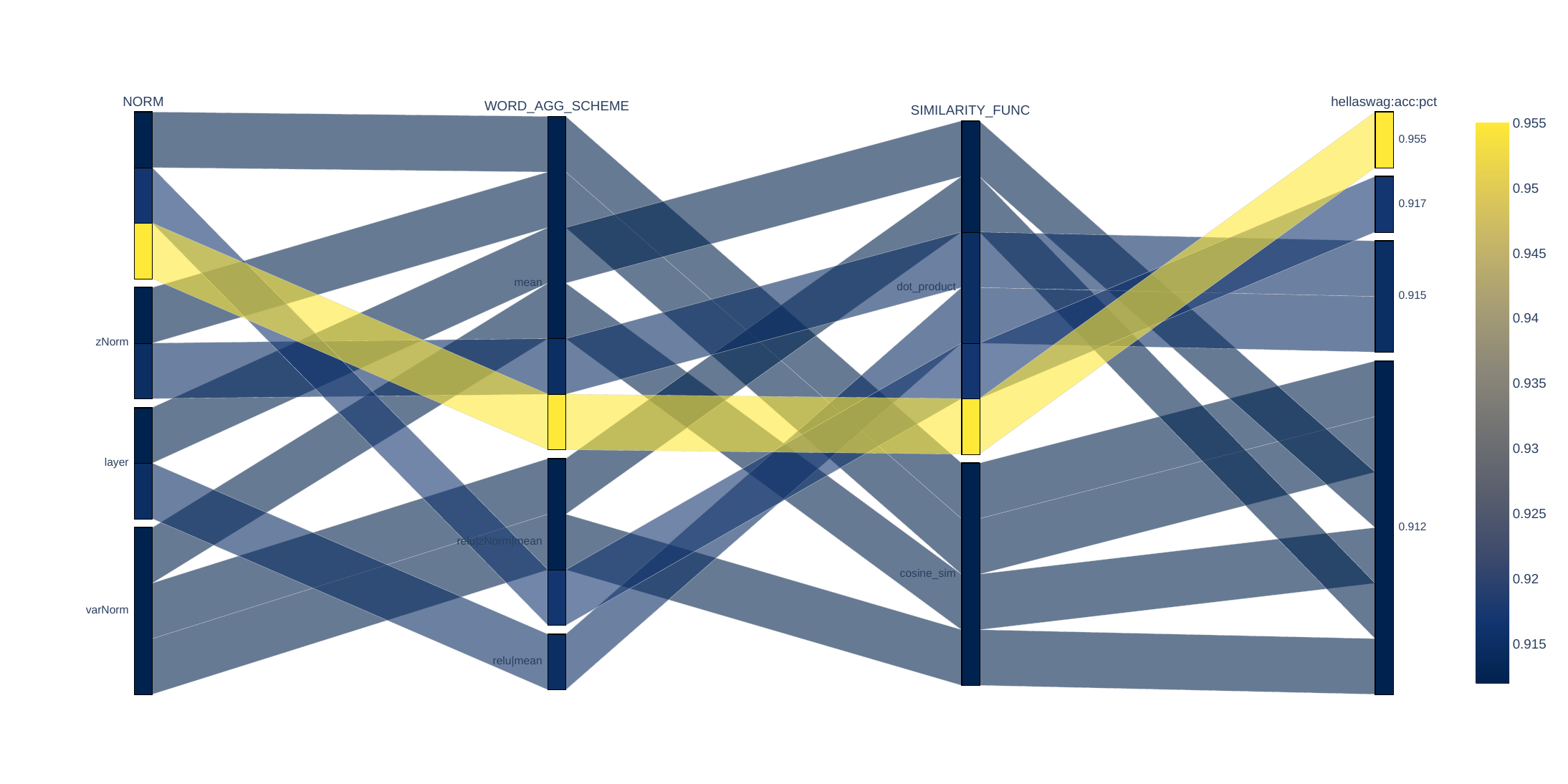}
    \caption{zero-shot, EXAMPLE\_AGG\_SCHEME=None, ENCODING\_LAYER=-1, encoding\_scheme=concat\_all\_examples,\\SEGMENT\_AGG\_SCHEME=None}
  \end{subfigure}
  \begin{subfigure}{1.0\textwidth}
    \includegraphics[width=\linewidth]{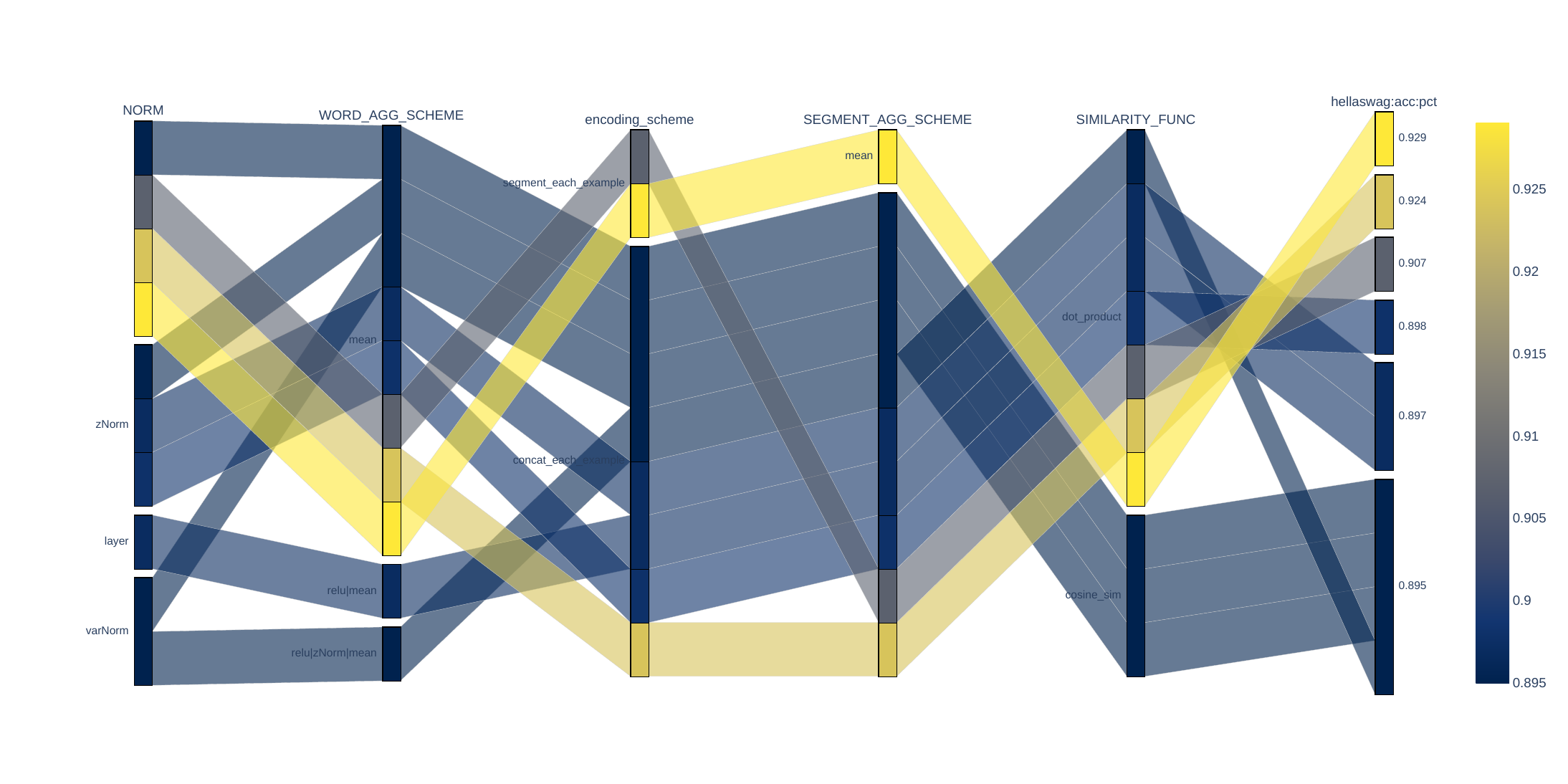}
    \caption{5-shot, ENCODING\_LAYER=-1, NORM=None}
  \end{subfigure}
\end{center}
\caption{Top performing 10 distributed composition schemes of test 2 for model google/flan-t5-xl. hellaswag:acc:pct stands for the normalized Hellaswag score reported by the test run.}
\label{fig-dist-sim-t5-top-configs}
\end{figure}

\end{document}